\documentclass{article}

\usepackage[textheight=8.8in,textwidth=6in,top=1in,headheight=12pt,headsep=25pt,footskip=30pt]{geometry}

\usepackage[utf8]{inputenc} 
\usepackage[T1]{fontenc}    
\usepackage{hyperref}       
\usepackage{url}            
\usepackage{booktabs}       
\usepackage{amsfonts}       
\usepackage{nicefrac}       
\usepackage{microtype}      
\usepackage{xcolor}         

\usepackage{newpxtext}
\usepackage{stmaryrd}		

\usepackage{wrapfig}        

\usepackage{graphicx}
\usepackage{enumitem}
\usepackage{float}
\usepackage{subfig}
\usepackage[font=small]{caption}
\usepackage{hyperref}
\usepackage{xcolor}
\usepackage{amsmath,amsthm,amsfonts,amssymb}
\theoremstyle{definition}
\newtheorem{theorem}{Theorem}[section]
\newtheorem*{theorem*}{Theorem}
\newtheorem*{lemma*}{Lemma}
\newtheorem{corollary}[theorem]{Corollary}
 
\newtheorem{proposition}[theorem]{Proposition}

\newtheorem{defn}[theorem]{Definition}
 
\newtheorem{example}[theorem]{Example} 

\newtheorem{example*}[theorem]{Example*}
\newtheorem{examples*}[theorem]{Examples*}
\newtheorem{remark}[theorem]{Remark}
\newtheorem{remark*}[theorem]{Remark*}

\newtheorem{construction}[theorem]{Construction}

\newtheorem{notation}[theorem]{Notation}

\usepackage{tikz-cd}
\usepackage{macros/tikzfig}

\tikzstyle{blanknode}=[circle, draw=black, fill=black, inner sep=0pt, minimum size=3.5pt]
\tikzstyle{dotblack}=[circle, draw=black, fill=black, inner sep=0pt, minimum size=2pt]
\tikzstyle{dotgray}=[circle, draw=black, fill=gray, inner sep=0pt, minimum size=3.5pt]
\tikzstyle{dotwhite}=[circle, draw=black, fill=white, inner sep=0.5pt, minimum size=3.5pt]
\tikzstyle{dotblue}=[circle, draw=black, fill=blue, inner sep=0pt, minimum size=3.5pt]
\tikzstyle{dotcyan}=[circle, draw=black, fill=cyan, inner sep=0pt, minimum size=3.5pt]
\tikzstyle{dotyellow}=[circle, draw=black, fill=yellow, inner sep=0pt, minimum size=3.5pt]
\tikzstyle{dotorange}=[circle, draw=black, fill=orange, inner sep=0pt, minimum size=3.5pt]
\tikzstyle{dotred}=[circle, draw=black, fill=cbred, inner sep=0pt, minimum size=3.5pt]
\tikzstyle{smalldotred}=[circle, draw=magenta, fill=cbred, inner sep=0pt, minimum size=2pt]
\tikzstyle{smalldotrred}=[circle, draw=magenta, fill=red, inner sep=0pt, minimum size=2pt]
\tikzstyle{smalldotorange}=[circle, draw=cborange, fill=orange, inner sep=0pt, minimum size=2pt]
\tikzstyle{smalldotblue}=[circle, draw=blue, fill=blue, inner sep=0pt, minimum size=2pt]
\tikzstyle{smalldotcyan}=[circle, draw=cbblue, fill=cyan, inner sep=0pt, minimum size=2pt]
\tikzstyle{smalldotcyan2}=[circle, draw=cbblue, fill=white, inner sep=0pt, minimum size=2pt]
\tikzstyle{smalldotgray}=[circle, draw=gray, fill=gray, inner sep=0pt, minimum size=2pt]
\tikzstyle{operadot}=[circle, draw=gray, fill=gray, inner sep=0pt, minimum size=5mm]
\tikzstyle{argoperadot}=[circle, draw=gray, style=dotted, line width = 1pt, fill=white, inner sep=0pt, minimum size=5mm]
\tikzstyle{tinydotwhite}=[inner sep=-0.45mm,minimum width=0mm,minimum height=0mm,draw=white,shape=circle,fill=white]
\tikzstyle{tinydotblack}=[inner sep=0mm,minimum width=1mm,minimum height=1mm,draw,shape=circle,fill=black]
\tikzstyle{blackdot}=[inner sep=0mm,minimum width=1mm,minimum height=1mm,draw,shape=circle,fill=black]

\tikzstyle{bigdotblack}=[circle, draw=black, fill=black, inner sep=0pt, minimum size=7pt]
\tikzstyle{bigdotwhite}=[circle, draw=black, fill=white, inner sep=0pt, minimum size=7pt]
\tikzstyle{dotbigblue}=[circle, draw=black, fill=blue, inner sep=0pt, minimum size=7pt]
\tikzstyle{dotbigcyan}=[circle, draw=black, fill=cyan, inner sep=0pt, minimum size=7pt]
\tikzstyle{dotbigyellow}=[circle, draw=black, fill=yellow, inner sep=0pt, minimum size=7pt]
\tikzstyle{dotbigorange}=[circle, draw=black, fill=orange, inner sep=0pt, minimum size=7pt]
\tikzstyle{dotbigred}=[circle, draw=black, fill=cbred, inner sep=0pt, minimum size=7pt]

\tikzstyle{thickdotblack}=[circle, draw=black, fill=black, inner sep=0pt, minimum size=3.5pt, line width=1pt]
\tikzstyle{thickdotgray}=[circle, draw=gray, fill=gray, inner sep=0pt, minimum size=3.5pt, line width=1pt]
\tikzstyle{thickdotwhite}=[circle, draw=black, fill=white, inner sep=0pt, minimum size=3.5pt, line width=1pt]
\tikzstyle{thickdotblue}=[circle, draw=black, fill=blue, inner sep=0pt, minimum size=3.5pt, line width=1pt]
\tikzstyle{thickdotcyan}=[circle, draw=black, fill=cyan, inner sep=0pt, minimum size=3.5pt, line width=1pt]
\tikzstyle{thickdotyellow}=[circle, draw=black, fill=yellow, inner sep=0pt, minimum size=3.5pt, line width=1pt]
\tikzstyle{thickdotorange}=[circle, draw=black, fill=orange, inner sep=0pt, minimum size=3.5pt, line width=1pt]
\tikzstyle{thickdotred}=[circle, draw=black, fill=cbred, inner sep=0pt, minimum size=3.5pt, line width=1pt]

\tikzstyle{basic rounded box}=[draw, rectangle, rounded corners, minimum height=1.2em, minimum width=1.4em]
\tikzstyle{small rounded box}=[draw, fill=white, rectangle, rounded corners, minimum height=1.2em, minimum width=1.4em, node font={\scriptsize}]

\tikzstyle{circl}=[draw, fill=white, circle, inner sep=0.5pt, node font={\scriptsize}]
\tikzstyle{smolsquare}=[inner sep=0mm,minimum width=3mm,minimum height=3mm,draw,shape=rectangle,fill=white]
\tikzstyle{square}=[inner sep=0mm,minimum width=5mm,minimum height=5mm,draw,shape=rectangle,fill=white]
\tikzstyle{thicksquare}=[draw, fill=white, regular polygon, regular polygon sides=4, inner sep=0pt, line width=1pt]
\tikzstyle{squareblue}=[draw, fill=blue, regular polygon, regular polygon sides=4, inner sep=0pt, minimum size=10pt]
\tikzstyle{squarecyan}=[draw, fill=cyan, regular polygon, regular polygon sides=4, minimum size=10pt, inner sep=0pt]
\tikzstyle{squareyellow}=[draw, fill=yellow, regular polygon, regular polygon sides=4, minimum size=10pt, inner sep=0pt]
\tikzstyle{squareorange}=[draw, fill=orange, regular polygon, regular polygon sides=4, minimum size=10pt, inner sep=0pt]
\tikzstyle{squarered}=[draw, fill=cbred, regular polygon, regular polygon sides=4, minimum size=10pt, inner sep=0pt]
\tikzstyle{thicksquareorange}=[draw, fill=orange, regular polygon, regular polygon sides=4, inner sep=0pt, minimum size=10pt, line width=1pt]

\tikzstyle{antipode}=[draw, fill=cbred, regular polygon, regular polygon sides=4, inner sep=0pt, minimum size=8pt, tikzit shape=rectangle]
\tikzstyle{leq}=[draw, fill=white, shape=regular polygon, regular polygon sides=3, rotate=90, scale=0.2]
\tikzstyle{geq}=[draw, fill=white, shape=regular polygon, regular polygon sides=3, rotate=270, scale=0.2]
\tikzstyle{geqgray}=[draw=gray, fill=white, shape=regular polygon, regular polygon sides=3, rotate=270, scale=0.5]
\tikzstyle{upground}=[circuit ee IEC,thick,ground,rotate=0,scale=0.35]
\tikzstyle{leqblue}=[draw, fill=blue, shape=regular polygon, regular polygon sides=3, rotate=90, scale=0.5]
\tikzstyle{leqcyan}=[draw, fill=cyan, shape=regular polygon, regular polygon sides=3, rotate=90, scale=0.5]
\tikzstyle{leqyellow}=[draw, fill=yellow, shape=regular polygon, regular polygon sides=3, rotate=90, scale=0.5]
\tikzstyle{leqorange}=[draw, fill=orange, shape=regular polygon, regular polygon sides=3, rotate=90, scale=0.5]
\tikzstyle{leqred}=[draw, fill=cbred, shape=regular polygon, regular polygon sides=3, rotate=90, scale=0.5]
\tikzstyle{leqthick}=[draw, fill=white, shape=regular polygon, regular polygon sides=3, rotate=90, scale=0.5, line width=1pt]
\tikzstyle{leqthickblue}=[draw, fill=blue, shape=regular polygon, regular polygon sides=3, rotate=90, scale=0.5, line width=1pt]
\tikzstyle{geqblue}=[draw, fill=blue, shape=regular polygon, regular polygon sides=3, rotate=270, scale=0.75]
\tikzstyle{geqcyan}=[draw, fill=cyan, shape=regular polygon, regular polygon sides=3, rotate=270, scale=0.75]
\tikzstyle{geqyellow}=[draw, fill=yellow, shape=regular polygon, regular polygon sides=3, rotate=270, scale=0.75]
\tikzstyle{geqorange}=[draw, fill=orange, shape=regular polygon, regular polygon sides=3, rotate=270, scale=0.75]
\tikzstyle{geqred}=[draw, fill=cbred, shape=regular polygon, regular polygon sides=3, rotate=270, scale=0.75]

\tikzstyle{H}=[-, style=dashed]
\tikzstyle{gray}=[-, draw=gray]
\tikzstyle{brace edge}=[-, decorate, decoration={brace,amplitude=1mm,raise=-1mm}]

\tikzstyle{blue}=[-, draw=blue]
\tikzstyle{cyan}=[-, draw=cbblue]
\tikzstyle{yellow}=[-, draw=Gold1]
\tikzstyle{orange}=[-, draw=cborange]
\tikzstyle{red}=[-, draw=magenta]
\tikzstyle{rred}=[-, draw=black, fill=cbjade]

\tikzstyle{v}=[->]
\tikzstyle{vgray}=[->, draw=gray]
\tikzstyle{vblue}=[->, draw=blue]
\tikzstyle{vcyan}=[->, draw=cbblue]
\tikzstyle{vyellow}=[->, draw=Gold1]
\tikzstyle{vorange}=[->, draw=cborange]
\tikzstyle{vred}=[->, draw=magenta]
\tikzstyle{vpurp}=[->, draw=gray, line width = 0.75pt]

\tikzstyle{K}=[-, line width=1pt]
\tikzstyle{Kblue}=[-, draw=blue, line width = 1pt]
\tikzstyle{Kcyan}=[-, draw=cbblue, line width = 1pt]
\tikzstyle{Kyellow}=[-, draw=Gold1, line width = 1pt]
\tikzstyle{Korange}=[-, draw=cborange, line width = 1pt]
\tikzstyle{Kred}=[-, draw=magenta, line width = 1pt]

\tikzstyle{D}=[-, style=dashed]
\tikzstyle{Dblue}=[-, draw=blue, style=dashed]
\tikzstyle{Dcyan}=[-, draw=cbblue, style=dashed]
\tikzstyle{Dyellow}=[-, draw=Gold1, style=dashed]
\tikzstyle{Dorange}=[-, draw=cborange, style=dashed]
\tikzstyle{Dred}=[-, draw=red, style=dashed]
\tikzstyle{Dpurp}=[-, draw=gray, style=dashed, line width = 0.75pt]

\tikzstyle{finner}=[-, draw=black, fill=white]
\tikzstyle{finnerd}=[-, draw=black, style=dashed, fill=white]
\tikzstyle{fo}=[-, draw=black, fill=orange]
\tikzstyle{fc}=[-, draw=cbblue, fill=cyan]
\tikzstyle{fb}=[-, draw=black, fill=blue]
\tikzstyle{tfc}=[-, draw=black, fill=cyan, line width = 1pt]
\tikzstyle{tfb}=[-, draw=black, fill=blue, line width = 1pt]
\tikzstyle{fr}=[-, draw=magenta, fill=cbred]
\tikzstyle{fy}=[-, draw=black, style=dashed, fill=cbjade]
\tikzstyle{kF}=[-, draw=none, fill=black]

\def\bR{\begin{color}{red}}  
\def\bB{\begin{color}{blue}}
\def\bM{\begin{color}{magenta}}  
\def\bC{\begin{color}{cyan}}
\def\bW{\begin{color}{white}}
\def\bBl{\begin{color}{black}}
\def\bG{\begin{color}{green}}
\def\bY{\begin{color}{yellow}}
\def\e{\end{color}}
\newcommand{\bit}{\begin{itemize}}
\newcommand{\eit}{\end{itemize}\par\noindent}
\newcommand{\ben}{\begin{enumerate}}
\newcommand{\een}{\end{enumerate}\par\noindent}
\newcommand{\beq}{\begin{equation}}
\newcommand{\eeq}{\end{equation}\par\noindent}
\newcommand{\beqa}{\begin{eqnarray*}}
\newcommand{\eeqa}{\end{eqnarray*}\par\noindent}
\newcommand{\beqn}{\begin{eqnarray}}
\newcommand{\eeqn}{\end{eqnarray}\par\noindent}

\definecolor{cbred}{RGB}{162,4,162}
\definecolor{cborange}{RGB}{251,145,10}
\definecolor{cbblue}{RGB}{5,162,162}
\definecolor{cbjade}{RGB}{204,234,207}

\title{On the Anatomy of Attention}

\newcommand\extrafootertext[1]{%
    \bgroup
    \renewcommand\thefootnote{\fnsymbol{footnote}}%
    \renewcommand\thempfootnote{\fnsymbol{mpfootnote}}%
    \footnotetext[0]{#1}%
    \egroup
}

\author{\makebox[3in]{\begin{minipage}{8in}\centering
  Nikhil Khatri$^{*}$ \quad 
  Tuomas Laakkonen$^{*}$ \quad 
  Jonathon Liu$^{*}$ \quad 
  Vincent Wang-Ma\'{s}cianica$^{*\ddag}$ \\[1em]
  Compositional Intelligence, Quantinuum \\
  17 Beaumont St., Oxford OX1 2NA, UK \\
  {\small \texttt{\{nikhil.khatri,tuomas.laakkonen,jonathon.liu,vincent.wang\}@quantinuum.com}}\\[1em]
  $^{\ddag}$Department of Computer Science, University of Oxford \\
  7 Parks Rd, Oxford OX1 3QG, UK
\end{minipage}}}

\date{}

\begin{document}

\maketitle
\extrafootertext{$^*$ Equal contribution}

\begin{abstract}

We introduce a category-theoretic diagrammatic formalism in order to systematically relate and reason about machine learning models. 
Our diagrams present architectures intuitively but without loss of essential detail, where natural relationships between models are captured by graphical transformations, and important differences and similarities can be identified at a glance. 
In this paper, we focus on attention mechanisms: translating folklore into mathematical derivations, and constructing a taxonomy of attention variants in the literature. 
As a first example of an empirical investigation underpinned by our formalism, we identify recurring anatomical components of attention, which we exhaustively recombine to explore a space of variations on the attention mechanism. 

\end{abstract}

\section{Introduction}

Many efforts have been directed towards providing an overarching framework for different deep learning (DL) architectures, which have gained interest in light of the massive proliferation of variants of the transformer \cite{vaswani2017attention}. These frameworks have come in various forms, such as taxonomic surveys \cite{linSurveyTransformers2021} that organise certain groups of architectures, and more prescriptive high-level conceptual characterisations~\cite{bronsteinGeometricDeepLearning2021a}. However, taxonomies risk arbitrariness in their organising criteria and theoretical frameworks risk abstracting away important practical details. The ideal would be to build taxonomies and organising theories starting from the precise computational descriptions of architectures as they occur ``in the wild''. How to systematically notate and communicate these computational descriptions in a human-friendly manner is a question as old as computer science \cite{goldstinePlanningCodingProblems1947}, and the tried-and-true solution remains the same even today: flowcharts. A mathematical fact that deserves to be better known is that flowcharts, of the same sort that are customary in DL papers introducing architectures, are often already formal representations with unambiguous (but implicit) semantics in the mathematical setting of smooth functions between Euclidean spaces and their sequential and parallel composites~\cite{selingerSurveyGraphicalLanguages2010a}. Even setting aside the formality of flowcharts, in our view there remain two fatal drawbacks from the practitioner's perspective that, to the best of our knowledge, every graphical notational system for DL currently on offer suffers from.

First, there is an inescapable tension between formal detail and the vantage point of abstraction that, for example, allows the practitioner to intuitively grasp a novel architecture and appreciate its conceptual differences from its predecessors. In practice this problem manifests as a dichotomy between overly informal presentations of architectures such as high-level flowcharts, with the only detailed alternatives being overly detailed prose or pseudocode. This problem cannot be solved by simply increasing the formal detail of a notational system: consider that in the limit, one could describe a transformer in GPU assembly language with perfect formal fidelity, but such a description would be completely unilluminating.

Second, even if one has a notational system for ML that is suitably formal and at the desired level of detail, without a \emph{rewrite system}, one has to step outside the notation to reason about its contents with other means. This problem cannot fundamentally be solved by any amount of descriptive adequacy for individual architectures that a notational system might have, without additional machinery to compare representations. All notational systems we know of for DL provide no framework for relating the computational structure of models outside of, at best, what is already granted by the equational theory in which they are grounded. Consider that from the practitioner's perspective, it is informative to know when two architectures only differ by ``introducing a residual around a particular subprocess'' e.g., but this would make the architectures mathematically unequal, and hence formally incomparable in purely equational theories, such as that of continuous maps between Euclidean spaces.

We introduce string diagrams\footnote{String diagrams are a diagrammatic syntax that have been used to formally represent and reason about linear and affine algebra \cite{sobocinski_graphical_2015,bonchi_interacting_2017,bonchi_graphical_2019}, first-order logic \cite{haydon_compositional_2020}, 
causal models \cite{lorenz2023causal,jacobs2019causal}, signal flow graphs \cite{bonchi_categorical_2014}, electrical circuits \cite{boisseau_string_2022}, game theory \cite{hedges_string_2015}, petri nets \cite{baez_open_2020}, probability theory \cite{fritz_finettis_2021}, 
formal linguistics \cite{coecke_mathematical_2010,wang-mascianica_distilling_2023}, quantum theory \cite{coecke_picturing_2017,poor_completeness_2023}, and aspects of machine learning such as backpropagation \cite{cruttwell_categorical_2022}.} for reasoning about arbitrary DL architectures, which we demonstrate in this paper in the context of attention mechanisms. Our notation is formally grounded in category theory and is similar in intent to the systems of \cite{abbottNeuralCircuitDiagrams2024} and \cite{taylorGraphicalTensorNotation}, but the crucial difference is that we address the two aforementioned issues. First, we solve the problem of trading-off between formal detail and visual abstraction by making all the choices simultaneously: we provide multiple abstraction levels that our formalism allows us to move between freely. This allows us to rise from the code-level formality suited for computers (Example \ref{ex:vanilla}), to an ``engineer's perspective'' suited for practitioners (Example \ref{ex:OG}). The novel mathematical machinery that enables this (elaborated in Appendix \ref{subsec:simd}) is the ability to formally decorate string diagrams (in symmetric monoidal categories) with instructions for reiteration of the same processes repeatedly in parallel --- similarly to box notation for iterated variables in plate notation \cite{buntineOperationsLearningGraphical1994} --- which allows us to compactly represent computation with tensors. In the language of high-performance computing, this is called \emph{Single Instruction (applied to) Multiple Data} (SIMD), which is particularly germane to 
transformers since parallelisation is one of the primary enablers of modern ML performance \cite{goodfellow_deep_2016, pandey2022transformational}.

Second, we address the problem of formal expressivity by introducing a rewrite system based on a minimal and principled basis: the Universal Approximation Theorem \cite{cybenkotApproximationSuperpositionsSigmoidal}, which may safely be assumed to hold for even relatively shallow feedforward networks \cite{ismailovThreeLayerNeural2022}. Fixed input-output universal function approximators may be formally viewed as the functional equivalent of variables waiting to be assigned concrete values, and so they admit a theory of rewrites that visually amounts to graphically substituting a universal approximator with any other composite function one likes --- even a composite that itself contains universal approximators. The categorical machinery that underpins the semantics of universal approximators as ``holes to be filled" is elaborated in Appendix~\ref{appendix:univ}, along with a proof of coherence in Appendix~\ref{appendix:coherence} that guarantees the well-definedness and well-behaviour of the parallelism notation with the non-equality universal approximator rewrites. What this buys us is the ability to continue abstracting from the engineering perspective to the ``scientist's perspective'' where we demonstrate that we can formally retrace the folkloric ``evolutionary process'' of architectures such as the development of the vanilla transformer \cite{vaswani2017attention} from its precursor \cite{bahdanauNeuralMachineTranslation2016}, the subsequent derivation of linear transformers (Section \ref{sec:evolution}), and ultimately accommodate many architectures at once in a family tree (Section \ref{sec:tax}).

We demonstrate the potential of our graphical perspective by posing and testing a question that we stand in a unique position to answer: is the structure of an attention mechanism important for its performance? In Section \ref{sec:experiments} we gather a family of commonly recurring anatomical components among transformer variants, which we are able to obtain by direct inspection of many architectures. We then exhaustively enumerate the possible composites for up to five such components and use our rewrite system to reduce trivial overparameterisations, thus obtaining 14 distinct attention mechanisms including the linear transformer \cite{wangLinformerSelfAttentionLinear2020} and the classic \cite{vaswani2017attention}. We answer in the negative: the structure of the attention mechanism does not seem to affect its performance on a representative task.

\section{Formally depicting architectures}
For our diagrams, we consider semantics in what is essentially the cartesian monoidal category of Euclidean spaces (viewed as vector spaces) and continuous functions between them, elaborated in Definition \ref{defn:SIMDformaldefn}, and Remark \ref{rem:porbs} for the immediate extension to the probabilistic setting. For now we are content with an accessible presentation. Wires represent spaces, and boxes represent functions. We read our diagrams from left to right, where sequential composition is the usual function composition, and parallel composition of spaces is by cartesian product. We include (for all spaces) the identity function, copy-maps, deletions, and swapping in our basic stock of diagrams.
\[\input{prelims.tikz}\]

String diagrams have the appealing characteristic of ensuring that visually intuitive equivalences in information flows correspond to symbolic derivations of behavioural equivalence. This means that the cumbersome algebraic proofs needed to demonstrate the equality of sequentially and parallel-composed processes are naturally handled by diagram isotopies. In diagrammatic syntax, such isomorphisms are conventionally written as plain equalities. Interested readers are directed to \cite{selingerSurveyGraphicalLanguages2010a} for more information. For example:
\[
\scriptsize{
\begin{aligned}
&(\mathbf{1} \oplus \theta) \circ (\Delta \oplus g) \circ (f \oplus \mathbf{1}) & \\
&\simeq (\mathbf{1} \oplus \theta) \circ (\mathbf{1} \oplus \mathbf{1} \oplus g) \circ (\Delta \oplus 1) \circ (f \oplus \mathbf{1}) & \text{[Identity, interchange]} \\
&\simeq (\mathbf{1} \oplus g \oplus \mathbf{1}) \circ (\mathbf{1} \oplus \theta) \circ (\Delta \oplus 1) \circ (f \oplus \mathbf{1}) & \text{[Braid naturality]} \\
&\simeq (\mathbf{1} \oplus g \oplus \mathbf{1}) \circ (\mathbf{1} \oplus \theta) \circ (f \oplus f \oplus 1) \circ (\Delta \oplus 1) & \text{[Copy naturality]}\\
&\simeq (\mathbf{1} \oplus g \oplus \mathbf{1}) \circ (f \oplus \mathbf{1} \oplus f) \circ (1 \oplus \theta) \circ (\Delta \oplus 1) & \text{[Braid naturality]} \\
\end{aligned}
\quad \Leftrightarrow \quad
\input{bur-combine.tikz}
}\]

Since we work with Euclidean spaces as wires, without loss of generality, an arbitrary $f:\mathbb{R}^N\to 
\mathbb{R}^M$ is equivalently expressed as a map from $N$ to $M$ parallel iterations of $\mathbb{R}$. As syntactic sugar, we indicate copies with a labelled tick under a wire.
\[\input{intro/eucfn.tikz}\]
Just as $\mathbb{R}^N$ may be viewed as the space of length-$N$ vectors taking values in $\mathbb{R}$, $\mathbb{R}^{(J \times K)}$ may be viewed as the space of $\mathbb{R}$-valued $J$-by-$K$ matrices. Generalising, we let ticks on wires indicate indices of tensors of arbitrary rank, and we allow functions between spaces of $\mathbb{R}$-valued tensors.
\[\input{intro/tensor_j.tikz}\]

A \emph{SIMD-box} is a dashed container that indicates parallel iteration of the enclosed map. 
By nesting SIMD-boxes we may iterate functions elementwise over arbitrary tensors, and we allow an optional residual to be passed along through iterates\footnote{Sequential iteration with a residual is precisely the computational graph of a recurrent neural network at inference-time. Mathematically, parallel composition is extensionally equivalent to residual-iteration with empty residual, but this is qualitatively different computationally, due to the improved runtime afforded by data-parallelisation.}.
\[\input{intro/SIMD_j.tikz}\]
Ticks and SIMD-boxes on string diagrams suffice to articulate several common operations that we sugar. From left to right, we \emph{share} an input by copying it; we \emph{reshape} tensors by rearranging their entries, which includes reversing their order, and combining tensors by concatenation; and we \emph{contract tensors}\footnote{Generally it should be clear which two indices have been contracted, but in cases with ambiguity this can be made explicit by adding superscripts to index labels (see opening remarks in Appendix~\ref{sec:detailed_appendix}).} along shared dimensions, which specialises to matrix multiplication for two rank-2 inputs, vector dot product for two rank-1 inputs, and scalar multiplication for two rank-0 inputs. 
\[\input{intro/SIMD2.tikz} \qquad \input{intro/reshape_j.tikz} \qquad \input{intro/tensorcontract_j.tikz}\]
The final notational aspect we introduce is a distinction between processes with learnable parameters -- which we depict as filled -- and processes with no learnable parameters, which are depicted without fill. Before continuing, we wish to impress that these trivial notational observations together form an unambiguous basis to describe arbitrary architectures, and the reader is referred to the appendix for formal semantics and comparisons to other notational systems in the literature.

\begin{figure}[H]
\centering
\makebox[\textwidth]{$$\scalebox{1.0}{\input{biblical_jono_pink.tikz}}$$}
\begin{example}[The vanilla transformer]\label{ex:vanilla} As defined by \cite{vaswani2017attention}, a transformer (encoder) consists of data indexed by arbitrary but fixed positive integers; the \emph{token vocabulary dimension} $v$; the number of \emph{attention heads} $h$; the \emph{head dimension} $d_h$; 
the \emph{model dimension} $d_m = h \times d_h$; the \emph{feedforward dimension} $d_f$; the number of \emph{layers} $N$; and the (supplied at inference time) \emph{sequence length} $s$.
For each choice of integers, the training process supplies concrete values for learnable tensors; the \emph{embedding} and \emph{unembedding} matrices $W^E \in \mathbb{R}^{(v \times d_m)}$ and $W^U \in \mathbb{R}^{(d_m \times v)}$; 
layer norm parameters $\beta_l, \gamma_l \in \mathbb{R}^{d_m}$ for $l\leqslant 2N+1$;
feedforward parameters $b^1_j \in \mathbb{R}^{d_f}$, 
$b^2_j \in \mathbb{R}^{d_m}$, 
$w^1_j \in \mathbb{R}^{(d_m \times d_f)}$, 
and $w^2_j \in \mathbb{R}^{(d_f \times d_m)}$ for each layer $j\leqslant N$;
$W^O_j \in \mathbb{R}^{(d_m \times d_m)}$ for each layer $j \leqslant N$; 
and famously, the \emph{value}, \emph{query}, and \emph{key} matrices $W^V_{j,k}, W^Q_{j,k}, W^K_{j,k} \in \mathbb{R}^{(d_m \times d_h)}$ for each attention-head $k \leqslant h$ in each layer $j\leqslant N$. 
The processes and values without learnable parameters are; the \emph{positional encoding} vectors
$(v_i)_k = \cos(i \times 10000^{-\frac{k-1}{d_m}})$ for odd $k$ and $\sin(i \times 10000^{-\frac{k}{d_m}})$ otherwise, where $i$ is the sequence index and $k\leq d_m$;
the \emph{layer norm} LN*$(v,\gamma,\beta)\triangleq \big( \frac{v - \mathop{\mathbb{E}}[v]}{\sqrt{\mathop{Var}(v) + \varepsilon}} \odot \gamma + \beta \big)$ where $\varepsilon$ is some small positive value; 
the \emph{scaling} $\eta(x) \triangleq \frac{x}{\sqrt{d_k}}$; 
and the \emph{softmax} $(\sigma(x))_k \triangleq \frac{e^{x_k}}{\sum\limits_{i \leqslant s} e^{x_i}}$.
Since the scaling and softmax often come together in Transformers, we use $\hat{\sigma}$ to denote the scaled softmax.
\end{example}
\end{figure}

\begin{figure}[H]
\centering
\[\scalebox{1.5}{\input{intro/simplify.tikz}}\]
\begin{example}[The original self-attention]\label{ex:OG}
Having described the transformer in detail above, we may now restrict and abstract to an appropriate vantage point. 
For our purposes, three moves suffice. 
First, we will mainly focus on the self-attention mechanism which is the core component of the Transformer. 
Second, we will only distinguish between two kinds of positive integers, which we notate with colours: those that are \textcolor{cyan}{\emph{fixed}} by the modeller before training, and the length of the \textcolor{orange}{\emph{sequence}} provided at inference-time by the environment. 
Third, without loss of generality (by definition) we consider all processes with learnable parameters to be \textcolor{magenta}{\emph{universal approximators}}. 
This process yields significant simplification relative to Example \ref{ex:vanilla}
\end{example}
\end{figure}

\pagebreak

\section{The evolution of architectures via rewrites}\label{sec:evolution}

\begin{defn}[Expressive Reductions]
Definitionally, a universal approximator of type $A \rightarrow B$ may, depending on its parameters, specialise to become any function of type $A \rightarrow B$. In particular, they may specialise to become arbitrary composites of concrete functions and even other universal approximators. We elaborate on the formal semantics of such \emph{expressivity reductions} in Appendix \ref{appendix:univ}. Below we give an illustration of how these rewrites may be applied. In the case where $f$ is chosen by the rewriter to be equivalent to addition, we have added a residual.
\[\input{expreduction.tikz}\]
\end{defn}

Equipped with diagrams and rewrites between them, we now present derivations for obtaining the Vaswani et al. self-attention mechanism, and the linearised attention mechanism proposed by ~\cite{katharopoulos2020transformers} from their respective predecessor architectures.
These two attention mechanisms incorporate all the ways one can generate and combine query, key, and values, and were thus chosen as the building blocks for our experimental component (Section~\ref{sec:experiments}).
Detailed versions of these derivations are in Appendices~\ref{ssec:appendix_bahd_vas},
\ref{ssec:appendix_generic_sim}, and
\ref{ssec:appendix_vas_to_linear}.

\subsection{Bahdanau et al. to Vaswani et al.}
First, we reformulate the folkloric evolution of \cite{vaswani2017attention} from the attention mechanism introduced in ~\cite{bahdanauNeuralMachineTranslation2016} for an encoder-decoder RNN.
\[\begin{tikzpicture}
	\begin{pgfonlayer}{nodelayer}
		\node [style=smalldotcyan] (206) at (-2.5, -1) {};
		\node [style=none] (216) at (-2, -0.25) {};
		\node [style=none] (218) at (-1.25, -0.25) {};
		\node [style=none] (195) at (-5.5, -1) {};
		\node [style=none] (196) at (-6.5, -1) {};
		\node [style=none] (197) at (-6, -1) {\tiny$\textcolor{orange}{\vert}$};
		\node [style=none] (198) at (-5.5, -0.625) {};
		\node [style=none] (199) at (-5.5, -1.375) {};
		\node [style=none] (200) at (-4.5, -1.375) {};
		\node [style=none] (201) at (-4.5, -0.625) {};
		\node [style=none] (202) at (-5, -1) {\tiny \textcolor{black}{\texttt{enc}}};
		\node [style=smalldotcyan] (203) at (-3.5, -1) {};
		\node [style=none] (204) at (-4.5, -1) {};
		\node [style=none] (205) at (-4, -1) {\tiny$\textcolor{orange}{\vert}$};
		\node [style=none] (207) at (-3.5, -1) {};
		\node [style=none] (208) at (-3, -1) {\tiny$\textcolor{orange}{\vert}$};
		\node [style=none] (209) at (-1.25, 0.25) {};
		\node [style=none] (210) at (-1.25, 0.375) {};
		\node [style=none] (211) at (-1.25, -0.375) {};
		\node [style=none] (212) at (0, -0.375) {};
		\node [style=none] (213) at (0, 0.375) {};
		\node [style=none] (214) at (-0.625, 0) {\tiny \textcolor{black}{\texttt{score}}};
		\node [style=none] (215) at (0, 0) {};
		\node [style=none] (217) at (-2, -0.25) {\tiny$\textcolor{orange}{\vert}$};
		\node [style=smalldotcyan] (219) at (-1.75, 0.25) {};
		\node [style=none] (220) at (0.5, 0) {};
		\node [style=none] (222) at (-1.75, 0.625) {};
		\node [style=none] (223) at (-1.75, -0.625) {};
		\node [style=none] (224) at (0.5, 0.625) {};
		\node [style=none] (225) at (0.5, -0.625) {};
		\node [style=none] (226) at (1, 0) {};
		\node [style=none] (227) at (2.475, 0) {};
		\node [style=tinydotwhite] (234) at (3.1, -1) {\small$\otimes$};
		\node [style=smalldotcyan] (238) at (-2.25, 0.875) {};
		\node [style=none] (249) at (4.225, 1.25) {};
		\node [style=none] (250) at (4.225, 0.5) {};
		\node [style=none] (251) at (4.725, 0.5) {};
		\node [style=none] (252) at (4.725, 1.25) {};
		\node [style=none] (253) at (4.225, 0.875) {};
		\node [style=none] (254) at (4.225, 0.625) {};
		\node [style=none] (255) at (4.725, 0.875) {};
		\node [style=none] (256) at (-3.5, 0.875) {};
		\node [style=none] (257) at (4.225, 1.125) {};
		\node [style=none] (258) at (-3.5, 1.125) {};
		\node [style=none] (259) at (-3.5, 1.5) {};
		\node [style=none] (260) at (6.625, 1.5) {};
		\node [style=none] (261) at (6.625, 0.875) {};
		\node [style=none] (262) at (6.625, 1.125) {};
		\node [style=none] (263) at (4.725, 1.125) {};
		\node [style=none] (264) at (6.625, -1.5) {};
		\node [style=none] (265) at (-3.5, -1.5) {};
		\node [style=smalldotcyan] (266) at (5.375, 1.125) {};
		\node [style=none] (267) at (6.625, -1) {};
		\node [style=none] (268) at (7.625, -1) {};
		\node [style=none] (269) at (7.125, -1) {\tiny$\textcolor{orange}{\vert}$};
		\node [style=none] (270) at (-4, 1.125) {};
		\node [style=none] (271) at (-4, 0.875) {};
		\node [style=none] (272) at (-4.375, 1.2) {};
		\node [style=none] (274) at (-4.25, 1.325) {};
		\node [style=none] (275) at (-4.25, 1.075) {};
		\node [style=none] (276) at (-4, 1.325) {};
		\node [style=none] (277) at (-4, 1.075) {};
		\node [style=none] (278) at (-4.375, 0.825) {};
		\node [style=none] (279) at (-4.25, 0.95) {};
		\node [style=none] (280) at (-4.25, 0.7) {};
		\node [style=none] (281) at (-4, 0.95) {};
		\node [style=none] (282) at (-4, 0.7) {};
		\node [style=none] (283) at (-0.625, -1) {\tiny$\textcolor{orange}{\vert}$};
		\node [style=smalldotcyan] (284) at (7.125, 1.125) {};
		\node [style=smalldotcyan] (285) at (7.125, 0.875) {};
		\node [style=none] (286) at (1.75, 0) {\tiny$\sigma$};
		\node [style=none] (287) at (1.5, 0.25) {};
		\node [style=none] (288) at (2, 0.25) {};
		\node [style=none] (289) at (1.5, -0.25) {};
		\node [style=none] (290) at (2, -0.25) {};
		\node [style=none] (291) at (2, 0) {};
		\node [style=none] (292) at (1.5, 0) {};
		\node [style=none] (293) at (1, 0.5) {};
		\node [style=none] (294) at (2.5, 0.5) {};
		\node [style=none] (295) at (1, -0.5) {};
		\node [style=none] (296) at (2.5, -0.5) {};
	\end{pgfonlayer}
	\begin{pgfonlayer}{edgelayer}
		\draw [style=gray] (215.center) to (220.center);
		\draw [style=cyan] (206)
			 to [in=-180, out=90] (216.center)
			 to (218.center);
		\draw [style=cyan] (219) to (209.center);
		\draw [style=orange] (220.center) to (226.center);
		\draw [style=cyan] (267.center) to (268.center);
		\draw [style=cyan, in=180, out=-90, looseness=0.75] (266) to (267.center);
		\draw [style=cyan] (271.center) to (256.center);
		\draw [style=cyan] (258.center) to (270.center);
		\draw [style=cyan] (261.center) to (285);
		\draw [style=cyan] (262.center) to (284);
		\draw [style=rred] (199.center)
			 to (198.center)
			 to (201.center)
			 to (200.center)
			 to cycle;
		\draw [style=cyan] (196.center) to (195.center);
		\draw [style=cyan] (204.center) to (203);
		\draw [style=cyan] (207.center) to (206);
		\draw [style=rred] (211.center)
			 to (210.center)
			 to (213.center)
			 to (212.center)
			 to cycle;
		\draw [style=orange] (224.center)
			 to (225.center)
			 to (223.center)
			 to (222.center)
			 to cycle;
		\draw [style=orange, in=360, out=90, looseness=1.25] (234) to (227.center);
		\draw [style=cyan, in=-180, out=-90] (238) to (219);
		\draw [style=fr] (249.center)
			 to (252.center)
			 to (251.center)
			 to (250.center)
			 to cycle;
		\draw [style=cyan] (238) to (253.center);
		\draw [style=cyan, in=-180, out=0] (234) to (254.center);
		\draw [style=cyan] (238) to (256.center);
		\draw [style=cyan] (258.center) to (257.center);
		\draw [style=Korange] (256.center)
			 to (259.center)
			 to (260.center)
			 to (261.center);
		\draw [style=cyan] (262.center) to (263.center);
		\draw [style=cyan] (261.center) to (255.center);
		\draw [style=orange] (256.center)
			 to (265.center)
			 to (264.center)
			 to (261.center);
		\draw [style=cyan] (206) to (234);
		\draw (275.center)
			 to (277.center)
			 to (276.center)
			 to (274.center)
			 to (272.center)
			 to cycle;
		\draw (280.center) to (282.center);
		\draw (282.center) to (281.center);
		\draw (281.center) to (279.center);
		\draw (279.center) to (278.center);
		\draw (278.center) to (280.center);
		\draw (289.center)
			 to (287.center)
			 to (288.center)
			 to (290.center)
			 to cycle;
		\draw (292.center) to (226.center);
		\draw (291.center) to (227.center);
		\draw [style=orange] (296.center)
			 to (295.center)
			 to (293.center)
			 to (294.center)
			 to cycle;
	\end{pgfonlayer}
\end{tikzpicture}
\]
The Bahdanau et al. architecture consists of a bi-RNN encoder, with an attention mechanism attached to an RNN decoder. We have abstracted the trainable \texttt{score} function used to calculate the similarity between a query (representing the current decoder output) and a key (representing an encoder output). The original paper uses `additive' attention, which was dropped in favour of a scaled dot product in the Transformer. We also abstract away the bi-RNN as a generic encoder. In fact, we can use reductions to turn the bi-RNN into a SIMD feedforward, which suffices for the rest of our formal derivation.

\[\begin{tikzpicture}
	\begin{pgfonlayer}{nodelayer}
		\node [style=smalldotcyan] (11) at (-2.5, -1) {};
		\node [style=none] (21) at (-2, -0.25) {};
		\node [style=none] (23) at (-1.25, -0.25) {};
		\node [style=none] (0) at (-5.5, -1) {};
		\node [style=none] (1) at (-6.5, -1) {};
		\node [style=none] (2) at (-6, -1) {\tiny$\textcolor{orange}{\vert}$};
		\node [style=none] (3) at (-5.5, -0.625) {};
		\node [style=none] (4) at (-5.5, -1.375) {};
		\node [style=none] (5) at (-4.5, -1.375) {};
		\node [style=none] (6) at (-4.5, -0.625) {};
		\node [style=none] (7) at (-5, -1) {\tiny \textcolor{black}{\texttt{enc}}};
		\node [style=smalldotcyan] (8) at (-3.5, -1) {};
		\node [style=none] (9) at (-4.5, -1) {};
		\node [style=none] (10) at (-4, -1) {\tiny$\textcolor{orange}{\vert}$};
		\node [style=none] (12) at (-3.5, -1) {};
		\node [style=none] (13) at (-3, -1) {\tiny$\textcolor{orange}{\vert}$};
		\node [style=none] (14) at (-1.25, 0.25) {};
		\node [style=none] (15) at (-1.25, 0.375) {};
		\node [style=none] (16) at (-1.25, -0.375) {};
		\node [style=none] (17) at (0, -0.375) {};
		\node [style=none] (18) at (0, 0.375) {};
		\node [style=none] (19) at (-0.625, 0) {\tiny \textcolor{black}{\texttt{score}}};
		\node [style=none] (20) at (0, 0) {};
		\node [style=none] (22) at (-2, -0.25) {\tiny$\textcolor{orange}{\vert}$};
		\node [style=smalldotcyan] (24) at (-1.75, 0.25) {};
		\node [style=none] (25) at (0.5, 0) {};
		\node [style=none] (27) at (-1.75, 0.625) {};
		\node [style=none] (28) at (-1.75, -0.625) {};
		\node [style=none] (29) at (0.5, 0.625) {};
		\node [style=none] (30) at (0.5, -0.625) {};
		\node [style=none] (31) at (1.25, 0) {};
		\node [style=none] (32) at (1.75, 0.025) {};
		\node [style=none] (33) at (1.25, 0.025) {};
		\node [style=tinydotwhite] (39) at (2.225, -0.5) {\small$\otimes$};
		\node [style=none] (41) at (3.1, -0.25) {};
		\node [style=none] (42) at (3.1, -0.75) {};
		\node [style=none] (43) at (3.6, -0.75) {};
		\node [style=none] (44) at (3.6, -0.25) {};
		\node [style=none] (45) at (3.1, -0.5) {};
		\node [style=none] (47) at (3.6, -0.5) {};
		\node [style=none] (48) at (-3.5, 0.25) {};
		\node [style=none] (51) at (-3.5, 1) {};
		\node [style=none] (52) at (4.375, 1) {};
		\node [style=none] (53) at (4.375, -0.5) {};
		\node [style=none] (56) at (4.375, -1.5) {};
		\node [style=none] (57) at (-3.5, -1.5) {};
		\node [style=none] (61) at (4.875, -0.5) {\tiny$\textcolor{orange}{\vert}$};
		\node [style=none] (63) at (-4, 0.25) {};
		\node [style=none] (69) at (-4.375, 0.25) {};
		\node [style=none] (70) at (-4.25, 0.5) {};
		\node [style=none] (71) at (-4.25, 0) {};
		\node [style=none] (72) at (-4, 0.5) {};
		\node [style=none] (73) at (-4, 0) {};
		\node [style=none] (74) at (-0.625, -1) {\tiny$\textcolor{orange}{\vert}$};
		\node [style=none] (75) at (5.375, -0.5) {};
		\node [style=none] (76) at (1.875, -1) {};
		\node [style=none] (77) at (1.25, 0.25) {};
		\node [style=none] (78) at (1.25, -0.25) {};
		\node [style=none] (79) at (1.75, -0.25) {};
		\node [style=none] (80) at (1.75, 0.25) {};
		\node [style=none] (81) at (1.5, 0) {\tiny$\sigma$};
	\end{pgfonlayer}
	\begin{pgfonlayer}{edgelayer}
		\draw [style=cyan] (53.center) to (47.center);
		\draw [style=cyan] (75.center) to (53.center);
		\draw [style=cyan] (1.center) to (0.center);
		\draw [style=cyan] (39) to (45.center);
		\draw [style=cyan] (9.center) to (8);
		\draw [style=gray] (20.center) to (25.center);
		\draw [style=cyan] (11)
			 to [in=-180, out=90] (21.center)
			 to (23.center);
		\draw [style=cyan] (24) to (14.center);
		\draw [style=orange, in=360, out=90, looseness=1.25] (39) to (32.center);
		\draw [style=orange] (25.center) to (31.center);
		\draw [style=rred] (4.center)
			 to (3.center)
			 to (6.center)
			 to (5.center)
			 to cycle;
		\draw [style=cyan] (12.center) to (11);
		\draw [style=rred] (18.center)
			 to (17.center)
			 to (16.center)
			 to (15.center)
			 to cycle;
		\draw [style=orange] (30.center)
			 to (28.center)
			 to (27.center)
			 to (29.center)
			 to cycle;
		\draw [style=fr] (41.center)
			 to (44.center)
			 to (43.center)
			 to (42.center)
			 to cycle;
		\draw [style=orange] (48.center)
			 to (51.center)
			 to (52.center)
			 to (53.center);
		\draw [style=orange] (48.center)
			 to (57.center)
			 to (56.center)
			 to (53.center);
		\draw [style=cyan] (63.center) to (48.center);
		\draw (72.center)
			 to (70.center)
			 to (69.center)
			 to (71.center)
			 to (73.center)
			 to cycle;
		\draw [style=cyan, in=-90, out=0, looseness=1.25] (76.center) to (39);
		\draw [style=cyan] (76.center) to (11);
		\draw [style=cyan] (48.center) to (24);
		\draw (78.center)
			 to (77.center)
			 to (80.center)
			 to (79.center)
			 to cycle;
	\end{pgfonlayer}
\end{tikzpicture}
\]
We rewrite the RNN cell, which discards the recurrent connections.
Note that discarding the recurrent streams means we gain parallelism.
In this intermediate form, we can further: shrink the large SIMD box; split off an invertible learner from the encoder, which may be copied; and specify the trainable score function to be a dot product with scaling, along with trainable `key' and `query' matrices.
\[\begin{tikzpicture}
	\begin{pgfonlayer}{nodelayer}
		\node [style=smalldotcyan] (11) at (0.5, -1.875) {};
		\node [style=none] (63) at (1.25, -2.5) {};
		\node [style=none] (68) at (2.25, -2.5) {};
		\node [style=tinydotwhite] (84) at (4.25, -0.625) {\small$\otimes$};
		\node [style=none] (92) at (2.5, 0) {};
		\node [style=none] (97) at (3.5, 0) {};
		\node [style=none] (79) at (2.5, -1.25) {};
		\node [style=none] (85) at (3.5, -1.25) {};
		\node [style=none] (0) at (-2, -1.875) {};
		\node [style=none] (1) at (-3, -1.875) {};
		\node [style=none] (2) at (-2.5, -1.875) {\tiny$\textcolor{orange}{\vert}$};
		\node [style=none] (3) at (-2, -1.5) {};
		\node [style=none] (4) at (-2, -2.25) {};
		\node [style=none] (5) at (-1, -2.25) {};
		\node [style=none] (6) at (-1, -1.5) {};
		\node [style=none] (7) at (-1.5, -1.875) {\tiny \textcolor{black}{\texttt{enc}}};
		\node [style=none] (9) at (-1, -1.875) {};
		\node [style=none] (10) at (-0.375, -1.875) {\tiny$\textcolor{orange}{\vert}$};
		\node [style=none] (21) at (1, -1.25) {};
		\node [style=none] (53) at (0.25, 0) {};
		\node [style=none] (54) at (-0.125, 0) {};
		\node [style=none] (55) at (0, 0.25) {};
		\node [style=none] (56) at (0, -0.25) {};
		\node [style=none] (57) at (0.25, 0.25) {};
		\node [style=none] (58) at (0.25, -0.25) {};
		\node [style=none] (61) at (9.75, -2.5) {};
		\node [style=none] (64) at (2, -2.25) {};
		\node [style=none] (65) at (2, -2.75) {};
		\node [style=none] (66) at (2.5, -2.75) {};
		\node [style=none] (67) at (2.5, -2.25) {};
		\node [style=none] (69) at (2.5, -2.5) {};
		\node [style=none] (70) at (1.5, -2.125) {};
		\node [style=none] (71) at (1.5, -2.875) {};
		\node [style=none] (72) at (3, -2.125) {};
		\node [style=none] (73) at (3, -2.875) {};
		\node [style=none] (74) at (2, -1) {};
		\node [style=none] (75) at (2, -1.5) {};
		\node [style=none] (76) at (2.5, -1.5) {};
		\node [style=none] (77) at (2.5, -1) {};
		\node [style=none] (78) at (2, -1.25) {};
		\node [style=none] (80) at (1.5, -0.75) {};
		\node [style=none] (81) at (1.5, -1.75) {};
		\node [style=none] (82) at (3, -0.75) {};
		\node [style=none] (83) at (3, -1.75) {};
		\node [style=none] (86) at (0.25, 0) {};
		\node [style=none] (87) at (2, 0.25) {};
		\node [style=none] (88) at (2, -0.25) {};
		\node [style=none] (89) at (2.5, -0.25) {};
		\node [style=none] (90) at (2.5, 0.25) {};
		\node [style=none] (91) at (2, 0) {};
		\node [style=none] (93) at (1.5, 0.5) {};
		\node [style=none] (94) at (1.5, -0.5) {};
		\node [style=none] (95) at (3, 0.5) {};
		\node [style=none] (96) at (3, -0.5) {};
		\node [style=none] (98) at (1, -1.25) {\tiny$\textcolor{orange}{\vert}$};
		\node [style=none] (99) at (1, -2.5) {\tiny$\textcolor{orange}{\vert}$};
		\node [style=none] (100) at (6.25, -0.625) {};
		\node [style=none] (102) at (6.25, -0.6) {};
		\node [style=none] (108) at (5.75, -0.125) {};
		\node [style=none] (109) at (5.75, -1.125) {};
		\node [style=none] (110) at (7.225, -0.125) {};
		\node [style=none] (111) at (7.225, -1.125) {};
		\node [style=none] (112) at (9.75, -0.625) {};
		\node [style=none] (113) at (5.75, -0.625) {};
		\node [style=none] (114) at (6.25, -0.625) {};
		\node [style=none] (115) at (6.75, -0.625) {};
		\node [style=none] (116) at (7.225, -0.625) {};
		\node [style=none] (117) at (5.25, 0.125) {};
		\node [style=none] (118) at (5.25, -1.375) {};
		\node [style=none] (119) at (8.625, 0.125) {};
		\node [style=none] (120) at (8.625, -1.375) {};
		\node [style=none] (121) at (4.75, -0.625) {\tiny$\textcolor{orange}{\vert}$};
		\node [style=none] (122) at (9.25, -0.625) {\tiny$\textcolor{orange}{\vert}$};
		\node [style=tinydotwhite] (123) at (10.5, -1.625) {\small$\otimes$};
		\node [style=none] (124) at (1, 0) {\tiny$\textcolor{orange}{\vert}$};
		\node [style=none] (125) at (12.25, -1.375) {};
		\node [style=none] (126) at (12.25, -1.875) {};
		\node [style=none] (127) at (12.75, -1.875) {};
		\node [style=none] (128) at (12.75, -1.375) {};
		\node [style=none] (129) at (12.25, -1.625) {};
		\node [style=none] (130) at (12.75, -1.625) {};
		\node [style=none] (131) at (11.75, -1.125) {};
		\node [style=none] (132) at (11.75, -2.125) {};
		\node [style=none] (133) at (13.25, -1.125) {};
		\node [style=none] (134) at (13.25, -2.125) {};
		\node [style=none] (135) at (14.25, -1.625) {};
		\node [style=none] (136) at (6.5, -2.5) {\tiny$\textcolor{orange}{\vert}$};
		\node [style=none] (137) at (11.25, -1.625) {\tiny$\textcolor{orange}{\vert}$};
		\node [style=none] (138) at (13.75, -1.625) {\tiny$\textcolor{orange}{\vert}$};
		\node [style=none] (139) at (3.5, 0) {\tiny$\textcolor{orange}{\vert}$};
		\node [style=none] (140) at (3.5, -1.25) {\tiny$\textcolor{orange}{\vert}$};
		\node [style=none] (141) at (6.5, -0.625) {\tiny$\eta$};
		\node [style=none] (142) at (6.25, -0.375) {};
		\node [style=none] (143) at (6.75, -0.375) {};
		\node [style=none] (144) at (6.25, -0.875) {};
		\node [style=none] (145) at (6.75, -0.875) {};
		\node [style=none] (146) at (7.625, -0.375) {};
		\node [style=none] (147) at (7.625, -0.875) {};
		\node [style=none] (148) at (8.125, -0.375) {};
		\node [style=none] (149) at (8.125, -0.875) {};
		\node [style=none] (150) at (7.875, -0.625) {\tiny$\sigma$};
		\node [style=none] (151) at (7.625, -0.625) {};
		\node [style=none] (152) at (8.125, -0.625) {};
	\end{pgfonlayer}
	\begin{pgfonlayer}{edgelayer}
		\draw [style=orange] (116.center) to (151.center);
		\draw [style=gray] (113.center) to (114.center);
		\draw [style=gray] (115.center) to (116.center);
		\draw [style=cyan] (68.center)
			 to (63.center)
			 to [in=-90, out=-180] (11);
		\draw [style=cyan] (91.center) to (86.center);
		\draw [style=cyan] (78.center) to (21.center);
		\draw [style=cyan] (92.center)
			 to (97.center)
			 to [in=90, out=0, looseness=1.25] (84);
		\draw [style=cyan] (79.center)
			 to (85.center)
			 to [in=-90, out=0, looseness=1.25] (84);
		\draw [style=cyan] (123) to (129.center);
		\draw [style=cyan] (130.center) to (135.center);
		\draw [style=rred] (5.center)
			 to (4.center)
			 to (3.center)
			 to (6.center)
			 to cycle;
		\draw [style=cyan] (1.center) to (0.center);
		\draw [style=cyan, in=-180, out=90] (11) to (21.center);
		\draw (56.center) to (58.center);
		\draw (58.center) to (57.center);
		\draw (57.center) to (55.center);
		\draw (55.center) to (54.center);
		\draw (54.center) to (56.center);
		\draw [style=fr] (65.center)
			 to (64.center)
			 to (67.center)
			 to (66.center)
			 to cycle;
		\draw [style=cyan] (123)
			 to [in=0, out=-90] (61.center)
			 to (69.center);
		\draw [style=orange] (73.center)
			 to (71.center)
			 to (70.center)
			 to (72.center)
			 to cycle;
		\draw [style=fr] (76.center)
			 to (75.center)
			 to (74.center)
			 to (77.center)
			 to cycle;
		\draw [style=orange] (81.center)
			 to (80.center)
			 to (82.center)
			 to (83.center)
			 to cycle;
		\draw [style=fr] (89.center)
			 to (88.center)
			 to (87.center)
			 to (90.center)
			 to cycle;
		\draw [style=orange] (94.center)
			 to (93.center)
			 to (95.center)
			 to (96.center)
			 to cycle;
		\draw [style=cyan] (9.center) to (11);
		\draw [style=orange] (111.center)
			 to (109.center)
			 to (108.center)
			 to (110.center)
			 to cycle;
		\draw [style=orange] (123)
			 to [in=0, out=90] (112.center)
			 to (152.center);
		\draw [style=orange] (113.center) to (84);
		\draw [style=orange] (118.center)
			 to (117.center)
			 to (119.center)
			 to (120.center)
			 to cycle;
		\draw [style=fr] (127.center)
			 to (126.center)
			 to (125.center)
			 to (128.center)
			 to cycle;
		\draw [style=orange] (133.center)
			 to (134.center)
			 to (132.center)
			 to (131.center)
			 to cycle;
		\draw (145.center)
			 to (144.center)
			 to (142.center)
			 to (143.center)
			 to cycle;
		\draw (146.center)
			 to (148.center)
			 to (149.center)
			 to (147.center)
			 to cycle;
	\end{pgfonlayer}
\end{tikzpicture}
\]
This is exactly one attention head in the cross attention part of the decoder of the Vaswani et al. Transformer. The decoder outputs, acting as queries, pay attention to the encoder outputs.
Feeding in the encoder input itself as the queries obtains self-attention as per Example~\ref{ex:OG}.

\subsection{Vaswani et al. to linearised attention}
\label{ssec:vas_to_lin}

The linearised variant was developed from the original self-attention.
The derivation of the linear transformation proceeds from the following abstraction:
forgetting that the scaled-dot product self-attention uses a scaled dot product with a softmax, we abstractly represent it using some unspecified \texttt{similarity} function.
\begin{equation}\label{eqn:similarityattention}
	z_i = \sum_j \frac{\texttt{sim}(q_i,k_j)}{\sum_{j'}\texttt{sim}(q_i,k_{j'})}v_j.
\end{equation}
This yields the following diagram (see Appendix~\ref{ssec:appendix_generic_sim} for details):
\begin{align}\label{diag:transformerwithsimilarity}
\begin{tikzpicture}
	\begin{pgfonlayer}{nodelayer}
		\node [style=tinydotwhite] (240) at (7.25, 1.625) {\small$\otimes$};
		\node [style=none] (241) at (6.375, 0.625) {};
		\node [style=none] (248) at (6.875, 0.625) {};
		\node [style=smalldotcyan] (220) at (-4, 0) {};
		\node [style=none] (50) at (-3, 1.25) {};
		\node [style=none] (55) at (-2, 1.25) {};
		\node [style=none] (56) at (-1.5, 1.25) {};
		\node [style=none] (27) at (-3, -1.25) {};
		\node [style=none] (32) at (-2, -1.25) {};
		\node [style=none] (26) at (12, -1.25) {};
		\node [style=none] (33) at (-1.5, -1.25) {};
		\node [style=tinydotwhite] (87) at (12.75, -0.25) {\small$\otimes$};
		\node [style=none] (3) at (2, 1.5) {};
		\node [style=none] (4) at (2, -0.25) {};
		\node [style=none] (5) at (3, -0.25) {};
		\node [style=none] (6) at (3, 1.5) {};
		\node [style=none] (7) at (2.5, 0.625) {\tiny \textcolor{black}{\texttt{sim}}};
		\node [style=none] (11) at (-3, 0) {};
		\node [style=none] (28) at (-2, -1) {};
		\node [style=none] (29) at (-2, -1.5) {};
		\node [style=none] (30) at (-1.5, -1.5) {};
		\node [style=none] (31) at (-1.5, -1) {};
		\node [style=none] (34) at (-2.5, -0.75) {};
		\node [style=none] (35) at (-2.5, -1.75) {};
		\node [style=none] (36) at (-1, -0.75) {};
		\node [style=none] (37) at (-1, -1.75) {};
		\node [style=none] (38) at (-2, 0.25) {};
		\node [style=none] (39) at (-2, -0.25) {};
		\node [style=none] (40) at (-1.5, -0.25) {};
		\node [style=none] (41) at (-1.5, 0.25) {};
		\node [style=none] (42) at (-2, 0) {};
		\node [style=none] (43) at (-1.5, 0) {};
		\node [style=none] (44) at (-2.5, 0.5) {};
		\node [style=none] (45) at (-2.5, -0.5) {};
		\node [style=none] (46) at (-1, 0.5) {};
		\node [style=none] (47) at (-1, -0.5) {};
		\node [style=none] (49) at (-0.5, 0) {};
		\node [style=none] (51) at (-2, 1.5) {};
		\node [style=none] (52) at (-2, 1) {};
		\node [style=none] (53) at (-1.5, 1) {};
		\node [style=none] (54) at (-1.5, 1.5) {};
		\node [style=none] (57) at (-2.5, 1.75) {};
		\node [style=none] (58) at (-2.5, 0.75) {};
		\node [style=none] (59) at (-1, 1.75) {};
		\node [style=none] (60) at (-1, 0.75) {};
		\node [style=none] (61) at (-0.5, 1.25) {};
		\node [style=none] (62) at (-3, 0) {\tiny$\textcolor{orange}{\vert}$};
		\node [style=none] (76) at (12, 0.625) {\tiny$\textcolor{orange}{\vert}$};
		\node [style=none] (81) at (10, 1.875) {};
		\node [style=none] (82) at (10, -0.625) {};
		\node [style=none] (83) at (11.5, 1.875) {};
		\node [style=none] (84) at (11.5, -0.625) {};
		\node [style=none] (88) at (-3, 1.25) {\tiny$\textcolor{orange}{\vert}$};
		\node [style=none] (93) at (13.75, -0.25) {};
		\node [style=none] (100) at (12, -1.25) {\tiny$\textcolor{orange}{\vert}$};
		\node [style=none] (101) at (13.25, -0.25) {\tiny$\textcolor{orange}{\vert}$};
		\node [style=none] (103) at (-0.5, 1.25) {\tiny$\textcolor{orange}{\vert}$};
		\node [style=none] (104) at (-0.5, 0) {\tiny$\textcolor{orange}{\vert}$};
		\node [style=none] (221) at (-3, -1.25) {\tiny$\textcolor{orange}{\vert}$};
		\node [style=none] (223) at (2, 1.25) {};
		\node [style=none] (224) at (2, 0) {};
		\node [style=none] (225) at (1, 1.75) {};
		\node [style=none] (226) at (1, -0.5) {};
		\node [style=none] (227) at (3.5, 1.75) {};
		\node [style=none] (228) at (3.5, -0.5) {};
		\node [style=smalldotcyan] (229) at (1, 1.25) {};
		\node [style=none] (230) at (0, 2) {};
		\node [style=none] (231) at (0, -0.75) {};
		\node [style=none] (232) at (4, 2) {};
		\node [style=none] (233) at (4, -0.75) {};
		\node [style=smalldotcyan] (234) at (0, 0) {};
		\node [style=none] (235) at (3.5, 0.625) {};
		\node [style=none] (236) at (3, 0.625) {};
		\node [style=smalldotorange] (237) at (5.125, 0.625) {};
		\node [style=none] (238) at (-5, 0) {};
		\node [style=none] (239) at (-4.5, 0) {\tiny$\textcolor{orange}{\vert}$};
		\node [style=none] (242) at (6.375, 0.875) {};
		\node [style=none] (243) at (6.125, 0.875) {};
		\node [style=none] (244) at (5.875, 0.625) {};
		\node [style=none] (245) at (6.125, 0.375) {};
		\node [style=none] (246) at (6.375, 0.375) {};
		\node [style=none] (247) at (6.225, 0.625) {\tiny $\mathbf{1}$};
		\node [style=none] (249) at (6, 1.625) {};
		\node [style=none] (250) at (8.125, 1.625) {};
		\node [style=none] (251) at (8.375, 1.625) {};
		\node [style=none] (252) at (8.375, 2) {};
		\node [style=none] (253) at (8.375, 1.25) {};
		\node [style=none] (254) at (9.125, 1.25) {};
		\node [style=none] (255) at (9.125, 2) {};
		\node [style=none] (256) at (8.75, 1.625) {\tiny $\frac{1}{x}$};
		\node [style=none] (257) at (9.125, 1.625) {};
		\node [style=none] (258) at (8.125, 2.25) {};
		\node [style=none] (259) at (9.375, 2.25) {};
		\node [style=none] (260) at (8.125, 1) {};
		\node [style=none] (261) at (9.375, 1) {};
		\node [style=none] (262) at (9.375, 1.625) {};
		\node [style=none] (263) at (10, -0.375) {};
		\node [style=none] (264) at (6, -0.375) {};
		\node [style=none] (265) at (10, 1.625) {};
		\node [style=tinydotwhite] (266) at (10.75, 0.625) {\small$\otimes$};
		\node [style=none] (267) at (4.5, 0.625) {\tiny$\textcolor{orange}{\vert}$};
		\node [style=none] (268) at (7.75, -0.375) {\tiny$\textcolor{orange}{\vert}$};
		\node [style=none] (269) at (6.5, 1.625) {\tiny$\textcolor{orange}{\vert}$};
		\node [style=none] (270) at (0.5, 0) {\tiny$\textcolor{orange}{\vert}$};
	\end{pgfonlayer}
	\begin{pgfonlayer}{edgelayer}
		\draw [style=orange] (240)
			 to [in=0, out=-105, looseness=0.75] (248.center)
			 to (241.center);
		\draw [style=orange] (237) to (235.center);
		\draw [style=gray] (236.center) to (235.center);
		\draw [style=cyan] (224.center) to (104.center);
		\draw [style=cyan] (223.center) to (103.center);
		\draw [style=cyan] (55.center)
			 to (50.center)
			 to [in=90, out=180] (220);
		\draw [style=cyan] (56.center) to (61.center);
		\draw [style=cyan] (42.center) to (11.center);
		\draw [style=cyan] (43.center) to (49.center);
		\draw [style=cyan] (32.center)
			 to (27.center)
			 to [in=-90, out=-180] (220);
		\draw [style=cyan] (87)
			 to [in=0, out=-90] (26.center)
			 to (33.center);
		\draw [style=rred] (5.center)
			 to (4.center)
			 to (3.center)
			 to (6.center)
			 to cycle;
		\draw [style=fr] (29.center)
			 to (28.center)
			 to (31.center)
			 to (30.center)
			 to cycle;
		\draw [style=orange] (35.center)
			 to (34.center)
			 to (36.center)
			 to (37.center)
			 to cycle;
		\draw [style=fr] (40.center)
			 to (39.center)
			 to (38.center)
			 to (41.center)
			 to cycle;
		\draw [style=orange] (44.center)
			 to (46.center)
			 to (47.center)
			 to (45.center)
			 to cycle;
		\draw [style=fr] (53.center)
			 to (52.center)
			 to (51.center)
			 to (54.center)
			 to cycle;
		\draw [style=orange] (57.center)
			 to (59.center)
			 to (60.center)
			 to (58.center)
			 to cycle;
		\draw [style=orange, in=0, out=90] (87) to (76.center);
		\draw [style=orange] (81.center)
			 to (83.center)
			 to (84.center)
			 to (82.center)
			 to cycle;
		\draw [style=cyan] (87) to (93.center);
		\draw [style=cyan] (62.center) to (220);
		\draw [style=orange] (225.center)
			 to (227.center)
			 to (228.center)
			 to (226.center)
			 to cycle;
		\draw [style=orange] (233.center)
			 to (231.center)
			 to (230.center)
			 to (232.center)
			 to cycle;
		\draw [style=cyan] (238.center) to (220);
		\draw (246.center) to (242.center);
		\draw (242.center) to (243.center);
		\draw (243.center) to (244.center);
		\draw (244.center) to (245.center);
		\draw (245.center) to (246.center);
		\draw [style=orange] (240)
			 to (249.center)
			 to [in=90, out=-180] (237);
		\draw [style=orange] (250.center) to (240);
		\draw (255.center)
			 to (252.center)
			 to (253.center)
			 to (254.center)
			 to cycle;
		\draw [style=orange] (258.center)
			 to (259.center)
			 to (261.center)
			 to (260.center)
			 to cycle;
		\draw [style=gray] (250.center) to (251.center);
		\draw [style=gray] (262.center) to (257.center);
		\draw [style=orange] (266)
			 to [in=0, out=-90, looseness=1.25] (263.center)
			 to (264.center)
			 to [in=-90, out=-180] (237);
		\draw [style=orange] (265.center) to (262.center);
		\draw [style=gray, in=90, out=0, looseness=1.25] (265.center) to (266);
		\draw [style=orange] (76.center) to (266);
	\end{pgfonlayer}
\end{tikzpicture}

\end{align}

We have made explicit here the process for calculating the normalisation of the attention matrix -- the $\mathbf{1}$ in the diagram is a column vector of 1's, which sums the rows of the attention matrix.
To recover Vaswani et al., one specifies $\texttt{sim}(x,y)=\exp\left(\frac{x\cdot y}{\sqrt{d_k}}\right)$.

Observe that there are three matrix multiplication operations -- with respect to the sequence length $s$, the first and last of these have $O(s^2)$ time complexity. 
The middle one is $O(s)$, but is in an $s$-SIMD, so overall the operation is also $O(s^2)$.
Observe also there are wires carrying $[s,s]$ sized tensors -- so this model is also $O(s^2)$ in space complexity. To linearize this, one first chooses the similarity function $\texttt{sim}(x,y)$ to be $\phi(x)\cdot\phi(y)$, where $\phi:\mathbb{R}^d\to \mathbb{R}^{d'}$ gives a kernel decomposition.
Diagrammatically, this yields:
\begin{align}\label{diag:linear_init}
    \begin{tikzpicture}
	\begin{pgfonlayer}{nodelayer}
		\node [style=tinydotwhite] (73) at (3.5, 1.375) {\small$\otimes$};
		\node [style=none] (74) at (2.375, 0.5) {};
		\node [style=none] (81) at (3.125, 0.5) {};
		\node [style=none] (5) at (-7.5, -0.25) {};
		\node [style=none] (6) at (8.5, -1.5) {};
		\node [style=none] (7) at (-7.5, -1.5) {};
		\node [style=none] (8) at (-6.5, -1.25) {};
		\node [style=none] (9) at (-6.5, -1.75) {};
		\node [style=none] (10) at (-6, -1.75) {};
		\node [style=none] (11) at (-6, -1.25) {};
		\node [style=none] (12) at (-6.5, -1.5) {};
		\node [style=none] (13) at (-6, -1.5) {};
		\node [style=none] (14) at (-7, -1) {};
		\node [style=none] (15) at (-7, -2) {};
		\node [style=none] (16) at (-5.5, -1) {};
		\node [style=none] (17) at (-5.5, -2) {};
		\node [style=none] (18) at (-6.5, 0) {};
		\node [style=none] (19) at (-6.5, -0.5) {};
		\node [style=none] (20) at (-6, -0.5) {};
		\node [style=none] (21) at (-6, 0) {};
		\node [style=none] (22) at (-6.5, -0.25) {};
		\node [style=none] (23) at (-6, -0.25) {};
		\node [style=none] (24) at (-7, 0.25) {};
		\node [style=none] (25) at (-7, -0.75) {};
		\node [style=none] (26) at (-5.5, 0.25) {};
		\node [style=none] (27) at (-5.5, -0.75) {};
		\node [style=none] (28) at (-5, -0.25) {};
		\node [style=none] (29) at (-7.5, 1) {};
		\node [style=none] (30) at (-6.5, 1.25) {};
		\node [style=none] (31) at (-6.5, 0.75) {};
		\node [style=none] (32) at (-6, 0.75) {};
		\node [style=none] (33) at (-6, 1.25) {};
		\node [style=none] (34) at (-6.5, 1) {};
		\node [style=none] (35) at (-6, 1) {};
		\node [style=none] (36) at (-7, 1.5) {};
		\node [style=none] (37) at (-7, 0.5) {};
		\node [style=none] (38) at (-5.5, 1.5) {};
		\node [style=none] (39) at (-5.5, 0.5) {};
		\node [style=none] (40) at (-5, 1) {};
		\node [style=none] (41) at (-7.5, -0.25) {\tiny$\textcolor{orange}{\vert}$};
		\node [style=none] (42) at (8.5, 0.5) {\tiny$\textcolor{orange}{\vert}$};
		\node [style=none] (43) at (6.75, 1.625) {};
		\node [style=none] (44) at (6.75, -0.75) {};
		\node [style=none] (45) at (8, 1.625) {};
		\node [style=none] (46) at (8, -0.75) {};
		\node [style=tinydotwhite] (47) at (9.25, -0.5) {\small$\otimes$};
		\node [style=none] (48) at (-7.5, 1) {\tiny$\textcolor{orange}{\vert}$};
		\node [style=none] (49) at (10.25, -0.5) {};
		\node [style=none] (50) at (8.5, -1.5) {\tiny$\textcolor{orange}{\vert}$};
		\node [style=none] (51) at (9.75, -0.5) {\tiny$\textcolor{orange}{\vert}$};
		\node [style=none] (52) at (-5, 1) {\tiny$\textcolor{orange}{\vert}$};
		\node [style=none] (53) at (-5, -0.25) {\tiny$\textcolor{orange}{\vert}$};
		\node [style=smalldotcyan] (54) at (-8.5, -0.25) {};
		\node [style=none] (55) at (-7.5, -1.5) {\tiny$\textcolor{orange}{\vert}$};
		\node [style=none] (56) at (-2.5, 1) {};
		\node [style=none] (57) at (-2.5, -0.25) {};
		\node [style=none] (58) at (-3.5, 1.5) {};
		\node [style=none] (59) at (-3.5, -0.75) {};
		\node [style=none] (60) at (-0.5, 1.5) {};
		\node [style=none] (61) at (-0.5, -0.75) {};
		\node [style=smalldotcyan] (62) at (-3.5, 1) {};
		\node [style=none] (63) at (-4.5, 1.75) {};
		\node [style=none] (64) at (-4.5, -1) {};
		\node [style=none] (65) at (0, 1.75) {};
		\node [style=none] (66) at (0, -1) {};
		\node [style=smalldotcyan] (67) at (-4.5, -0.25) {};
		\node [style=none] (68) at (-0.5, 0.375) {};
		\node [style=smalldotorange] (70) at (1.125, 0.375) {};
		\node [style=none] (71) at (-9.5, -0.25) {};
		\node [style=none] (72) at (-9, -0.25) {\tiny$\textcolor{orange}{\vert}$};
		\node [style=none] (75) at (2.375, 0.75) {};
		\node [style=none] (76) at (2.125, 0.75) {};
		\node [style=none] (77) at (1.875, 0.5) {};
		\node [style=none] (78) at (2.125, 0.25) {};
		\node [style=none] (79) at (2.375, 0.25) {};
		\node [style=none] (80) at (2.225, 0.5) {\tiny $\mathbf{1}$};
		\node [style=none] (82) at (2, 1.375) {};
		\node [style=none] (83) at (4.5, 1.375) {};
		\node [style=none] (84) at (5, 1.375) {};
		\node [style=none] (85) at (5, 1.75) {};
		\node [style=none] (86) at (5, 1) {};
		\node [style=none] (87) at (5.75, 1) {};
		\node [style=none] (88) at (5.75, 1.75) {};
		\node [style=none] (89) at (5.375, 1.375) {\tiny $\frac{1}{x}$};
		\node [style=none] (90) at (5.75, 1.375) {};
		\node [style=none] (91) at (4.5, 2) {};
		\node [style=none] (92) at (6.25, 2) {};
		\node [style=none] (93) at (4.5, 0.75) {};
		\node [style=none] (94) at (6.25, 0.75) {};
		\node [style=none] (95) at (6.25, 1.375) {};
		\node [style=none] (96) at (6.75, -0.5) {};
		\node [style=none] (97) at (2, -0.5) {};
		\node [style=none] (98) at (6.75, 1.375) {};
		\node [style=tinydotwhite] (99) at (7.5, 0.5) {\small$\otimes$};
		\node [style=none] (100) at (0.5, 0.375) {\tiny$\textcolor{orange}{\vert}$};
		\node [style=none] (101) at (4.125, -0.5) {\tiny$\textcolor{orange}{\vert}$};
		\node [style=none] (102) at (2.75, 1.375) {\tiny$\textcolor{orange}{\vert}$};
		\node [style=none] (103) at (-2.25, 1) {\tiny $\textcolor{black}{\phi}$};
		\node [style=none] (104) at (-2.5, 1.25) {};
		\node [style=none] (105) at (-2.5, 0.75) {};
		\node [style=none] (106) at (-2, 0.75) {};
		\node [style=none] (107) at (-2, 1.25) {};
		\node [style=none] (112) at (-2, 1) {};
		\node [style=none] (114) at (-2.25, -0.25) {\tiny $\textcolor{black}{\phi}$};
		\node [style=none] (115) at (-2.5, 0) {};
		\node [style=none] (116) at (-2.5, -0.5) {};
		\node [style=none] (117) at (-2, -0.5) {};
		\node [style=none] (118) at (-2, 0) {};
		\node [style=none] (123) at (-2, -0.25) {};
		\node [style=tinydotwhite] (124) at (-1.25, 0.375) {\small$\otimes$};
		\node [style=none] (125) at (-4, -0.25) {\tiny$\textcolor{orange}{\vert}$};
	\end{pgfonlayer}
	\begin{pgfonlayer}{edgelayer}
		\draw [style=orange] (73)
			 to [in=0, out=-105, looseness=0.75] (81.center)
			 to [in=360, out=180] (74.center);
		\draw [style=cyan] (12.center)
			 to (7.center)
			 to [in=-90, out=-180] (54);
		\draw [style=fr] (11.center)
			 to (10.center)
			 to (9.center)
			 to (8.center)
			 to cycle;
		\draw [style=cyan, in=0, out=-90] (47) to (6.center);
		\draw [style=cyan] (6.center) to (13.center);
		\draw [style=orange] (15.center)
			 to (14.center)
			 to (16.center)
			 to (17.center)
			 to cycle;
		\draw [style=fr] (19.center)
			 to (18.center)
			 to (21.center)
			 to (20.center)
			 to cycle;
		\draw [style=orange] (26.center)
			 to (27.center)
			 to (25.center)
			 to (24.center)
			 to cycle;
		\draw [style=cyan] (22.center) to (5.center);
		\draw [style=cyan] (23.center) to (28.center);
		\draw [style=fr] (30.center)
			 to (33.center)
			 to (32.center)
			 to (31.center)
			 to cycle;
		\draw [style=orange] (39.center)
			 to (37.center)
			 to (36.center)
			 to (38.center)
			 to cycle;
		\draw [style=cyan] (34.center)
			 to (29.center)
			 to [in=90, out=-180] (54);
		\draw [style=cyan] (35.center) to (40.center);
		\draw [style=orange, in=0, out=90] (47) to (42.center);
		\draw [style=orange] (46.center)
			 to (44.center)
			 to (43.center)
			 to (45.center)
			 to cycle;
		\draw [style=cyan] (47) to (49.center);
		\draw [style=cyan] (41.center) to (54);
		\draw [style=cyan] (56.center) to (52.center);
		\draw [style=cyan] (57.center) to (53.center);
		\draw [style=orange] (61.center)
			 to (59.center)
			 to (58.center)
			 to (60.center)
			 to cycle;
		\draw [style=orange] (63.center)
			 to (65.center)
			 to (66.center)
			 to (64.center)
			 to cycle;
		\draw [style=orange] (70) to (68.center);
		\draw [style=cyan] (71.center) to (54);
		\draw (79.center) to (75.center);
		\draw (75.center) to (76.center);
		\draw (76.center) to (77.center);
		\draw (77.center) to (78.center);
		\draw (78.center) to (79.center);
		\draw [style=orange] (73)
			 to (82.center)
			 to [in=90, out=-180] (70);
		\draw [style=orange] (83.center) to (73);
		\draw (86.center)
			 to (87.center)
			 to (88.center)
			 to (85.center)
			 to cycle;
		\draw [style=orange] (94.center)
			 to (93.center)
			 to (91.center)
			 to (92.center)
			 to cycle;
		\draw [style=gray] (83.center) to (84.center);
		\draw [style=gray] (95.center) to (90.center);
		\draw [style=orange] (99)
			 to [in=0, out=-90, looseness=1.25] (96.center)
			 to (97.center)
			 to [in=-90, out=-180] (70);
		\draw [style=orange] (98.center) to (95.center);
		\draw [style=gray, in=90, out=0, looseness=1.25] (98.center) to (99);
		\draw [style=orange] (42.center) to (99);
		\draw [style=rred] (107.center)
			 to (106.center)
			 to (105.center)
			 to (104.center)
			 to cycle;
		\draw [style=rred] (115.center)
			 to (118.center)
			 to (117.center)
			 to (116.center)
			 to cycle;
		\draw [style=gray] (68.center) to (124);
		\draw [style=cyan, in=0, out=90, looseness=1.25] (124) to (112.center);
		\draw [style=cyan, in=0, out=-90, looseness=1.25] (124) to (123.center);
	\end{pgfonlayer}
\end{tikzpicture}

\end{align}
Then, by unfolding the SIMD boxes in the resulting diagram, and applying three instances of the associativity of matrix multiplication, we arrive at the following diagram, corresponding to the computational graph of the linear transformer:
\begin{align}\label{diag:lineartransformer}
    \begin{tikzpicture}
	\begin{pgfonlayer}{nodelayer}
		\node [style=none] (0) at (1.5, 0.75) {};
		\node [style=tinydotwhite] (1) at (2, 0) {\tiny$\otimes$};
		\node [style=none] (2) at (2, 0.25) {};
		\node [style=none] (3) at (1, 0.75) {};
		\node [style=none] (4) at (-3, 0) {};
		\node [style=none] (5) at (-3, 1.25) {};
		\node [style=none] (6) at (-3, -1.25) {};
		\node [style=tinydotwhite] (8) at (1.25, -1.25) {\tiny$\otimes$};
		\node [style=none] (9) at (-1.75, -1.25) {\tiny$\textcolor{orange}{\vert}$};
		\node [style=smalldotcyan] (10) at (-5.25, 0) {};
		\node [style=none] (11) at (-3.25, 0) {};
		\node [style=none] (12) at (-4.25, 0) {};
		\node [style=none] (13) at (-3, 0) {};
		\node [style=none] (14) at (-3, 1.25) {};
		\node [style=none] (15) at (-3, -1.25) {};
		\node [style=none] (16) at (-3.25, -1.25) {};
		\node [style=none] (17) at (-4.25, -1.25) {};
		\node [style=none] (18) at (-3, 0) {};
		\node [style=none] (19) at (-3, 1.25) {};
		\node [style=none] (20) at (-2.75, -1.25) {};
		\node [style=none] (21) at (-3.25, 1.5) {};
		\node [style=none] (22) at (-3.25, 1) {};
		\node [style=none] (23) at (-2.75, 1) {};
		\node [style=none] (24) at (-2.75, 1.5) {};
		\node [style=none] (25) at (-3.25, 1.25) {};
		\node [style=none] (26) at (-3.25, 0.25) {};
		\node [style=none] (27) at (-3.25, -0.25) {};
		\node [style=none] (28) at (-2.75, -0.25) {};
		\node [style=none] (29) at (-2.75, 0.25) {};
		\node [style=none] (30) at (-3.25, -1) {};
		\node [style=none] (31) at (-3.25, -1.5) {};
		\node [style=none] (32) at (-2.75, -1.5) {};
		\node [style=none] (33) at (-2.75, -1) {};
		\node [style=none] (34) at (-4.25, 1.25) {};
		\node [style=none] (35) at (-4.25, 1.25) {\tiny$\textcolor{orange}{\vert}$};
		\node [style=none] (36) at (-4.25, 0) {\tiny$\textcolor{orange}{\vert}$};
		\node [style=none] (37) at (-4.25, -1.25) {\tiny$\textcolor{orange}{\vert}$};
		\node [style=none] (38) at (-6.25, 0) {};
		\node [style=none] (39) at (-5.75, 0) {\tiny$\textcolor{orange}{\vert}$};
		\node [style=none] (40) at (-2, 1.25) {\tiny $\textcolor{black}{\phi}$};
		\node [style=none] (41) at (-2.25, 1.5) {};
		\node [style=none] (42) at (-2.25, 1) {};
		\node [style=none] (43) at (-1.75, 1) {};
		\node [style=none] (44) at (-1.75, 1.5) {};
		\node [style=none] (45) at (-3.75, 1.75) {};
		\node [style=none] (46) at (-1.25, 1.75) {};
		\node [style=none] (47) at (-3.75, 0.75) {};
		\node [style=none] (48) at (-1.25, 0.75) {};
		\node [style=none] (49) at (-1.75, 1.25) {};
		\node [style=none] (50) at (-2.25, 1.25) {};
		\node [style=none] (51) at (-2, 0) {\tiny $\textcolor{black}{\phi}$};
		\node [style=none] (52) at (-2.25, 0.25) {};
		\node [style=none] (53) at (-2.25, -0.25) {};
		\node [style=none] (54) at (-1.75, -0.25) {};
		\node [style=none] (55) at (-1.75, 0.25) {};
		\node [style=none] (56) at (-3.75, 0.5) {};
		\node [style=none] (57) at (-1.25, 0.5) {};
		\node [style=none] (58) at (-3.75, -0.5) {};
		\node [style=none] (59) at (-1.25, -0.5) {};
		\node [style=none] (60) at (-1.75, 0) {};
		\node [style=none] (61) at (-2.25, 0) {};
		\node [style=none] (63) at (-1.25, 1.25) {};
		\node [style=smalldotcyan] (64) at (-0.25, 1.25) {};
		\node [style=smalldotcyan] (65) at (-0.25, 0) {};
		\node [style=none] (66) at (-0.75, 1.25) {\tiny$\textcolor{orange}{\vert}$};
		\node [style=none] (67) at (-0.75, 0) {\tiny$\textcolor{orange}{\vert}$};
		\node [style=tinydotwhite] (68) at (0.25, -1.25) {\tiny$\otimes$};
		\node [style=none] (69) at (-0.25, 1.25) {};
		\node [style=none] (70) at (-0.25, 0.75) {};
		\node [style=none] (71) at (0.25, 0.25) {};
		\node [style=none] (72) at (-0.25, -0.25) {};
		\node [style=none] (73) at (0, -0.5) {};
		\node [style=none] (74) at (0.25, -1) {};
		\node [style=none] (75) at (1.25, -0.25) {};
		\node [style=none] (76) at (0.75, 0.25) {};
		\node [style=none] (77) at (3.5, 1.25) {};
		\node [style=none] (78) at (1, 1) {};
		\node [style=none] (79) at (0.75, 1) {};
		\node [style=none] (80) at (0.5, 0.75) {};
		\node [style=none] (81) at (0.75, 0.5) {};
		\node [style=none] (82) at (1, 0.5) {};
		\node [style=none] (83) at (0.85, 0.75) {\tiny $\mathbf{1}$};
		\node [style=tinydotwhite] (85) at (2.5, 1.25) {\tiny$\otimes$};
		\node [style=none] (86) at (2, 0) {};
		\node [style=none] (87) at (2.5, 0.5) {};
		\node [style=none] (89) at (4, 1.625) {};
		\node [style=none] (90) at (4, 0.875) {};
		\node [style=none] (91) at (4.75, 0.875) {};
		\node [style=none] (92) at (4.75, 1.625) {};
		\node [style=none] (93) at (4.375, 1.25) {\tiny $\frac{1}{x}$};
		\node [style=none] (94) at (5.25, 1.25) {};
		\node [style=none] (95) at (3.025, 1.25) {\tiny$\textcolor{orange}{\vert}$};
		\node [style=none] (96) at (3.5, 1.75) {};
		\node [style=none] (97) at (5.25, 1.75) {};
		\node [style=none] (98) at (3.5, 0.75) {};
		\node [style=none] (99) at (5.25, 0.75) {};
		\node [style=none] (100) at (5.75, 1.25) {};
		\node [style=none] (101) at (0.25, 1.25) {\tiny$\textcolor{orange}{\vert}$};
		\node [style=none] (102) at (2.75, -1.25) {\tiny$\textcolor{orange}{\vert}$};
		\node [style=none] (103) at (5.75, -1.25) {};
		\node [style=none] (105) at (0.75, -1.25) {\tiny$\textcolor{cyan}{\vert}$};
		\node [style=none] (106) at (5.75, 1.75) {};
		\node [style=none] (107) at (5.75, -1.75) {};
		\node [style=none] (108) at (6.75, 1.75) {};
		\node [style=none] (109) at (6.75, -1.75) {};
		\node [style=none] (110) at (6.25, 0.75) {};
		\node [style=none] (111) at (6.25, -0.75) {};
		\node [style=tinydotwhite] (112) at (6.25, 0) {\tiny$\otimes$};
		\node [style=none] (113) at (7.75, 0) {};
		\node [style=none] (114) at (7.25, 0) {\tiny$\textcolor{orange}{\vert}$};
		\node [style=none] (115) at (0.25, 0) {\tiny$\textcolor{orange}{\vert}$};
		\node [style=none] (116) at (4, 1.25) {};
		\node [style=none] (117) at (4.75, 1.25) {};
		\node [style=none] (118) at (-3.75, -0.75) {};
		\node [style=none] (119) at (-2.25, -0.75) {};
		\node [style=none] (120) at (-3.75, -1.75) {};
		\node [style=none] (121) at (-2.25, -1.75) {};
	\end{pgfonlayer}
	\begin{pgfonlayer}{edgelayer}
		\draw [style=cyan] (68) to (20.center);
		\draw [style=orange] (3.center)
			 to (0.center)
			 to [in=90, out=0] (2.center)
			 to (1);
		\draw [style=cyan] (112) to (113.center);
		\draw [style=cyan] (103.center)
			 to [in=270, out=0] (111.center)
			 to (112);
		\draw [style=gray] (100.center)
			 to [in=90, out=0] (110.center)
			 to (112);
		\draw [style=orange] (94.center) to (100.center);
		\draw [style=orange] (85) to (77.center);
		\draw [style=cyan] (50.center) to (25.center);
		\draw [style=cyan] (60.center) to (65);
		\draw [style=cyan] (61.center) to (18.center);
		\draw [style=cyan] (49.center) to (64);
		\draw [style=cyan] (10) to (12.center);
		\draw [style=cyan] (12.center) to (11.center);
		\draw [style=cyan, in=-180, out=90] (10) to (34.center);
		\draw [style=cyan] (34.center) to (25.center);
		\draw [style=cyan, in=-180, out=-90] (10) to (17.center);
		\draw [style=cyan] (17.center) to (16.center);
		\draw [style=fr] (24.center)
			 to (23.center)
			 to (22.center)
			 to (21.center)
			 to cycle;
		\draw [style=fr] (29.center)
			 to (28.center)
			 to (27.center)
			 to (26.center)
			 to cycle;
		\draw [style=fr] (32.center)
			 to (31.center)
			 to (30.center)
			 to (33.center)
			 to cycle;
		\draw [style=cyan] (38.center) to (36.center);
		\draw [style=rred] (44.center)
			 to (43.center)
			 to (42.center)
			 to (41.center)
			 to cycle;
		\draw [style=orange] (48.center)
			 to (47.center)
			 to (45.center)
			 to (46.center)
			 to cycle;
		\draw [style=rred] (52.center)
			 to (55.center)
			 to (54.center)
			 to (53.center)
			 to cycle;
		\draw [style=orange] (56.center)
			 to (57.center)
			 to (59.center)
			 to (58.center)
			 to cycle;
		\draw [style=cyan] (64) to (69.center);
		\draw [style=cyan] (65)
			 to (72.center)
			 to [in=-180, out=-90] (73.center)
			 to [in=90, out=0, looseness=1.25] (74.center)
			 to (68);
		\draw [style=cyan] (64)
			 to (69.center)
			 to (70.center)
			 to [in=-180, out=-90] (71.center)
			 to (76.center)
			 to [in=90, out=0] (75.center)
			 to (8);
		\draw (82.center)
			 to (78.center)
			 to (79.center)
			 to (80.center)
			 to (81.center)
			 to cycle;
		\draw [style=cyan] (65) to (1);
		\draw [style=cyan] (1)
			 to (86.center)
			 to [in=270, out=0] (87.center)
			 to (85);
		\draw [style=cyan] (69.center) to (85);
		\draw (89.center)
			 to (90.center)
			 to (91.center)
			 to (92.center)
			 to cycle;
		\draw [style=orange] (97.center)
			 to (99.center)
			 to (98.center)
			 to (96.center)
			 to cycle;
		\draw [style=cyan] (8) to (103.center);
		\draw [style=cyan] (68) to (8);
		\draw [style=orange] (107.center)
			 to (106.center)
			 to (108.center)
			 to (109.center)
			 to cycle;
		\draw [style=gray] (77.center) to (116.center);
		\draw [style=gray] (117.center) to (94.center);
		\draw [style=orange] (119.center)
			 to (121.center)
			 to (120.center)
			 to (118.center)
			 to cycle;
	\end{pgfonlayer}
\end{tikzpicture}

\end{align}
Since the rewrites we used were mathematical identities, this diagram computes the same function as Diagram~\ref{diag:linear_init}.
However, its computational properties are different -- we can easily see that all instances of matrix multiplication are $O(s)$, and similarly the wires only carry $O(s)$ sized tensors. 

\pagebreak
\section{Taxonomizing architectures}\label{sec:tax}
\begin{figure}[p]
    \vspace{-2cm}
    \centering
    \makebox[\textwidth][c]{\scalebox{1.3}{$$\input{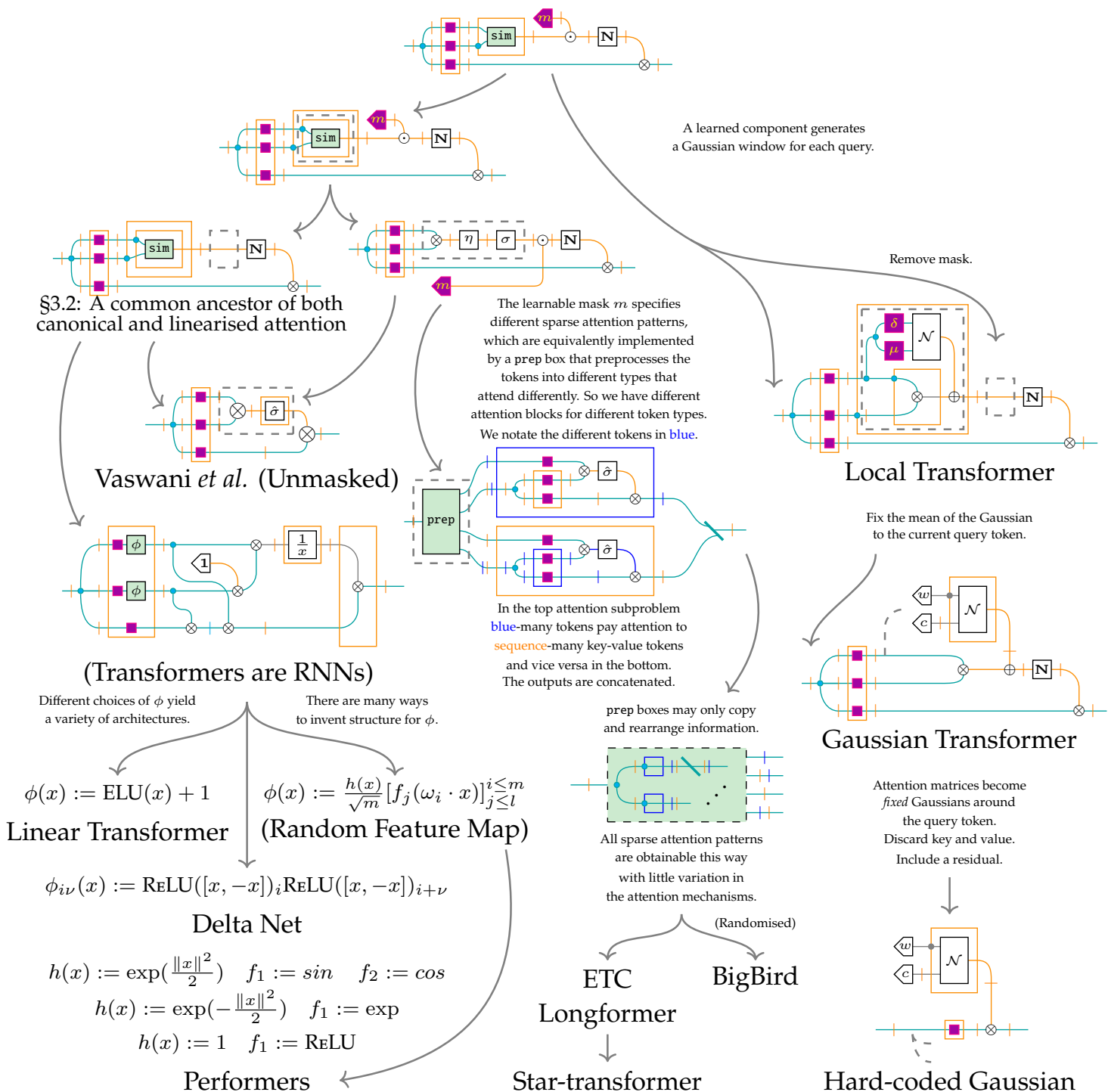}$$}}
    \caption{The taxonomy generated by starting with a `primordial attention' mechanism and applying specialisations and expressivity rewrites. Details of the notation and rewrites are given in Section \ref{sec:tax}. \label{fig:bigtax}}
\end{figure}

In Figure \ref{fig:bigtax}, we provide a ``primordial attention'' from which we derive a variety of transformers by expressivity reductions.
Here we depict three broad families, each corresponding to structural variations on a particular part of the primordial attention mechanism. On the left are the ``Linear Transformers'' which vary in the abstract \texttt{similarity} computation performed on keys and queries. In the middle are ``sparse attentions'' that vary up to the mask, which is equivalent to variations in an abstract \texttt{preparation} procedure that copies and reshapes incoming data.
On the right are ``Gaussian attention'' variants where the attention pattern is scaled by a Gaussian window.

Highlighted boxes indicate abstract families of functions to be instantiated by the user. For example, the \texttt{sim} box denotes an abstract ``similarity" to be computed. $\mathbf{N}$ denotes normalisation of the attention matrix.
Recall $\eta$ is the scaling, $\sigma$ is the softmax, and the composite $\hat{\sigma}$ is the scaled softmax. $\mathcal{N}$ generates a Gaussian window vector parameterised by the mean and standard deviation.
Grey arrows indicate expressivity rewrites, and dashed boxes indicate how child architectures relate to their parents. 

\section{Empirically exploring a space of attention mechanisms}
\label{sec:experiments}

As an empirical first test of the utility of our diagrams, in this section we define, explore, and exhaustively test a space of attention mechanisms. We select two generators from the transformer: \texttt{AttPrep} and \texttt{AttApply}, responsible for preparing the attention matrix, and applying it to the values matrix, respectively. 
We also select the analogous generators from the linear transformer \cite{katharopoulos_transformers_2020}: \texttt{LinAttPrep} and \texttt{LinAttApply}, where $\phi(x) := \text{ELU}(x)+1$.
In addition to these attention-related components, we require additional generators to allow for non-attentional information interactions. 
We introduce a \texttt{ConcatFF} generator, which allows for the merging of two data streams (of the same shape as the original input).

\begin{center}
    \begin{tabular}{ccccc}
        \texttt{ConcatFF} & \texttt{AttPrep} & \texttt{AttApply} & \texttt{LinAttPrep} & \texttt{LinAttApply} \\[0.5em]
        $\input{generators/concatFF.tikz}$ & $\input{generators/AttPrep.tikz}$ & $\input{generators/AttApply.tikz}$ & $\input{generators/LinAttPrep.tikz}$ & $\input{generators/LinAttApply.tikz}$
    \end{tabular}
\end{center}

We consider all attention mechanisms composed of at most five generators. Each diagram must have a single sequence length wire as input and output. Arbitrarily many copies of the input sequence are allowed, as long as all are consumed by the remaining generators in the diagrams. Even at just five generators, this enumeration yields many potential attention heads, which is a difficult space to evaluate numerically. 
To reduce the search space, we employ the following reductions, motivated by our diagrammatic rewrite system. First, we remove all diagrams which are sequential composites of simpler diagrams. This limits sequential composition to layers with residuals, as in the original transformer architecture. Second, for diagrams which are equivalent up to local applications of rewrites \eqref{eq:rewrite1} and \eqref{eq:rewrite2}, we remove the one with more generators, which eliminates trivial overparameterisations. This reduces the number to 14 distinct attention mechanisms, which we display in Figure \ref{fig:attmechs}. See Appendix \ref{app:lemmaproofs} for proofs of \eqref{eq:rewrite1} and \eqref{eq:rewrite2}.
\begin{align}
    \input{greenequivs/copmon.tikz} \quad &\equiv \quad \input{greenequivs/proc.tikz} \label{eq:rewrite1}\\
    \input{greenequivs/bmon.tikz} \quad &\equiv \quad \input{greenequivs/bom.tikz} \label{eq:rewrite2}
\end{align}

Given the generated attention mechanisms, we constructed decoder-only Transformer-style models from each -- the construction is the same as the decoder component of \cite{vaswani2017attention}, but replacing the scaled-dot-product attention heads with each attention variant (all other parts, e.g. residuals and feedforward networks, are kept the same). In line with \cite{qin2022devil}, we observed that variants containing \texttt{LinAttPrep} and \texttt{LinAttApply} were unstable during training -- therefore, as suggested there, we added additional layer normalisation after each \texttt{LinAttApply} or \texttt{AttApply} generator. For all models, we used the causally-masked variant of \texttt{LinAttPrep}/\texttt{LinAttApply} given by \cite{katharopoulos2020transformers}. An implementation of these models is provided in the supplementary material, and see Appendix \ref{app:experiment} for more details.

\begin{figure}[p]
    \centering
    \textbf{(a)}\\
    \makebox[5.1in][l]{\includegraphics[height=3in]{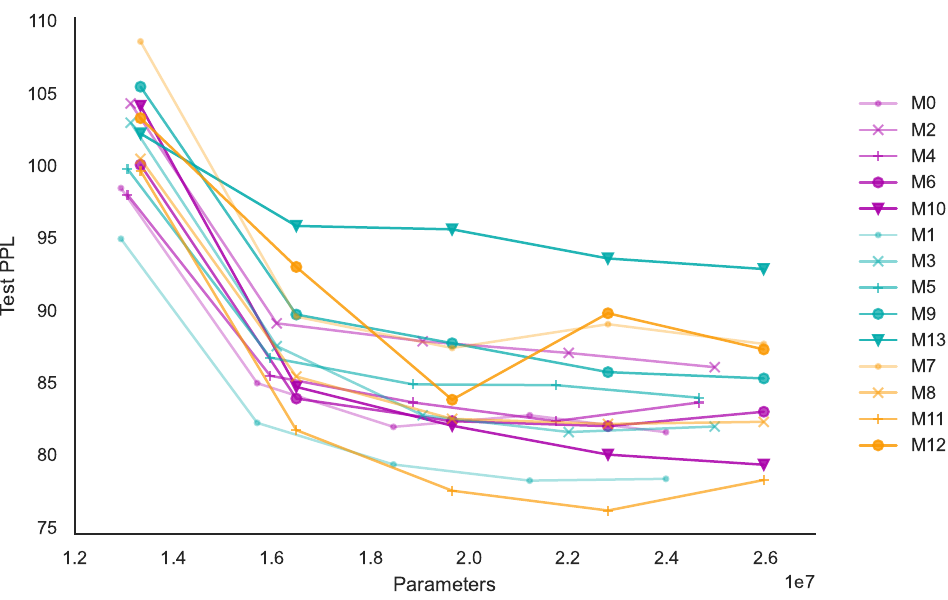}} \\[1em]
    \textbf{(b)}\\
    \makebox[5.1in][l]{\includegraphics[height=3in]{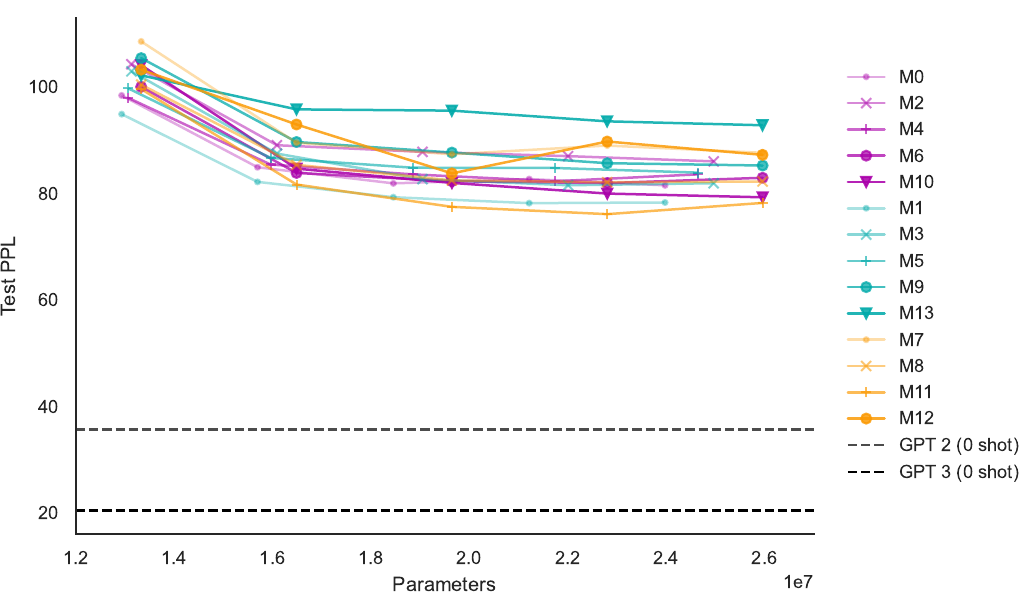}}
    \caption{\textbf{(a)} The results of training Transformer models based on the 14 attention variants identified above. They were trained \emph{ab initio} on word-level language modelling of the Penn Treebank corpus - all models have four attention heads per layer and an embedding dimension of 512. We used the learning-rate scheduler given by \cite{vaswani2017attention}, with initial learning rate tuned per-model. Each line on the plot shows the test-set perplexity of one model for between one and five layers, as compared to total trainable parameter count. The models are coloured according to whether they contain only linear attention generators (magenta), only dot-product attention generators (cyan), or both (orange). \textbf{(b)} The same with the results from \cite{radford2019language,brown2020language} for scale. Note that PPL is an exponential scale, so differences matter less as the PPL value increases.}
    \label{fig:results}
\end{figure}

To compare the 14 proposed attention mechanisms, we trained them \emph{ab initio} on a word-level language modelling task on the Penn Treebank corpus \cite{marcus1993building}. While specific attention mechanisms may be preferable for certain tasks, we believe this task is suitably representative to compare across different variants. We performed word-level tokenization on the dataset for a final vocabulary size of $\sim$10K. All trainable universal approximators specified in the generators are implemented as an MLP with one hidden layer with the same dimension as the input - note that this includes the transformations creating the key and query inputs for the \texttt{AttPrep} generator and value inputs for the \texttt{AttApply} generator, in contrast with the usual implementations of these transformations as linear mappings.  The results are given in Figure \ref{fig:results}. We observe that despite the significant differences between attention mechanisms, all models perform comparably -- indeed, the gap between the best- and worst-performing models is narrower than the range of performance we observed for any particular model during hyperparameter tuning. Furthermore, insofar as performance varies, there does not appear to be any obvious connection between the structure of each attention head and its performance.

\begin{figure}[p]
    \makebox[\textwidth]{$$\scalebox{0.95}{\input{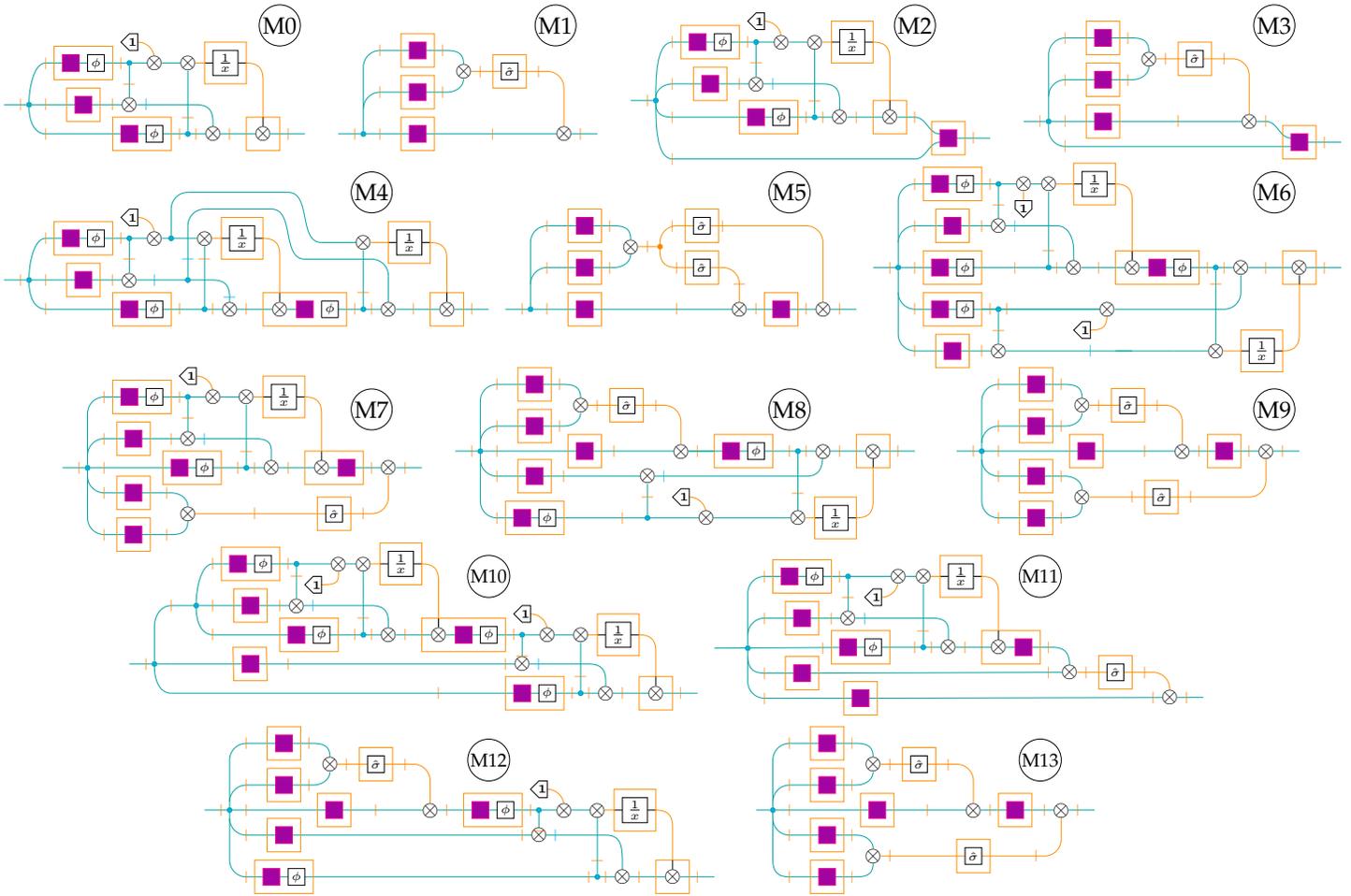}}$$}
    \caption{The attention mechanisms generated by exhaustively recombining the given generators, after removing redundant models using the criteria described above. Note that M0 corresponds precisely to the linear attention mechanism presented in \cite{katharopoulos2020transformers}, and M1 corresponds to scaled dot-product attention as presented in \cite{vaswani2017attention}.}
    \label{fig:attmechs}
\end{figure}

\section{Discussion}

We introduced string diagrams with rewrites for DL models, and demonstrated their conceptual utility in describing, comparing, and probing the structure of attention mechanisms. Our notation critically underpinned our ability to study a wide range of architectures in detail, and proved its worth in shaping and enabling our scientific inquiry; in a companion paper, we demonstrate the application of string diagrammatic notation for objective functions \cite{patternlanguage}. Our experimental findings from exploring the space of possible attention mechanisms suggest the following two possibilities:
\begin{itemize}[leftmargin=*]
\item{
Many popular explanations have been developed for the inner workings of Transformer models, based on the specific structure of the attention mechanism, e.g. \cite{alammar2018illustrated,sanderson2024attention}. Since all our attention variants performed comparably, one possibility is that these conceptual explanations, while intuitive and perhaps even locally correct, are not crucial for performance and do not constitute an explanation for why transformer-like models work in general. 
Rather, any expressive method of exchanging data between tokens ought to suffice -- this conclusion is borne out by other models such as FNet \cite{lee2022fnet} and MLPMixer \cite{tolstikhin2021mlpmixer} which work by a completely different mechanism.
}
\item{
Otherwise, if the anatomy of attention is important for performance, we have also observed that one model (M11 in Figure \ref{fig:attmechs}) slightly outperforms the classic scaled dot product attention, which is otherwise the most performant model. If this finding is reproducible at scale, then another possibility is that there exist very large attention mechanisms that are highly performant. Further investigation along these lines again critically depends on notation: these large mechanisms might be discovered more effectively by combinatorial search instead of refining particular anatomical elements by hand, and just describing them may already necessitate analysis in terms of components of the sort we have demonstrated.
}
\end{itemize}

\section*{Acknowledgements} The authors would like to thank Dimitri Kartsaklis, Steve Clark, and Bob Coecke for helpful feedback, Benjamin Rodatz, Ian Fan, Neil Ortega, and Konstantinos Meichanetzidis for useful discussions, and the rest of the Quantinuum Oxford office for their assistance.

\nocite{schlag_linear_2021}
\nocite{beltagy2020longformer}
\nocite{ainslie2020etc}
\nocite{guo2019star}
\nocite{guo2019gaussian}
\nocite{zaheer2021big}
\nocite{choromanski_rethinking_2022}
\nocite{yang2018modeling}
\nocite{you2020hard}

\bibliographystyle{apalike}
\bibliography{bibliography}

\appendix
\section{Formal semantics}\label{appendix:semantics}

Throughout, we assume familiarity with symmetric monoidal categories (SMCs), for which a standard reference is \cite{laneCategoriesWorkingMathematician2010a}. In this section we detail the formal semantics of our diagrammatic system. In a nutshell, each of our string diagrams is a indexed set of concrete programs, where the indexing corresponds to the user's choices in instantiating an architecture. There are several novel aspects of our diagrammatic notation that merit comment and mathematical elaboration, as they extend beyond the customary formal semantics of string diagrams in terms of symmetric monoidal functors \cite{selingerSurveyGraphicalLanguages2010a}.

\paragraph{Tick-notation for arbitrary dimensions}
We allow our diagrams to express (and thus hide the complexity of) choice-dependent dimensions, which wins us two novel benefits. First, we formally capture the intention of the informal flowcharts frequently used to illustrate architectures as referents to indexed families of models rather than specific programs; a transformer is still a transformer whether its programmer chooses the context window size to be $2^8$ or $2^{24}$.

Second, by analogy with game-theoretical semantics \cite{hintikkaChapterGameTheoreticalSemantics1997}, we can distil away numerals altogether from our diagrams to leave only the two kinds of integer-variables of interest to the engineer: arbitrary-but-fixed values that the programmer chooses, and the sequence-length of the input that the environment chooses. These two abstractions are valuable enough to merit the effort of establishing their formality. As is customary with the mathematics of string diagrams, the way to do so is to describe what individual pieces of diagrams mean in isolation, how they can be put together, and how their meanings interact in their composites.

\paragraph{Diagrammatic rewriting: universal approximation}
We opt to express the universal approximation theorem as the capacity for a universal approximator to be diagrammatically substituted for any other composite diagram with equal input and output. This amounts to asking that universal approximators are treated as ``typed contexts" in diagrams that are strict in the sense that different such contexts also compose in the same symmetric monoidal manner as the ambient processes. This leads to technical difficulties that rule out direct application of off-the-shelf methods for depicting open string diagrams derived from the coend calculus \cite{loregianCoendCalculus2021}, some notable examples being \cite{huExternalTracedMonoidala,Roman2020OpenDV,boisseauStringDiagramsOptics2020}. 
We define the coloured operad of a symmetric monoidal category in Construction \ref{cons:homop}, in which setting we are able to identify universal approximators with operad morphisms, viewed as the parse-trees of morphisms in the ambient SMC, thus directly interpreting them as typed contexts, and the expressivity reduction rewrites as operadic composition. For technical reasons involving scalars (the endomorphisms of the monoidal unit), this construction only works in semicartesian settings, i.e. where the monoidal unit is also terminal, but that is sufficiently general to admit our use cases, which are primary cartesian monoidal settings \cite{foxCoalgebrasCartesianCategories1976}, and are at worst Markov categories \cite{nlab_markov_category}.

\paragraph{Confluence}
Finally, because we define rewrites in the PROP setting, not the free tensoring, we have to show that the SIMD notation "plays nicely together" with rewrites. In this context, the relevant safety property to show is that the order in which rewrites and SIMD-reasoning occurs is immaterial. We achieve this by constructing a symmetric monoidal category from the hom-operad of a PROP in Construction \ref{cons:backtobasics}, which we demonstrate in Theorem \ref{thm:bigboy} is equivalent to the original PROP decorated with a novel formal black box for each input-output typing, such that we may completely delegate the symmetric monoidal semantics to the PROP and the expressivity reduction rewrites to the hom-operad.

\paragraph{On tensor-notation in machine learning and related work}
There are usually tradeoffs involved in tensor-notation for machine learning, but we sidestep these tradeoffs altogether with tick notation, underpinned by a free-tensoring construction on PROPs that we introduce. Thus we may win all of the benefits of tensor notation while keeping semantics formal and elementary, allowing us to further establish useful rewrite systems for calculation, and extend our consideration to virtually all sequence-to-sequence architectures. Simplifying tensor contraction for calculative purposes was in fact one of the original use-cases of string diagrams \cite{penroseApplicationsNegativeDimensional1971}, and the formal semantics of tensor-network notation in hypergraph categories \cite{fongHypergraphCategories2019b} is well understood. However, there is a fundamental mathematical obstacle in bringing together tensor notation and machine learning: any computational setting in which arbitrary data can be faithfully copied has string-diagrammatic semantics in a cartesian monoidal category by Fox's theorem \cite{foxCoalgebrasCartesianCategories1976}, and a monoidal product can only support both cartesian and hypergraph structures simultaneously if the category is trivial (\cite{coecke_picturing_2017}: Proposition 4.74). This mathematical fact has unfortunately forced various compromises in other tensorial approaches to string diagrams for transformers in particular. For instance, the "graphical tensor notation" approach \cite{taylorGraphicalTensorNotation} stays in the tensorial setting by employing a visually similar but informal analogue of functor boxes \cite{melliesFunctorialBoxesString2006b} to perform nonlinear transformations borrowed from the cartesian side, but this comes at steep costs to expressivity and formal difficulty: categories of smooth maps tend to be cartesian, so the tensorial monoidal product is undefined unless bubbles are restricted to be unconstrained endocombinators on homsets of linear maps. To keep what appears to be the desired expressivity, semantics may be declared in 2-categories of profunctor representations of open diagrams as previously mentioned to maintain both compact closure and access to generic nonlinear maps, at the expense of broad accessibility and verifiability of the formalism. Without addressing the underlying mathematical mismatch, increasing formal correctness also comes with its own drawbacks. For instance, the "neural circuit architecture" approach \cite{abbottNeuralCircuitDiagrams2024} attempts to incorporate both the tensorial and cartesian worlds by what appears to be an independent rediscovery of relational tape-diagrams \cite{bonchiDeconstructingCalculusRelations2023} and sheet-diagrams \cite{comfort2020sheet} in distributive monoidal settings. While in principle formally unambiguous, too much detail arguably comes at the expense of the raison d'\^{e}tre of string-diagrammatic notation: visual clarity and simplicity. Even discounting the lack of calculative and inferential power of the aforementioned approaches, we consider these shortcomings to be unacceptable and wholly unnecessary impediments to scientific inquiry in deep learning. Thus we state an alternative.

\subsection{SIMD notation}\label{subsec:simd}

\subsubsection{Free tensoring of a PROP}

We adopt the convention that $\mathbb{N}$ does not contain 0. We use the terms PROP and coloured PROP (for which a standard reference is \cite{yauHigherDimensionalAlgebras2008}) interchangeably after their introduction, and we assume that they are combinatorially generated by signatures so that we can speak unambiguously of the "data of a PROP". We refer to relations as subsets without loss of generality in the symmetric monoidal setting of $\textbf{Rel}^\times$, where strong compact closure makes input-output typing a matter of convention.

\paragraph{Ticks as tensor indices}

\begin{defn}[PROP]
A PROP is a strict symmetric monoidal category generated by a single object $x$: every object is of the form
$$\bigotimes^n x = x \underbrace{\otimes \cdots \otimes}_{n} x$$
PROPs may be generated by, and presented as \emph{signatures} $(\Sigma,E)$ consisting of generating morphisms $\Sigma$ with arity and coarity in $\mathbb{N}$, and equations $E$ relating symmetric monoidal composites of generators.
\end{defn}

\begin{example}\label{ex:cartsp}
The category \textbf{CartSp} of cartesian spaces $\mathbb{R}^n = \mathbb{R} \times \cdots \times \mathbb{R}$ and structured maps between them (\emph{i.e.} linear, continuous, smooth -- as one likes) is a PROP. \textbf{CartSp} is the intended mathematical setting for the final evaluation of our diagrams, and arguably of most deep learning.
\end{example}

\begin{defn}[Coloured PROP]
A \emph{multi-sorted} or \emph{coloured} PROP with set of colours $\mathfrak{C}$ has a monoid of objects generated by $\mathfrak{C}$.
\end{defn}

\begin{defn}[Cartesian PROP]
By Fox's theorem \cite{foxCoalgebrasCartesianCategories1976}, a cartesian PROP is one in which every object (wire) is equipped with a cocommutative comonoid (copy) with counit (delete) such that all morphisms in the category are comonoid cohomomorphisms.
\end{defn}

\begin{construction}[Tensoring of a PROP]\label{cons:tensoring}
Let $\mathcal{P}(i,j)$ for $i,j \in \mathbb{N}$ notate the homsets of a prop $\mathcal{P}$. Define the coloured PROP $\mathcal{P}^\otimes$ to have as colours finite lists of integers $\texttt{List}(\mathbb{N})$, and as morphisms, tensored copies $f^{\otimes k}$ of morphisms $f$ from the $\mathbf{P}$.
\end{construction}

\begin{defn}[Reshaping]
In any tensored PROP, we freely obtain the following family of \emph{reshaping} bijections on homsets by colouring and forgetting colours:
$$\mathcal{P}^\otimes(\bigotimes^k [i_1\cdots i_{n^k}],\bigotimes^l [j_1\cdots j_{m^l}]) \leftrightarrow \mathcal{P}(\prod^k\sum^{n^k} i_{n^k},\prod^l\sum^{m^l} j_{m^l})$$
\end{defn}

\begin{example}
Continuing with Example \ref{ex:cartsp}, the tensoring of $\textbf{CartSp}$ is denoted $\textbf{CartSp}^\otimes$. It has real-valued tensors as objects, denoted without loss of generality as ordered lists of indices $[i,j,k]$ decorating collections of objects $\mathbb{R}^{i \times j \times k}$ in the underlying PROP $\textbf{CartSp}$. As in all tensored-PROPs, every object comes equipped with a family of reshape isomorphisms:
$$[i_1\cdots i_n] \simeq [j^1_1\cdots j^1_{m^1}] \times \cdots \times [j^k_1, \cdots, j^k_{m^k}]$$
Where $\prod^n i_n = \sum^k \prod^{m^k} j^k_{m^k}$; \emph{i.e.} the integer totals match in the underlying PROP \textbf{CartSp}. Reshaping subsumes tensor concatenation along matching dimensions and reordering indices of tensors, which in turn subsumes matrix transposition as the special case for two indices. $\textbf{CartSp}^\otimes$ moreover has a family of \emph{tensor contraction} morphisms:
$$[a,b] \times [b,c] \rightarrow [a,c]$$
Which are comprised of the usual addition and multiplication monoids on $\mathbb{R}$ in the underlying PROP \textbf{CartSp}. Together, $\textbf{CartSp}^\otimes$ is an adequately expressive semantic category for all merely descriptive purposes in tensorial approaches to machine learning in nonlinear settings.
\end{example}

\subsubsection{Functorial semantics for arbitrary-choice indices on tensored PROPs}

Strategically, we aim to obtain functorial semantics for our diagrams by exploiting the observation that $\textbf{CartSp}^\otimes$ is a symmetric monoidal subcategory of the distributive monoidal category $\textbf{Rel}^\times$. So, analogously to Fock spaces \cite{fockKonfigurationsraumUndZweite1932}, we may use coproducts to "bundle together" objects and morphisms of $\textbf{CartSp}^\otimes$ according to their number of distinct indices, corresponding to the number of distinct choices to be made. Tactically, we prosecute this strategy in two steps. First, we define the required family of sets which will be our bundled-objects, and this immediately induces a full symmetric monoidal subcategory $\mathbb{F}(\mathbb{R}^\otimes)$ of $\textbf{Rel}^\times$ which will be the codomain of the semantic functor. Second, we define the functorial semantics of elements of our notation as indexed embeddings of $\textbf{CartSp}^\otimes$ in $\mathbb{F}(\mathbb{R}^\otimes)$, which obtains the intended semantics of sets of processes indexed by choices in the case of SIMD notation without traces, which we demonstrate by desugaring tensor contraction notation.
Finally, we deal with SIMD notation in generality as \emph{finite iteration}, and we list the the tracelike \cite{andreTracedMonoidalCategories1996} coherence laws governing the notation.

\paragraph{Target category}
\begin{defn}[$\mathbb{F}(\mathbb{R}^\otimes)$]\label{defn:targetcat}
For each $n \in \mathbb{N}$, let
\[\mathfrak{X}_n \triangleq \bigg(\coprod\limits_{i_1 \in \mathbb{N}}\cdots\coprod\limits_{i_n \in \mathbb{N}} \big( \mathbb{R}^{i_1} \times \cdots \times \mathbb{R}^{i_n} \big) \bigg) \times \underbrace{\mathbb{N} \times \cdots \times \mathbb{N}}_{n} \]
Let $\mathbb{F}(\mathbb{R}^\otimes)$ be the full symmetric monoidal subcategory of $\textbf{Rel}^\times$ generated by object-sets $\bigcup\limits_{n \in \mathbb{N}}\mathfrak{X}_n$.
\end{defn}

\paragraph{Functorial semantics for index-tick notation}
\begin{remark}\label{rem:specific}
The fundamental conceptual difficulty we seek to address in this section is that functorial semantics can only depend upon type-correctness of composing local components, but matching indices is a global correctness condition. This motivates our resort to relational semantics. We elaborate on generality and practicality in Remark \ref{rem:general}.
\end{remark}
Establishing functorial semantics in this instance amounts to declaring a compositional interpretation of our notational elements in $\mathbb{F}(\mathbb{R}^\otimes)$. For formal clarity, we switch to an austere variant of tick-notation. Instead of referring to choices using colours, we will use greeks.\\

\begin{defn}[Object semantics]\label{defn:obj}
Objects are ticked-wires, which we now interpret in $\mathbb{F}(\mathbb{R}^\otimes)$. The type of a vector with dimension varying with a choice $\alpha \in \mathbb{N}$ is interpreted as a relation (white dot) that selects the identity relation $\mathbb{R}^n \rightarrow \mathbb{R}^n$ for each $n \in \mathbb{N}$, combined with the special frobenius algebra (grey dot) on the copy-relation.
\[\input{austere/austereticksingle.tikz} \quad\triangleq\quad \input{austere/semanticssingle.tikz} \quad\triangleq\quad \{\big( (i,i) \ , (x,x) \big) \ | \ i \in \mathbb{N}, x \in \mathbb{R}^i \} \]
The type of a tensor with dimensions varying with $n$ (ordered) integer-choices $\alpha, \beta, \ldots, \omega$ is similarly:
\[\input{austere/multichoice.tikz} \quad\triangleq\quad \input{austere/semanticsmulti.tikz} \]
Where the choices $\alpha,\ldots,\omega$ correspond to the $n$ open $\mathbb{N}$ wires, and each relation with integer index $j$ selects identities as before, on the $j^{\text{th}}$ dimension. In particular, this means they commute. Thus the resulting relation is:
\[\{ \bigg( (i_1, i_1) , (i_2, i_2), \cdots , (i_n, i_n) , \Big( \overbrace{(x_{i_1}, x_{i_2}, \cdots x_{i_n})}^{n}, \overbrace{(x_{i_1}, x_{i_2}, \cdots, x_{i_n})}^{n} \Big) \bigg) \ | \ i_j \in \mathbb{N}, x_{i_j} \in \mathbb{R}^{i_j}\}\]
\end{defn}

This definition already allows us to account for several of our diagrammatic phenomena.

\begin{proposition}[Sequential composition of objects identifies choices]\label{prop:seqcompchoices}
\begin{proof}
For single ticks the verification is as follows; we show that the sequential composite of two single-ticked wires with differently labelled choices is definitionally equivalent to having a single choice. The multi-tick case is conceptually identical.
\[(\input{austere/singlealpha.tikz}) \fatsemi (\input{austere/singlebeta.tikz}) \quad\triangleq\quad \input{austere/simplesemsingle.tikz} \quad = \quad \input{austere/simplesemsingle2.tikz} \quad\triangleq\quad \input{austere/singlealpha.tikz}\]
\end{proof}
\end{proposition}

\begin{proposition}[Reshaping: transposition and rearranging indices]
Matrix transposition is a special case of index-rearrangement, which corresponds semantically to braiding the auxiliary $\mathbb{N}$ wires.
\[\input{austere/transpose.tikz} \quad\triangleq\quad \input{austere/reshapesem.tikz}\]
\begin{proof}
The integer-indexed identity selectors are commutative, and so in conjunction with Proposition \ref{prop:seqcompchoices}, transposition inherits naturality with respect to integer-indexed relations from the naturality of braids. All rearrangements of $n$ indices are elements of the permutation group of $n$ elements, which correspond to distinct braidings on $n$ wires, up to isotopy.
\end{proof}
\end{proposition}

\begin{proposition}[Fixed dimensions]\label{cons:fixdim}
We can recover variable-choice iteration over fixed dimensions via the injective embedding of $\coprod\limits_{\beta}(\mathbb{R}^k)^{\beta} \hookrightarrow \coprod\limits_{\alpha}\coprod\limits_{\beta}(\mathbb{R}^\alpha \times \mathbb{R}^\beta)$.
\[\input{austere/choosek.tikz} \quad\triangleq\quad \input{austere/chooseksem.tikz} \quad\simeq\quad \input{austere/Rknotation.tikz} \quad\triangleq\quad \input{austere/tickrkreal.tikz}\]
\begin{proof}
In prose: where $k$ is fixed, iterating $\beta$-times over $\mathbb{R}^k$ is equivalent to iterating $\alpha$ followed by $\beta$ times over $\mathbb{R}$, and then \emph{choosing} $\alpha \leftarrow: k$, which we can interpret as relational postselection. The claim then follows by inspection.
\end{proof}
\end{proposition}

\begin{defn}[Morphism semantics: Same Instruction, Multiple Data]\label{defn:simd}
For each morphism $f: \mathbb{R}^j \rightarrow \mathbb{R}^k$ in \textbf{CartSp}, we can define its iteration over a list of index choices $\Phi$ in a similar fashion as we defined objects.
\[\input{austere/f-alph.tikz} \quad\triangleq\quad \input{austere/SIMDfsemantics.tikz}\]
Without loss of generality, let $\Phi = \langle \alpha, \cdots \omega \rangle$ and $|\Phi| = n$. We define the selection relation $\ulcorner f^{\Phi} \urcorner$ to be the following subset of $\underbrace{\mathbb{N} \times \cdots \times \mathbb{N}}_{n} \times \coprod\limits_{\Phi}(\mathbb{R}^j)^\Phi \times \coprod\limits_{\Phi}(\mathbb{R}^k)^\Phi$:
\[\{ \Big( \alpha\cdots \omega, \mathbf{x}^\Phi, f(\mathbf{x}^\Phi) \Big) \ | \ \alpha, \cdots, \omega \in \mathbb{N} \}\]
Where $\mathbf{x}^\Phi$ denotes a (choice-variant) $\Phi$-indexed tensor of values $x_{i_\alpha\cdots i_\omega} \in \mathbb{R}^j$, and $f(\mathbf{x}^\Phi)$ denotes the $\Phi$-indexed tensor of outputs $f(x_{i_\alpha\cdots i_\omega}) \in \mathbb{R}^k$.
\end{defn}

\begin{proposition}\label{prop:simdcat}
Definition \ref{defn:simd} yields a symmetric monoidal category, where in particular Definition \ref{defn:obj} yields the identities. \begin{proof}
Unitality follows by construction, similarly to Proposition \ref{prop:seqcompchoices}. Associativity of composition and symmetric monoidal structure is inherited from $\textbf{Rel}^\times$, by Definition \ref{defn:targetcat}.
\end{proof}
\end{proposition}

\begin{defn}[\textbf{SIMD}(\textbf{CartSp})]
Let \textbf{SIMD}(\textbf{CartSp}) be the category defined by Proposition \ref{prop:simdcat}, which is the intended semantic category for SIMD diagrams.
\end{defn}

\begin{remark}\label{rem:general}
We have presented an elementary intended semantics in \textbf{Rel}, as opposed to the general construction for equipping arbitrary PROPs with arbitrary-indexing, which would extend consideration to nonclassical computational settings. The notation is arguably intuitive enough to not require further formal elaboration, so we are content with a sketch of the general construction, which may be intuited by the following visual correspondence:
\[\input{austere/HtoOptic.tikz} \quad\mapsto\quad \input{austere/HtoOptic2.tikz}\]
In prose, starting with a PROP $\mathcal{P}$, apply the \textbf{Optic} construction on the actegorical parameterisation \cite{capucciFoundationsCategoricalCybernetics2022b} of profunctorial open diagrams \cite{Roman2020OpenDV} in $\mathbb{F}(\mathcal{P}^\otimes)$ with a PROP evaluated at $\mathbb{N}$. While using \textbf{Prof} in this way as a categorification of \textbf{Rel} is pedestrian, the use of the forward-backward passes of the \textbf{Optic} construction however conceptually underpins the practical possibility of efficient functional (\emph{cf.} relational in Remark \ref{rem:specific}) implementation.
\end{remark}

\subsubsection{SIMD as finite iteration}

Executing the same process in parallel may be viewed (up to extensional equivalence) as a special case of finite recurrence with a residual, where the residual carries no data. Expanding the notion of SIMD to incorporate such \emph{finite iteration} is both practically and mathematically natural \emph{cf.} recurrent neural networks, and traced \cite{andreTracedMonoidalCategories1996}. The familial resemblance to traces suggests that we may provide a stock of diagrammatic laws to aid reasoning, which is what we aim for in this section.

\begin{defn}[Finite iteration]
\emph{Finite iteration} on a symmetric monoidal category $\mathcal{M}$ is a family of functions indexed over triples $(A,B,U)$ of objects of $\mathcal{M}$ and all positive integers $k \in \mathbb{N}$: 
$$\mathcal{M}(U \otimes A, U \otimes B) \overset{\langle - \rangle^k_U}\rightarrow \mathcal{M}(U \otimes A^{\otimes k}, U \otimes B^{\otimes k})$$
Given by the following inductive definition for all $f: U \otimes A \rightarrow U \otimes B$, where $\theta_{(X,Y)}$ denotes the braiding:
$$\begin{cases}
\langle f \rangle^1_U \triangleq f \\
\langle f \rangle^{k+1}_U \triangleq (\theta_{(B^{\otimes k}, U)} \otimes 1_B) \circ (1_{B^{\otimes k}} \otimes f) \circ ((\theta_{(B^{\otimes k}, U)} \circ \langle f \rangle^k_U) \otimes 1_A)
\end{cases}$$
The inductive case is defined up to braid isomorphism. Since the ambient category is symmetric monoidal, the right-to-left order of passing the residual $U$ is a choice of convention. We may present this inductive definition diagrammatically as:
\[\input{SIMD/inductivebase.tikz} \quad\triangleq\quad \input{SIMD/f.tikz}\]
\[\input{SIMD/inductivestep.tikz} \quad\triangleq\quad \input{SIMD/inductivestepexpand.tikz}\]
\end{defn}

\begin{proposition}
SIMD is a special case of finite iteration with residual $I$, the monoidal unit. \begin{proof} Immediate, from monoidal coherence. \end{proof}
\end{proposition}

\begin{notation}[Finite iteration]
To more explicitly relate finite iteration to feedback and traced structure, we notate $\langle f \rangle^\alpha_U$ following the same conventions as an operator on processes with a bolded feedback edge. We remove the bolding in the special case of SIMD. Given a morphism $f: U \times A \rightarrow U \times B$ in \textbf{CartSp}, we define in \textbf{SIMD(CartSp)} the relation $\ulcorner \langle f \rangle ^{-}_U \urcorner: \mathbb{N} \times U \times \coprod\limits_{\alpha}A^{\otimes\alpha} \times \coprod\limits_{\alpha}B^{\otimes\alpha} \rightarrow \coprod\limits_{\alpha}A^{\otimes\alpha} \times \coprod\limits_{\alpha}B^{\otimes\alpha}$ to select the appropriate finite iteration given a positive integer parameter, and we define the following notation:
\[\input{SIMD/tracenotation.tikz} \quad\triangleq\quad \input{austere/SIMDproper.tikz}\]
\end{notation}

Before we explicitly relate finite iteration notation to tracelike structure, we introduce some common notational tools for manipulating indices via examples.

\begin{example}[Reshaping: splitting and concatenation]\label{ex:concat}
Concatenation amounts to tensoring wires and adding indices, and splitting is co-concatenation, which relies on the converse of the addition operation in \textbf{Rel}.
\[\input{austere/concatnotation.tikz} \quad\triangleq\quad \input{austere/concatsemantics.tikz}\]
\[\input{austere/splitnotation.tikz} \quad\triangleq\quad \input{austere/splitsemantics.tikz}\]

\end{example}

\begin{proposition}\label{prop:finitgeneralise}
Modulo notation, finite iteration satisfies \emph{tightening}, and \emph{yanking}.
\begin{proof} The following laws hold by symmetric monoidal coherence, and unpacking the definitions of finite iteration.
\[\input{SIMD/tightening0.tikz} \quad =\quad \input{SIMD/tightening1.tikz} \tag{\text{Tightening}}\]
A precondition for yanking is that the residual matches the input and output: $U = A = B$.
\[\input{SIMD/yanking0.tikz} \quad =\quad \input{SIMD/yanking1.tikz} \tag{\text{Yanking}}\]
\end{proof}
\end{proposition}

\begin{proposition}\label{prop:finitsamish}
Finite iteration satisfies a variant \emph{strength} condition, similar to monoidal strength.
\[\input{SIMD/strength0.tikz} \quad =\quad \input{SIMD/strength1.tikz} \tag{\text{Strength}}\]
\begin{proof}
By inspection.
\end{proof}
\end{proposition}

Owing to the integer parameter on each finite iteration structure, finite iteration case-splits the \emph{sliding} and \emph{vanishing} laws, the former of which demands more subtle interactions with the iteration count, and the latter of which must account for the possibility of differing iteration choices in nested structure.

\begin{proposition}\label{prop:finitslide}
\[\input{SIMD/Lsliding0.tikz} \quad =\quad \input{SIMD/Rsliding1.tikz} \tag{\text{Sliding-1}}\]
\[\input{SIMD/Rsliding0.tikz} \quad =\quad \input{SIMD/Lsliding1.tikz} \tag{\text{Sliding-2}}\]
\begin{proof}
By inspection.
\end{proof}
\end{proposition}

\begin{proposition}\label{prop:finitvanish}
\[\input{SIMD/vanishing0.tikz} \quad =\quad \input{SIMD/vanishing1.tikz} \tag{\text{Vanishing}}\]
\begin{proof}
By inspection.
\end{proof}
\end{proposition}

\begin{corollary}
The above laws characterise finite iteration.
\begin{proof}
Finite iteration yields the laws by Propositions \ref{prop:finitgeneralise}, \ref{prop:finitsamish}, \ref{prop:finitslide}, and \ref{prop:finitvanish}. Conversely, instantiating $g$ as the identity in the sliding laws recovers the inductive definition of finite iteration.
\end{proof}
\end{corollary}

\subsection{Universal approximators and expressive reductions}\label{appendix:univ}

In this section, we deal with equipping generic coloured cartesian PROPs $\mathcal{P}$ with universal approximators with the intuitively direct interpretation as typed holes in diagrams. The formal underpinning is via considering universal approximators to be morphisms in an operad algebra valued in the homsets of $\mathcal{P}$.

\begin{defn}[(symmetric, unital) coloured operad]
Where $(\mathcal{V},\boxtimes,J)$ is a symmetric monoidal category and $\mathfrak{C}$ denotes a set of \emph{colours} $c_i$, a coloured operad $\mathcal{O}$ consists of:
\begin{itemize}
\item{For each $n \in \mathbb{N}$ and each $(n+1)$-tuple $(c_1, \cdots, c_n; c)$, an object $\mathcal{O}(c_1, \cdots, c_n; n) \in \mathcal{V}$}
\item{For each $c \in \mathfrak{C}$, a morphism $1_c : J \rightarrow \mathcal{O}(c;c)$ called the \emph{identity of} $c$}
\item{For each $(n+1)$-tuple $(c_1 \cdots c_n; c)$ and $n$ other tuples $(d^1_1\cdots d^1_{k_1})\cdots (d^n_1\cdots d^n_{k_n})$ a \emph{composition morphism} \[\mathcal{O}(c_1, \cdots, c_n; c) \boxtimes \mathcal{O}(d^1_1\cdots d^1_{k_1}) \boxtimes \cdots \boxtimes \mathcal{O}(d^n_1\cdots d^n_{k_n}) \rightarrow \mathcal{O}(d^1_1\cdots d^1_{k_1}\cdots d^n_1\cdots d^n_{k_n};c)\]}
\item{for all $n \in \mathbb{N}$, all tuples of colours, and each permutation $\sigma \in S_n$ the symmmetric group on $n$, a morphism:
\[\sigma^*: \mathcal{O}(c_1\cdots c_n;c) \rightarrow \mathcal{O}(c_{\sigma^*(1)}\cdots c_{\sigma^*(n)};c)\]}
\end{itemize}
The $\sigma^*$ must represent $S_n$, and composition must satisfy associativity and unitality in a $S_n$-invariant manner.
\end{defn}

\begin{construction}[Hom-Operad of coloured PROP]\label{cons:homop}
Where ($\mathcal{P},\otimes,I)$ is a coloured PROP with colours $\mathfrak{C}_\mathcal{P}$, we construct $\mathcal{O}_\mathcal{P}$, the \emph{hom-operad} of $\mathcal{P}$. We do so in two stages, by first defining an ambient operad, and then restricting to the operad obtained by a collection of generators. Let the ambient symmetric monoidal category be $(\mathbf{Set},\times,\{\star\})$. Let the colours $\mathfrak{C}_\mathcal{O}$ be the set of all tuples $(\mathbf{A},\mathbf{B})$, each denoting a pair of tuples $(A_1 \otimes A_n, B_1 \otimes B_n)$ of $A_i,B_i \in \mathfrak{C}_\mathcal{P}$.

\begin{itemize}
\item{The tuple $\big((\mathbf{A}^1,\mathbf{B}^1)\cdots(\mathbf{A}^n,\mathbf{B}^n);(\mathbf{A},\mathbf{B})\big)$ is assigned the set $[\mathcal{P}(\mathbf{A}^1,\mathbf{B}^1) \times \cdots \times \mathcal{P}(\mathbf{A}^n,\mathbf{B}^n) \rightarrow \mathcal{P}(\mathbf{A},\mathbf{B})] \in \mathbf{Set}$; the set of all \emph{generated} functions from the product of homsets $\mathcal{P}(\mathbf{A}^i,\mathbf{B}^i)$ to the homset $\mathcal{P}(\mathbf{A},\mathbf{B})$.}
\item{$1_{(\mathbf{A},\mathbf{B})} : \{\star\} \rightarrow [\mathcal{P}(\mathbf{A},\mathbf{B}) \rightarrow \mathcal{P}(\mathbf{A},\mathbf{B})]$ is the identity functional that maps $f: \mathbf{A} \rightarrow \mathbf{B}$ in $\mathcal{P}(\mathbf{A},\mathbf{B})$ to itself.}
\item{The composition operations correspond to function composition in $\mathbf{Set}$, where $[X \rightarrow Y] \times [Y \rightarrow Z] \rightarrow [X \rightarrow Z]$ sends $(f_{: X \rightarrow Y},g_{: Y \rightarrow Z}) \mapsto (g \circ f)_{: X \rightarrow Z}$; appropriately generalised to the multi-argument case. The permutations are similarly defined, inheriting their coherence conditions from the commutativity isomorphisms of the categorical product $\times$.}
\end{itemize}

The generators are:
\begin{itemize}
\item{For every $f \in \mathcal{P}(\mathbf{A},\mathbf{B})$ that is a generator of $\mathcal{P}$, define a corresponding generator of type $\{\star\} \rightarrow [\mathcal{P}(I,I) \rightarrow \mathcal{P}(\mathbf{A},\mathbf{B})]$, which is the functional $\big(- \mapsto (f \otimes -)\big)$ that sends endomorphisms of the monoidal unit of $\mathcal{P}$ to their tensor with $f$, viewed as an element of the set $[\mathcal{P}(I,I) \rightarrow \mathcal{P}(\mathbf{A},\mathbf{B})]$.}
\item{For every pair of tuples $\big((\mathbf{X}^1,\mathbf{Y}^1) \cdots (\mathbf{X}^m,\mathbf{Y}^m) ;(\mathbf{A},\mathbf{B})\big)$ and $\big((\mathbf{J}^1,\mathbf{K}^1) \cdots (\mathbf{J}^n,\mathbf{K}^n) ;(\mathbf{B},\mathbf{C})\big)$ in $\mathfrak{C}_\mathcal{O}$, a corresponding \emph{sequential composition} operation of type:
\[[\prod\limits_{i \leqslant m} \mathcal{P}(\mathbf{X}^i,\mathbf{Y}^i) \rightarrow \mathcal{P}(\mathbf{A},\mathbf{B})] \times [\prod\limits_{j \leqslant n} \mathcal{P}(\mathbf{J}^j,\mathbf{K}^j) \rightarrow \mathcal{P}(\mathbf{B},\mathbf{C})]\] \[\rightarrow [\big(\prod\limits_{i \leqslant m} \mathcal{P}(\mathbf{X}^i,\mathbf{Y}^i) \times \prod\limits_{j \leqslant n} \mathcal{P}(\mathbf{J}^j,\mathbf{K}^j)\big) \rightarrow \mathcal{P}(\mathbf{A},\mathbf{C})]\]
Which maps pairs of functionals $(F_{: \prod\limits_{i \leqslant m} \mathcal{P}(\mathbf{X}^i,\mathbf{Y}^i) \rightarrow \mathcal{P}(\mathbf{A},\mathbf{B})}, G_{: \prod\limits_{j \leqslant n} \mathcal{P}(\mathbf{J}^j,\mathbf{K}^j) \rightarrow \mathcal{P}(\mathbf{B},\mathbf{C})})$ to the functional which sends $p^i: \mathbf{X}^i \rightarrow \mathbf{Y}^i$ and $q^j: \mathbf{X}^j \rightarrow \mathbf{Y}^j$ to $G(p_1\cdots p_m) \circ F(q_1 \cdots q_n)$.
}
\item{An analogous \emph{parallel composition} for every pair of tuples, which sends pairs of functionals $(F,G)$ to $G(p_1\cdots p_m) \otimes F(q_1 \cdots q_n)$.}
\end{itemize}
\end{construction}

\begin{example}\label{ex:bridge} Construction \ref{cons:homop} can be thought of as bridging diagrams with their specific algebraic descriptions using just the basic constructors $\circ, \otimes$; the hom-operad (when notated suggestively in the usual tree-notation, found e.g. in \cite{marklOperadsAlgebraTopology2007}) plays the role of the syntactic tree of $\circ, \otimes$ operators. For instance, given the composite morphism $(g \otimes 1_E) \circ (1_A \otimes f)$ in PROP $\mathcal{P}$, the corresponding diagram and operad-state in $\mathcal{O}_\mathcal{P}$ is:
\[\begin{tikzpicture}
	\begin{pgfonlayer}{nodelayer}
		\node [style=none] (0) at (1, 1.25) {};
		\node [style=none] (1) at (1, -0.25) {};
		\node [style=none] (2) at (1.5, -0.25) {};
		\node [style=none] (3) at (1.5, 1.25) {};
		\node [style=none] (4) at (-0.5, 0.25) {};
		\node [style=none] (5) at (-0.5, -1.25) {};
		\node [style=none] (6) at (0, -1.25) {};
		\node [style=none] (7) at (0, 0.25) {};
		\node [style=none] (8) at (0, 0) {};
		\node [style=none] (9) at (1, 0) {};
		\node [style=none] (10) at (-1, 1) {};
		\node [style=none] (11) at (1, 1) {};
		\node [style=none] (12) at (-0.5, -0.5) {};
		\node [style=none] (13) at (-1, -0.5) {};
		\node [style=none] (14) at (1.5, 0.5) {};
		\node [style=none] (15) at (2, 0.5) {};
		\node [style=none] (16) at (-0.25, -0.5) {\tiny $f$};
		\node [style=none] (17) at (1.25, 0.5) {\tiny $g$};
		\node [style=none] (18) at (-1.25, 1) {\tiny$A$};
		\node [style=none] (19) at (-1.25, -0.5) {\tiny$B$};
		\node [style=none] (21) at (2.25, 0.5) {\tiny$D$};
		\node [style=none] (22) at (0.5, 1.5) {};
		\node [style=none] (23) at (0.5, -1.5) {};
		\node [style=none] (24) at (2, 1.5) {};
		\node [style=none] (25) at (2, -1.5) {};
		\node [style=none] (26) at (-1, 1.5) {};
		\node [style=none] (27) at (-1, -1.5) {};
		\node [style=none] (28) at (-1, 0.5) {};
		\node [style=none] (29) at (0.5, 0.5) {};
		\node [style=none] (31) at (-0.5, -1) {};
		\node [style=none] (32) at (0, -1) {};
		\node [style=none] (33) at (2, -1) {};
		\node [style=none] (34) at (0.5, -0.5) {};
		\node [style=none] (35) at (2, -0.5) {};
		\node [style=none] (36) at (2.25, -1) {\tiny$E$};
		\node [style=none] (37) at (3, 0) {$\leftrightarrow$};
		\node [style=none] (41) at (0.5, 0.25) {\tiny$C$};
		\node [style=operadot] (48) at (4.25, 2.25) {\tiny$\mathbf{1}_A$};
		\node [style=operadot] (49) at (4.25, 0.75) {\footnotesize$f$};
		\node [style=operadot] (51) at (4.25, -2.25) {\tiny$\mathbf{1}_E$};
		\node [style=operadot] (52) at (4.25, -0.75) {\footnotesize$g$};
		\node [style=operadot] (53) at (17.5, 0) {\footnotesize$\circ$};
		\node [style=operadot] (54) at (10.5, 1.5) {\footnotesize$\otimes$};
		\node [style=operadot] (55) at (10.5, -1.5) {\footnotesize$\otimes$};
		\node [style=none] (56) at (25.25, 0) {};
		\node [style=none] (57) at (8, 2.5) {\tiny \textcolor{gray}{$[\mathcal{P}(I,I) \rightarrow \mathcal{P}(A,A)]$}};
		\node [style=none] (58) at (8, 0.5) {\tiny \textcolor{gray}{$[\mathcal{P}(I,I) \rightarrow \mathcal{P}(B,C \otimes E)]$}};
		\node [style=none] (59) at (8, -0.5) {\tiny \textcolor{gray}{$[\mathcal{P}(I,I) \rightarrow \mathcal{P}(A \otimes C, D)]$}};
		\node [style=none] (60) at (8, -2.5) {\tiny \textcolor{gray}{$[\mathcal{P}(I,I) \rightarrow \mathcal{P}(E,E)]$}};
		\node [style=none] (61) at (15.25, 1.75) {\tiny \textcolor{gray}{$[\mathcal{P}(I,I)^2 \rightarrow \mathcal{P}(A \otimes B,A \otimes C \otimes E)]$}};
		\node [style=none] (62) at (15.25, -1.75) {\tiny \textcolor{gray}{$[\mathcal{P}(I,I)^2 \rightarrow \mathcal{P}(A \otimes C \otimes E, D \otimes E)]$}};
		\node [style=none] (63) at (21.75, 0.5) {\tiny \textcolor{gray}{$[\mathcal{P}(I,I)^3 \rightarrow \mathcal{P}(A \otimes B, D \otimes E)]$}};
	\end{pgfonlayer}
	\begin{pgfonlayer}{edgelayer}
		\draw (1.center)
			 to (2.center)
			 to (3.center)
			 to (0.center)
			 to cycle;
		\draw (5.center)
			 to (6.center)
			 to (7.center)
			 to (4.center)
			 to cycle;
		\draw (8.center) to (9.center);
		\draw (10.center) to (11.center);
		\draw (13.center) to (12.center);
		\draw (14.center) to (15.center);
		\draw [style=Dcyan] (22.center) to (23.center);
		\draw [style=Dcyan] (26.center)
			 to (22.center)
			 to (24.center)
			 to (25.center)
			 to (23.center)
			 to (27.center)
			 to cycle;
		\draw [style=Dcyan] (29.center) to (28.center);
		\draw (32.center) to (33.center);
		\draw [style=Dcyan] (34.center) to (35.center);
		\draw [style=gray] (48) to (54);
		\draw [style=gray] (49) to (54);
		\draw [style=gray] (52) to (55);
		\draw [style=gray] (51) to (55);
		\draw [style=gray] (55) to (53);
		\draw [style=gray] (54) to (53);
		\draw [style=gray] (53) to (56.center);
	\end{pgfonlayer}
\end{tikzpicture}
\]
Since the PROPs \textbf{CartSp} and its free tensoring are cartesian, $\mathcal{P}(I,I)$ is a singleton containing only the identity of the monoidal unit, so in the settings we are concerned with, we may simplify colours of the form $[\mathcal{P}(I,I)^N \rightarrow \mathcal{P}(\mathbf{A},\mathbf{B})]$ to just $\mathcal{P}(\mathbf{A},\mathbf{B})$, and operad-states $\{\star\} \rightarrow \mathcal{P}(\mathbf{A},\mathbf{B})$ are in bijective correpondence with morphisms $f: \mathbf{A} \rightarrow \mathbf{B}$ of $\mathcal{P}$; the fact that all $f: \mathbf{A} \rightarrow \mathbf{B}$ are representable as operad states follows by construction, since any $f$ in $\mathcal{P}$ is by definition expressible in terms of the generators of $\mathcal{P}$, and sequential and parallel composition $\circ,\otimes$. As we assume homsets are already quotiented by the equational theory of $\mathcal{P}$ and the symmetric monoidal coherences, our operadic representations inherit them: for example, we obtain interchange equalities such as the one below for free:
\[\begin{tikzpicture}
	\begin{pgfonlayer}{nodelayer}
		\node [style=operadot] (65) at (-13.75, 0.75) {\footnotesize$v$};
		\node [style=none] (72) at (-12.25, 2.5) {\tiny \textcolor{gray}{$\mathcal{P}(A,C)$}};
		\node [style=none] (76) at (-19, 0.75) {};
		\node [style=none] (77) at (-19, 0.25) {};
		\node [style=none] (78) at (-18.5, 0.25) {};
		\node [style=none] (79) at (-18.5, 0.75) {};
		\node [style=none] (80) at (-19, -0.25) {};
		\node [style=none] (81) at (-19, -0.75) {};
		\node [style=none] (82) at (-18.5, -0.75) {};
		\node [style=none] (83) at (-18.5, -0.25) {};
		\node [style=none] (84) at (-17, 0.75) {};
		\node [style=none] (85) at (-17, 0.25) {};
		\node [style=none] (86) at (-16.5, 0.25) {};
		\node [style=none] (87) at (-16.5, 0.75) {};
		\node [style=none] (88) at (-17, -0.25) {};
		\node [style=none] (89) at (-17, -0.75) {};
		\node [style=none] (90) at (-16.5, -0.75) {};
		\node [style=none] (91) at (-16.5, -0.25) {};
		\node [style=none] (92) at (-19.5, 0.5) {};
		\node [style=none] (93) at (-19, 0.5) {};
		\node [style=none] (94) at (-18.5, 0.5) {};
		\node [style=none] (95) at (-17, 0.5) {};
		\node [style=none] (96) at (-16.5, 0.5) {};
		\node [style=none] (97) at (-16, 0.5) {};
		\node [style=none] (98) at (-19.5, -0.5) {};
		\node [style=none] (99) at (-19, -0.5) {};
		\node [style=none] (100) at (-18.5, -0.5) {};
		\node [style=none] (101) at (-17, -0.5) {};
		\node [style=none] (102) at (-16.5, -0.5) {};
		\node [style=none] (103) at (-16, -0.5) {};
		\node [style=none] (104) at (-19.75, 0.5) {\tiny $A$};
		\node [style=none] (105) at (-19.75, -0.5) {\tiny $B$};
		\node [style=none] (106) at (-17.75, 0.25) {\tiny $C$};
		\node [style=none] (107) at (-17.75, -0.25) {\tiny $D$};
		\node [style=none] (108) at (-15.75, 0.5) {\tiny $E$};
		\node [style=none] (109) at (-15.75, -0.5) {\tiny $F$};
		\node [style=none] (110) at (-18.75, 0.5) {\tiny $u$};
		\node [style=none] (111) at (-18.75, -0.5) {\tiny $v$};
		\node [style=none] (112) at (-16.75, 0.5) {\tiny $w$};
		\node [style=none] (113) at (-16.75, -0.5) {\tiny $x$};
		\node [style=none] (114) at (-19.5, 1) {};
		\node [style=none] (115) at (-19.5, -1) {};
		\node [style=none] (116) at (-16, 1) {};
		\node [style=none] (117) at (-16, -1) {};
		\node [style=none] (118) at (-19.5, 0) {};
		\node [style=none] (119) at (-16, 0) {};
		\node [style=none] (120) at (-17.75, 1) {};
		\node [style=none] (121) at (-17.75, -1) {};
		\node [style=operadot] (122) at (-13.75, 2.25) {\footnotesize$u$};
		\node [style=operadot] (123) at (-13.75, -0.75) {\footnotesize$w$};
		\node [style=operadot] (124) at (-13.75, -2.25) {\footnotesize$x$};
		\node [style=none] (125) at (-14.75, 0) {$\leftrightarrow$};
		\node [style=operadot] (126) at (-12.25, -1.5) {\footnotesize$\otimes$};
		\node [style=operadot] (127) at (-12.25, 1.5) {\footnotesize$\otimes$};
		\node [style=operadot] (128) at (-9.25, 0) {\footnotesize$\circ$};
		\node [style=none] (129) at (-5, 0) {};
		\node [style=none] (130) at (-12.25, 0.5) {\tiny \textcolor{gray}{$\mathcal{P}(B,D)$}};
		\node [style=none] (131) at (-12.25, -0.5) {\tiny \textcolor{gray}{$\mathcal{P}(C,E)$}};
		\node [style=none] (132) at (-12.25, -2.5) {\tiny \textcolor{gray}{$\mathcal{P}(D,F)$}};
		\node [style=none] (133) at (-9.25, 1.25) {\tiny \textcolor{gray}{$\mathcal{P}(A \otimes B ,C \otimes D)$}};
		\node [style=none] (134) at (-9.25, -1.25) {\tiny \textcolor{gray}{$\mathcal{P}(C \otimes D, E \otimes F)$}};
		\node [style=none] (135) at (-6.75, 0.5) {\tiny \textcolor{gray}{$\mathcal{P}(A \otimes B, E \otimes F)$}};
		\node [style=operadot] (136) at (-2.75, 0.75) {\footnotesize$w$};
		\node [style=none] (137) at (-1.25, 2.5) {\tiny \textcolor{gray}{$\mathcal{P}(A,C)$}};
		\node [style=operadot] (138) at (-2.75, 2.25) {\footnotesize$u$};
		\node [style=operadot] (139) at (-2.75, -0.75) {\footnotesize$v$};
		\node [style=operadot] (140) at (-2.75, -2.25) {\footnotesize$x$};
		\node [style=operadot] (141) at (-1.25, -1.5) {\footnotesize$\circ$};
		\node [style=operadot] (142) at (-1.25, 1.5) {\footnotesize$\circ$};
		\node [style=operadot] (143) at (0.5, 0) {\footnotesize$\otimes$};
		\node [style=none] (144) at (4.75, 0) {};
		\node [style=none] (145) at (-1.25, -0.5) {\tiny \textcolor{gray}{$\mathcal{P}(B,D)$}};
		\node [style=none] (146) at (-1.25, 0.5) {\tiny \textcolor{gray}{$\mathcal{P}(C,E)$}};
		\node [style=none] (147) at (-1.25, -2.5) {\tiny \textcolor{gray}{$\mathcal{P}(D,F)$}};
		\node [style=none] (148) at (0.5, 1.25) {\tiny \textcolor{gray}{$\mathcal{P}(A,E)$}};
		\node [style=none] (149) at (0.5, -1.25) {\tiny \textcolor{gray}{$\mathcal{P}(B,F)$}};
		\node [style=none] (150) at (3, 0.5) {\tiny \textcolor{gray}{$\mathcal{P}(A \otimes B, E \otimes F)$}};
		\node [style=none] (151) at (-4, 0) {$=$};
	\end{pgfonlayer}
	\begin{pgfonlayer}{edgelayer}
		\draw (76.center)
			 to (79.center)
			 to (78.center)
			 to (77.center)
			 to cycle;
		\draw (80.center)
			 to (83.center)
			 to (82.center)
			 to (81.center)
			 to cycle;
		\draw (84.center)
			 to (87.center)
			 to (86.center)
			 to (85.center)
			 to cycle;
		\draw (88.center)
			 to (91.center)
			 to (90.center)
			 to (89.center)
			 to cycle;
		\draw (94.center) to (95.center);
		\draw (96.center) to (97.center);
		\draw (92.center) to (93.center);
		\draw (98.center) to (99.center);
		\draw (100.center) to (101.center);
		\draw (102.center) to (103.center);
		\draw [style=Dcyan] (118.center) to (119.center);
		\draw [style=Dcyan] (116.center)
			 to (114.center)
			 to (115.center)
			 to (117.center)
			 to cycle;
		\draw [style=Dcyan] (120.center) to (121.center);
		\draw [style=gray] (122) to (127);
		\draw [style=gray] (65) to (127);
		\draw [style=gray] (123) to (126);
		\draw [style=gray] (124) to (126);
		\draw [style=gray] (127) to (128);
		\draw [style=gray] (126) to (128);
		\draw [style=gray] (128) to (129.center);
		\draw [style=gray] (138) to (142);
		\draw [style=gray] (136) to (142);
		\draw [style=gray] (139) to (141);
		\draw [style=gray] (140) to (141);
		\draw [style=gray] (142) to (143);
		\draw [style=gray] (141) to (143);
		\draw [style=gray] (143) to (144.center);
	\end{pgfonlayer}
\end{tikzpicture}
\]
\end{example}

\begin{defn}[Universal approximators and expressivity reductions]
A morphism of a coloured PROP $\mathcal{P}$ of type $(\mathbf{A},\mathbf{B})$ containing universal approximators as black-boxes of types $\mathbf{A}^{i \leqslant n} \rightarrow \mathbf{B}^{i \leqslant n}$ is a morphism $\big((\mathbf{A}^1,\mathbf{B}^1)\cdots(\mathbf{A}^n,\mathbf{B}^n);(\mathbf{A},\mathbf{B})\big)$ of $\mathcal{O}_\mathcal{P}$, and by construction, vice versa. Expressivity reductions correspond to precomposition in $\mathcal{O}_\mathcal{P}$.
\end{defn}

\begin{example} The inputs of open morphisms in $\mathcal{O}_\mathcal{P}$ correspond to "typed holes", and operadic precomposition corresponds to "filling holes", with contents that may themselves also contain typed holes. This precisely formalises the intuition that expressive reductions correspond to the ability of a universal approximator to simulate anything, including composites containing other universal approximators.
\[\begin{tikzpicture}
	\begin{pgfonlayer}{nodelayer}
		\node [style=none] (0) at (-8, 0.75) {};
		\node [style=none] (1) at (-8, -0.25) {};
		\node [style=none] (2) at (-7.5, -0.25) {};
		\node [style=none] (3) at (-7.5, 0.75) {};
		\node [style=none] (4) at (-7.75, 0.25) {\tiny$f$};
		\node [style=none] (5) at (-9.5, 0.5) {};
		\node [style=none] (6) at (-8, 0.5) {};
		\node [style=none] (7) at (-7.5, 0.5) {};
		\node [style=none] (8) at (-7, 0.5) {};
		\node [style=none] (9) at (-8, 0) {};
		\node [style=none] (10) at (-9.5, 0) {};
		\node [style=none] (11) at (-7.5, 0) {};
		\node [style=none] (12) at (-7, 0) {};
		\node [style=none] (13) at (-9.5, -0.5) {};
		\node [style=none] (14) at (-7, -0.5) {};
		\node [style=none] (15) at (-9, 0) {};
		\node [style=none] (16) at (-8.5, 0) {};
		\node [style=none] (17) at (-9, -0.5) {};
		\node [style=none] (18) at (-8.5, -0.5) {};
		\node [style=none] (19) at (-9, 0.25) {};
		\node [style=none] (20) at (-9, -0.75) {};
		\node [style=none] (21) at (-8.5, -0.75) {};
		\node [style=none] (22) at (-8.5, 0.25) {};
		\node [style=none] (23) at (-2.75, 1.25) {};
		\node [style=none] (24) at (-2.75, -0.25) {};
		\node [style=none] (25) at (-1.75, -0.25) {};
		\node [style=none] (26) at (-1.75, 1.25) {};
		\node [style=none] (27) at (-2.25, 0.5) {\tiny$f$};
		\node [style=none] (28) at (-4.5, 1) {};
		\node [style=none] (29) at (-2.75, 1) {};
		\node [style=none] (30) at (-1.75, 1) {};
		\node [style=none] (31) at (-1.5, 1) {};
		\node [style=none] (32) at (-2.75, 0) {};
		\node [style=none] (33) at (-4.5, 0) {};
		\node [style=none] (34) at (-1.75, 0) {};
		\node [style=none] (35) at (-1.5, 0) {};
		\node [style=none] (36) at (-4.5, -1) {};
		\node [style=none] (37) at (-1.5, -1) {};
		\node [style=none] (38) at (-4.125, 0) {};
		\node [style=none] (39) at (-3.375, 0) {};
		\node [style=none] (40) at (-4.125, -1) {};
		\node [style=none] (41) at (-3.375, -1) {};
		\node [style=none] (42) at (-3, 1.5) {};
		\node [style=none] (43) at (-3, -1.5) {};
		\node [style=none] (44) at (-4.5, 1.5) {};
		\node [style=none] (45) at (-4.5, -1.5) {};
		\node [style=none] (46) at (-1.5, -1.5) {};
		\node [style=none] (47) at (-1.5, 1.5) {};
		\node [style=none] (48) at (-3, -0.5) {};
		\node [style=none] (49) at (-1.5, -0.5) {};
		\node [style=none] (50) at (-4.5, 0.5) {};
		\node [style=none] (51) at (-3, 0.5) {};
		\node [style=none] (52) at (-4.25, 0.25) {};
		\node [style=none] (53) at (-3.25, 0.25) {};
		\node [style=none] (54) at (-4.25, -1.25) {};
		\node [style=none] (55) at (-3.25, -1.25) {};
		\node [style=none] (56) at (-6, 0) {$\triangleq$};
		\node [style=none] (57) at (-4.75, 1) {\tiny $A$};
		\node [style=none] (58) at (-4.75, 0) {\tiny $B$};
		\node [style=none] (59) at (-4.75, -1) {\tiny $C$};
		\node [style=none] (60) at (-3.625, 0) {\tiny $D$};
		\node [style=none] (61) at (-1.25, 1) {\tiny $E$};
		\node [style=none] (62) at (-1.25, 0) {\tiny $F$};
		\node [style=none] (63) at (-3.625, -1) {\tiny $G$};
		\node [style=none] (64) at (-1.25, -1) {\tiny $G$};
		\node [style=operadot] (65) at (4, -2) {\tiny$\mathbf{1}_G$};
		\node [style=none] (66) at (6.5, -0.25) {\tiny \textcolor{gray}{$\mathcal{P}(A\otimes D,E \otimes F)$}};
		\node [style=operadot] (67) at (4, -0.5) {\footnotesize$f$};
		\node [style=operadot] (68) at (5.5, -1.25) {\footnotesize$\otimes$};
		\node [style=operadot] (69) at (11.75, 0) {\footnotesize$\circ$};
		\node [style=none] (70) at (5.5, -2.25) {\tiny \textcolor{gray}{$\mathcal{P}(G,G)$}};
		\node [style=none] (71) at (9.25, -1.5) {\tiny \textcolor{gray}{$\mathcal{P}(A\otimes D \otimes G,E \otimes F \otimes G)$}};
		\node [style=operadot] (72) at (4, 1.75) {\tiny$\mathbf{1}_A$};
		\node [style=none] (73) at (5.5, 2) {\tiny \textcolor{gray}{$\mathcal{P}(A,A)$}};
		\node [style=operadot] (74) at (5.5, 1) {\footnotesize$\otimes$};
		\node [style=none] (75) at (1, 1) {};
		\node [style=none] (76) at (9.25, 1.25) {\tiny \textcolor{gray}{$\mathcal{P}(A \otimes B \otimes C,A\otimes D \otimes G)$}};
		\node [style=none] (77) at (3, 0.5) {\tiny \textcolor{gray}{$\mathcal{P}(B \otimes C,D \otimes G)$}};
		\node [style=none] (78) at (18, 0) {};
		\node [style=none] (79) at (15.25, -0.5) {\tiny \textcolor{gray}{$\mathcal{P}(A \otimes B \otimes C,E \otimes F \otimes G)$}};
		\node [style=none] (80) at (0, 0) {$\leftrightarrow$};
		\node [style=none] (81) at (-8, -5) {};
		\node [style=none] (82) at (-8, -6.5) {};
		\node [style=none] (83) at (-7.5, -6.5) {};
		\node [style=none] (84) at (-7.5, -5) {};
		\node [style=none] (85) at (-7.75, -5.75) {\tiny$f$};
		\node [style=none] (86) at (-9.5, -5.25) {};
		\node [style=none] (87) at (-8, -5.25) {};
		\node [style=none] (88) at (-7.5, -5.25) {};
		\node [style=none] (89) at (-7, -5.25) {};
		\node [style=none] (90) at (-8, -6.25) {};
		\node [style=none] (91) at (-9.5, -6.25) {};
		\node [style=none] (92) at (-7.5, -6.25) {};
		\node [style=none] (93) at (-7, -6.25) {};
		\node [style=none] (94) at (-9.5, -7) {};
		\node [style=none] (95) at (-7, -7) {};
		\node [style=none] (96) at (-9, -6.25) {};
		\node [style=none] (97) at (-8.5, -6.25) {};
		\node [style=none] (98) at (-9, -7) {};
		\node [style=none] (99) at (-8.5, -7) {};
		\node [style=none] (100) at (-9, -6.75) {};
		\node [style=none] (101) at (-9, -7.25) {};
		\node [style=none] (102) at (-8.5, -7.25) {};
		\node [style=none] (103) at (-8.5, -6.75) {};
		\node [style=none] (104) at (-9, -6.5) {};
		\node [style=none] (105) at (-9, -6) {};
		\node [style=none] (106) at (-8.5, -6) {};
		\node [style=none] (107) at (-8.5, -6.5) {};
		\node [style=none] (108) at (-8.75, -6.25) {\tiny$h$};
		\node [style=none] (109) at (-8.25, -7.5) {};
		\node [style=none] (110) at (-9.25, -7.5) {};
		\node [style=none] (111) at (-9.25, -5.75) {};
		\node [style=none] (112) at (-8.25, -5.75) {};
		\node [style=none] (113) at (-2.75, -5) {};
		\node [style=none] (114) at (-2.75, -6.5) {};
		\node [style=none] (115) at (-1.75, -6.5) {};
		\node [style=none] (116) at (-1.75, -5) {};
		\node [style=none] (117) at (-2.25, -5.75) {\tiny$f$};
		\node [style=none] (118) at (-4.5, -5.25) {};
		\node [style=none] (119) at (-2.75, -5.25) {};
		\node [style=none] (120) at (-1.75, -5.25) {};
		\node [style=none] (121) at (-1.5, -5.25) {};
		\node [style=none] (122) at (-2.75, -6.25) {};
		\node [style=none] (123) at (-4.5, -6.25) {};
		\node [style=none] (124) at (-1.75, -6.25) {};
		\node [style=none] (125) at (-1.5, -6.25) {};
		\node [style=none] (126) at (-4.5, -7.5) {};
		\node [style=none] (127) at (-1.5, -7.5) {};
		\node [style=none] (130) at (-4.125, -7.5) {};
		\node [style=none] (131) at (-3.375, -7.5) {};
		\node [style=none] (132) at (-3, -4.75) {};
		\node [style=none] (133) at (-3, -8.25) {};
		\node [style=none] (134) at (-4.5, -4.75) {};
		\node [style=none] (135) at (-4.5, -8.25) {};
		\node [style=none] (136) at (-1.5, -8.25) {};
		\node [style=none] (137) at (-1.5, -4.75) {};
		\node [style=none] (138) at (-4.5, -6.75) {};
		\node [style=none] (139) at (-1.5, -6.75) {};
		\node [style=none] (140) at (-4.5, -5.75) {};
		\node [style=none] (141) at (-3, -5.75) {};
		\node [style=none] (142) at (-4.25, -7) {};
		\node [style=none] (143) at (-3.25, -7) {};
		\node [style=none] (144) at (-4.25, -8) {};
		\node [style=none] (145) at (-3.25, -8) {};
		\node [style=none] (146) at (-6, -6.25) {$\triangleq$};
		\node [style=none] (147) at (-4.75, -5.25) {\tiny $A$};
		\node [style=none] (148) at (-4.75, -6.25) {\tiny $B$};
		\node [style=none] (149) at (-4.75, -7.5) {\tiny $C$};
		\node [style=none] (151) at (-1.25, -5.25) {\tiny $E$};
		\node [style=none] (152) at (-1.25, -6.25) {\tiny $F$};
		\node [style=none] (153) at (-3.625, -7.5) {\tiny $G$};
		\node [style=none] (154) at (-1.25, -7.5) {\tiny $G$};
		\node [style=none] (157) at (-4, -6.5) {};
		\node [style=none] (158) at (-4, -6) {};
		\node [style=none] (159) at (-3.5, -6) {};
		\node [style=none] (160) at (-3.5, -6.5) {};
		\node [style=none] (161) at (-3.75, -6.25) {\tiny$h$};
		\node [style=none] (162) at (-4, -6.25) {};
		\node [style=none] (163) at (-3.5, -6.25) {};
		\node [style=operadot] (164) at (10.5, -8.25) {\tiny$\mathbf{1}_G$};
		\node [style=operadot] (166) at (10.5, -6.25) {\footnotesize$f$};
		\node [style=operadot] (167) at (13.5, -7.25) {\footnotesize$\otimes$};
		\node [style=operadot] (168) at (16.5, -6.25) {\footnotesize$\circ$};
		\node [style=operadot] (171) at (10.5, -4.25) {\tiny$\mathbf{1}_A$};
		\node [style=operadot] (173) at (13.5, -5.25) {\footnotesize$\otimes$};
		\node [style=none] (177) at (18, -6.25) {};
		\node [style=none] (178) at (3, -3) {};
		\node [style=none] (179) at (18, -3) {};
		\node [style=none] (180) at (18, -3.25) {};
		\node [style=none] (181) at (10.5, -3.25) {};
		\node [style=none] (182) at (-8, -3) {\rotatebox{-90}{$\textcolor{purple}{\mapsto}$}};
		\node [style=operadot] (183) at (2.75, -6) {\footnotesize$h$};
		\node [style=operadot] (184) at (5.75, -6.25) {\footnotesize$\otimes$};
		\node [style=none] (185) at (4.25, -5.75) {\tiny \textcolor{gray}{$\mathcal{P}(B,D)$}};
		\node [style=none] (186) at (2, -7) {};
		\node [style=none] (187) at (4.25, -7) {\tiny \textcolor{gray}{$\mathcal{P}(C,G)$}};
		\node [style=none] (188) at (8.25, -5.5) {\tiny \textcolor{gray}{$\mathcal{P}(B \otimes C,D \otimes G)$}};
		\node [style=none] (189) at (7.25, -6.25) {};
		\node [style=none] (190) at (2, -4) {};
		\node [style=none] (191) at (2, -8.5) {};
		\node [style=none] (192) at (1, -7) {};
		\node [style=none] (193) at (0, -6.25) {$\leftrightarrow$};
		\node [style=none] (194) at (-3, -3) {\rotatebox{-90}{$\textcolor{purple}{\mapsto}$}};
	\end{pgfonlayer}
	\begin{pgfonlayer}{edgelayer}
		\draw (2.center)
			 to (1.center)
			 to (0.center)
			 to (3.center)
			 to cycle;
		\draw (5.center) to (6.center);
		\draw (7.center) to (8.center);
		\draw (11.center) to (12.center);
		\draw (10.center) to (15.center);
		\draw (16.center) to (9.center);
		\draw (13.center) to (17.center);
		\draw (18.center) to (14.center);
		\draw [style=kF] (19.center)
			 to (22.center)
			 to (21.center)
			 to (20.center)
			 to cycle;
		\draw (25.center)
			 to (24.center)
			 to (23.center)
			 to (26.center)
			 to cycle;
		\draw (28.center) to (29.center);
		\draw (30.center) to (31.center);
		\draw (34.center) to (35.center);
		\draw (33.center) to (38.center);
		\draw (39.center) to (32.center);
		\draw (36.center) to (40.center);
		\draw (41.center) to (37.center);
		\draw [style=Dcyan] (42.center) to (43.center);
		\draw [style=Dcyan] (45.center)
			 to (46.center)
			 to (47.center)
			 to (44.center)
			 to cycle;
		\draw [style=Dcyan] (48.center) to (49.center);
		\draw [style=Dcyan] (50.center) to (51.center);
		\draw [style=H] (52.center)
			 to (53.center)
			 to (55.center)
			 to (54.center)
			 to cycle;
		\draw [style=gray] (67) to (68);
		\draw [style=gray] (65) to (68);
		\draw [style=gray] (68) to (69);
		\draw [style=gray] (72) to (74);
		\draw [style=gray] (75.center) to (74);
		\draw [style=gray] (74) to (69);
		\draw [style=gray] (69) to (78.center);
		\draw (83.center)
			 to (82.center)
			 to (81.center)
			 to (84.center)
			 to cycle;
		\draw (86.center) to (87.center);
		\draw (88.center) to (89.center);
		\draw (92.center) to (93.center);
		\draw (91.center) to (96.center);
		\draw (97.center) to (90.center);
		\draw (94.center) to (98.center);
		\draw (99.center) to (95.center);
		\draw [style=kF] (100.center)
			 to (103.center)
			 to (102.center)
			 to (101.center)
			 to cycle;
		\draw (105.center)
			 to (106.center)
			 to (107.center)
			 to (104.center)
			 to cycle;
		\draw [style=Dpurp] (110.center)
			 to (111.center)
			 to (112.center)
			 to (109.center)
			 to cycle;
		\draw (115.center) to (114.center);
		\draw (114.center) to (113.center);
		\draw (113.center) to (116.center);
		\draw (116.center) to (115.center);
		\draw (118.center) to (119.center);
		\draw (120.center) to (121.center);
		\draw (124.center) to (125.center);
		\draw (126.center) to (130.center);
		\draw (131.center) to (127.center);
		\draw [style=Dcyan] (132.center) to (133.center);
		\draw [style=Dcyan] (135.center)
			 to (136.center)
			 to (137.center)
			 to (134.center)
			 to cycle;
		\draw [style=Dcyan] (138.center) to (139.center);
		\draw [style=Dcyan] (140.center) to (141.center);
		\draw [style=H] (142.center)
			 to (143.center)
			 to (145.center)
			 to (144.center)
			 to cycle;
		\draw (158.center)
			 to (159.center)
			 to (160.center)
			 to (157.center)
			 to cycle;
		\draw (123.center) to (162.center);
		\draw (163.center) to (122.center);
		\draw [style=Dpurp] (133.center)
			 to (135.center)
			 to (140.center)
			 to (141.center)
			 to cycle;
		\draw [style=gray] (166) to (167);
		\draw [style=gray] (164) to (167);
		\draw [style=gray] (167) to (168);
		\draw [style=gray] (171) to (173);
		\draw [style=gray] (173) to (168);
		\draw [style=gray] (168) to (177.center);
		\draw [style=gray] (183) to (184);
		\draw [style=gray] (192.center)
			 to (186.center)
			 to [in=-180, out=0] (184);
		\draw [style=gray] (184)
			 to (189.center)
			 to [in=-180, out=0] (173);
		\draw [style=Dpurp] (189.center)
			 to (191.center)
			 to (190.center)
			 to cycle;
		\draw [style=brace edge] (179.center) to (178.center);
		\draw [style=brace edge] (181.center) to (180.center);
	\end{pgfonlayer}
\end{tikzpicture}
\]
\end{example}

\begin{remark}
The extension of the current theory to accommodate parameter sharing between universal approximators is conceptually straightforward but technically involved. Parameter sharing corresponds to the ability to reuse -- i.e. copy -- data between open wires in the operad $\mathcal{O}_\mathcal{P}$, which amounts to having a cartesian operad.
\end{remark}

\subsection{Coherence of SIMD and universal approximators}\label{appendix:coherence}
Now, given a (presumed cartesian) coloured PROP $\mathcal{P}$, we have two extensions: the free tensoring $\mathbb{F}(\mathcal{P})^\otimes$ which formalises tick-notation for SIMD, and the hom-operad $\mathcal{O}_\mathcal{P}$ in which universal approximators and expressive reductions are formalised as open morphisms and operad precomposition respectively. The remaining formal question is how to combine the extensions in a consistent manner. We achieve this by constructing a symmetric monoidal structure $\mathcal{C}(\mathcal{O}_\mathcal{P})$ on top of $\mathcal{O}_\mathcal{P}$ which will contain $\mathcal{P}$ as a full (symmetric monoidal) subcategory and will be equivalent to the PROP $\hat{\mathcal{P}}$ obtained by extending the signature generating $\mathcal{P}$ with a black-box for every input-output type: in this way $\mathcal{C}(\mathcal{O}_\mathcal{P})$ will syntactically represent extending $\mathcal{P}$ with black-boxes, while the underlying $\mathcal{O}_\mathcal{P}$ structure will modularly govern the behaviour of expressive reductions.

\begin{construction}[The symmetric monoidal category $\mathcal{C}(\mathcal{O}_\mathcal{P})$ of the hom-operad $\mathcal{O}_\mathcal{P}$ of the PROP $\mathcal{P}$]\label{cons:backtobasics}
We define a symmetric monoidal category $\mathcal{C}(\mathcal{O}_\mathcal{P})$ from the data of $\mathcal{O}_\mathcal{P}$ of a cartesian $\mathcal{P}$ as follows. First, recall from Example \ref{ex:bridge} that by cartesianity of $\mathcal{P}$, $\mathcal{P}(I,I)$ is a singleton, so without loss of generality we may denote the arbitrary morphism of $\mathcal{O}_\mathcal{P}$ as $f: \mathbf{X} \rightarrow \mathcal{P}(\mathbf{A},\mathbf{B})$ of $\mathcal{O}_\mathcal{P}$, for some $\mathbf{X}$, and for $\mathbf{A},\mathbf{B}$ objects of $\mathcal{P}$. For each such $f$, we define a morphism $f: \mathbf{A} \rightarrow \mathbf{B}$ in $\mathcal{C}(\mathcal{O}_\mathcal{P})$, and we define the objects $\mathcal{C}(\mathcal{O}_\mathcal{P})$ by proxy, via the identity morphisms $1_\mathbf{X}$ in $\mathbf{P}$, which become states $1_\mathbf{X}: \{\star\} \rightarrow \mathcal{P}(\mathbf{X},\mathbf{X})$, which become morphisms $1_\mathbf{X}: \mathbf{X} \rightarrow \mathbf{X}$ $\mathcal{C}(\mathcal{O}_\mathcal{P})$. Similarly, we define the braidings $\theta_{\mathbf{X},\mathbf{Y}}: \mathbf{X} \otimes \mathbf{Y} \rightarrow \mathbf{Y} \otimes \mathbf{X}$.
\[\begin{tikzpicture}
	\begin{pgfonlayer}{nodelayer}
		\node [style=operadot] (0) at (-4, 0) {\tiny$\mathbf{1}_X$};
		\node [style=none] (1) at (-1.75, 0) {};
		\node [style=none] (2) at (-2.5, 0.25) {\tiny \textcolor{gray}{$\mathcal{P}(X,X)$}};
		\node [style=operadot] (3) at (2, 0) {\tiny$\theta_{X,Y}$};
		\node [style=none] (4) at (6.5, 0) {};
		\node [style=none] (5) at (4.75, 0.25) {\tiny \textcolor{gray}{$\mathcal{P}(X \otimes Y,Y \otimes X)$}};
	\end{pgfonlayer}
	\begin{pgfonlayer}{edgelayer}
		\draw [style=gray] (0) to (1.center);
		\draw [style=gray] (3) to (4.center);
	\end{pgfonlayer}
\end{tikzpicture}
\]
We define sequential and parallel composition in $\mathcal{C}(\mathcal{O}_\mathcal{P})$ using the $\circ$ and $\otimes$ constructors that are definitionally available in $\mathcal{O}_\mathcal{P}$.
\[\begin{tikzpicture}
	\begin{pgfonlayer}{nodelayer}
		\node [style=argoperadot] (0) at (0.125, 0.75) {\footnotesize$f$};
		\node [style=operadot] (1) at (1.75, 0) {\footnotesize$\otimes$};
		\node [style=operadot] (2) at (-11, 0) {\footnotesize$\circ$};
		\node [style=argoperadot] (3) at (0.125, -0.75) {\footnotesize$g$};
		\node [style=argoperadot] (4) at (-12.625, 0.75) {\footnotesize$f$};
		\node [style=argoperadot] (5) at (-12.625, -0.75) {\footnotesize$g$};
		\node [style=none] (6) at (6.25, 0) {};
		\node [style=none] (7) at (-8.5, 0) {};
		\node [style=none] (11) at (-11, 1) {\tiny \textcolor{gray}{$\mathcal{P}(A,B)$}};
		\node [style=none] (12) at (1.75, 1) {\tiny \textcolor{gray}{$\mathcal{P}(A,B)$}};
		\node [style=none] (13) at (-11, -1) {\tiny \textcolor{gray}{$\mathcal{P}(B,C)$}};
		\node [style=none] (14) at (-9.25, 0.25) {\tiny \textcolor{gray}{$\mathcal{P}(A,C)$}};
		\node [style=none] (15) at (1.75, -1) {\tiny \textcolor{gray}{$\mathcal{P}(C,D)$}};
		\node [style=none] (16) at (4.5, 0.25) {\tiny \textcolor{gray}{$\mathcal{P}(A \otimes C,B \otimes D)$}};
		\node [style=operadot] (20) at (-6.5, 0) {\tiny$g \circ f$};
		\node [style=none] (21) at (-4, 0) {};
		\node [style=none] (22) at (-4.75, 0.25) {\tiny \textcolor{gray}{$\mathcal{P}(A,C)$}};
		\node [style=none] (23) at (-7.75, 0) {$\triangleq$};
		\node [style=operadot] (24) at (8.25, 0) {\tiny$f \otimes g$};
		\node [style=none] (25) at (12.75, 0) {};
		\node [style=none] (27) at (11, 0.25) {\tiny \textcolor{gray}{$\mathcal{P}(A \otimes C,B \otimes D)$}};
		\node [style=none] (28) at (7, 0) {$\triangleq$};
		\node [style=none] (29) at (-13.75, 1.25) {};
		\node [style=none] (30) at (-13.75, 0.25) {};
		\node [style=none] (31) at (-13.75, -0.25) {};
		\node [style=none] (32) at (-13.75, -1.25) {};
		\node [style=none] (33) at (-1, 1.25) {};
		\node [style=none] (34) at (-1, 0.25) {};
		\node [style=none] (35) at (-1, -0.25) {};
		\node [style=none] (36) at (-1, -1.25) {};
	\end{pgfonlayer}
	\begin{pgfonlayer}{edgelayer}
		\draw [style=gray] (0) to (1);
		\draw [style=gray] (3) to (1);
		\draw [style=gray] (4) to (2);
		\draw [style=gray] (5) to (2);
		\draw [style=gray] (1) to (6.center);
		\draw [style=gray] (2) to (7.center);
		\draw [style=gray] (20) to (21.center);
		\draw [style=gray] (24) to (25.center);
		\draw [style=gray] (4) to (29.center);
		\draw [style=gray] (4) to (30.center);
		\draw [style=gray] (5) to (31.center);
		\draw [style=gray] (5) to (32.center);
		\draw [style=gray] (0) to (33.center);
		\draw [style=gray] (0) to (34.center);
		\draw [style=gray] (3) to (35.center);
		\draw [style=gray] (3) to (36.center);
	\end{pgfonlayer}
\end{tikzpicture}
\]
By the construction of $\mathcal{O}_\mathcal{P}$, $\mathcal{C}(\mathcal{O}_\mathcal{P})$ satisfies all symmetric monoidal coherences.
\end{construction}

\begin{construction}[$\hat{\mathcal{P}}$]
Given a cartesian coloured PROP $\mathcal{P}$ generated by some signature $\Sigma$ consisting of colours, morphisms and equational relations, define $\hat{\mathcal{P}}$ to be the coloured PROP obtained by extending $\Sigma$ with (1) a morphism $\bot: \mathbf{X} \rightarrow \mathbf{Y}$ for each pair of objects of $\mathcal{P}$ (2) copy-delete coherence equations for the $\bot$s; by Fox's theorem, the presence of such copy-delete coherences for all morphisms is equivalent to cartesian monoidality \cite{foxCoalgebrasCartesianCategories1976}.
\end{construction}

\begin{theorem}\label{thm:bigboy}
If a coloured PROP $\mathcal{P}$ is cartesian, then $\mathcal{P} \hookrightarrow \mathcal{C}(\mathcal{O}_\mathcal{P}) \simeq \hat{\mathcal{P}}$. In prose, $\mathcal{P}$ embeds (as a cartesian monoidal category) into $\mathcal{C}(\mathcal{O}_\mathcal{P})$, which is equivalent to $\hat{\mathcal{P}}$ (as a cartesian monoidal category).
\begin{proof}
First, to see that $\mathcal{P}$ embeds as a full subcategory of $\mathcal{C}(\mathcal{O}_\mathcal{P})$, observe that (1) morphisms $f: \mathbf{A} \rightarrow \mathbf{B}$ of $\mathcal{P}$ are in definitional one-to-one correspondence with states $f: \{\star\} \rightarrow \mathcal{P}(\mathbf{A},\mathbf{B})$ of $\mathcal{O}_\mathcal{P}$, (2) the constructors available in $\mathcal{C}(\mathcal{O}_\mathcal{P})$ can only combine states of $\mathcal{O}_\mathcal{P}$ to form other states of $\mathcal{O}_\mathcal{P}$, and (3) all morphisms of $\mathcal{P}$ can be obtained in this way, as all admit a decomposition in terms of $\circ$, $\otimes$, and the generators of $\mathcal{P}$.

Second, to see that $\mathcal{C}(\mathcal{O}_\mathcal{P})$ is (cartesian monoidally) equivalent to $\hat{\mathcal{P}}$, observe that the open inputs $\mathbf{X}$ of morphisms $f: \mathbf{X} \rightarrow \mathcal{P}(\mathbf{A},\mathbf{B})$ may be written without loss of generality as lists $[\mathcal{P}(\mathbf{X}_1,\mathbf{Y}_1),\cdots,\mathcal{P}(\mathbf{X}_n,\mathbf{Y}_n)]$, which fall in bijective correspondence with states $\{\star\} \rightarrow \mathbf{X}$ by first sending $\mathbf{X}$ to $\mathcal{C}(\mathcal{O}_{\hat{\mathcal{P}}})$ along the embedding $\mathcal{P} \hookrightarrow \hat{\mathcal{P}}$, and then precomposing each open $\mathcal{P}(\mathbf{X}_n,\mathbf{Y}_n)$ with $\bot: \{\star\} \rightarrow \mathcal{P}(\mathbf{X}_i,\mathbf{Y}_i)$ in $\mathcal{O}_{\hat{\mathcal{P}}}$, obtained from the $\bot: \mathbf{X}_i \rightarrow \mathbf{Y}_i$ that are definitionally available in $\hat{\mathcal{P}}$. So $\mathcal{C}(\mathcal{O}_\mathcal{P})$ is equivalent to the full subcategory of $\mathcal{C}(\mathcal{O}_{\hat{\mathcal{P}}})$ generated by states of $\mathcal{O}_{\hat{\mathcal{P}}}$, which is by part one of this proof equivalent to $\hat{\mathcal{P}}$, so we have the claim.
\end{proof}
\end{theorem}

\begin{defn}[SIMD notation with universal approximators]\label{defn:SIMDformaldefn}
Our string diagrams take their formal semantics in $\mathbb{F}(\mathcal{C}(\mathcal{O}_{\mathbf{CartSp}}))^\otimes$, and the semantics of expressivity reduction of universal approximators is formalised by precomposition in $\mathcal{O}_{\mathbf{CartSp}}$.
\end{defn}

\begin{remark}\label{rem:porbs}
This construction extends to any symmetric monoidal category $(\mathcal{C},\otimes,I)$ with $\mathcal{C}(I,I)$ a singleton, which would notably include \textbf{Set} and Markov categories \cite{nlab_markov_category} that formalise probabilistic processes. The key step is the ability to remove the composition-dependence of the domains $\mathcal{P}(I,I)^n$ in Example \ref{ex:bridge}, which may alternately be achieved by e.g. the use of a comonoid $\delta: \mathcal{P}(I,I) \rightarrow \mathcal{P}(I,I) \times \mathcal{P}(I,I)$ to compose e.g. $F(-) \triangleq (- \mapsto - \otimes f) \in [\mathcal{P}(I,I) \rightarrow \mathcal{A,B}]$ and $G(-) \triangleq (- \mapsto - \otimes g) \in [\mathcal{P}(I,I) \rightarrow \mathcal{B,C}]$ sequentially as $\big(- \mapsto G(\delta_1(-)) \circ F(\delta_2(-))\big) \in [\mathcal{P}(I,I) \rightarrow \mathcal{A,C}]$.
\end{remark}
\section{Detailed graphical derivations}
\label{sec:detailed_appendix}

In this section we present in detail several sequences of graphical rewrites.
Some (Appendix~\ref{ssec:appendix_bahd_vas}, 
\ref{app:lemmaproofs}) will use the the less precise notation with coloured wires with ticks, which in principle could leave some ambiguity with respect to indices and dimension sizes, but is easier to work with, especially if we are only interested in the high-level structure of the model.
Others (Appendices~\ref{ssec:appendix_generic_sim}, \ref{ssec:appendix_vas_to_linear}) will use fully precise notation to demonstrate that one can always be fully precise in this framework if desired. 
In particular, in the precise form, we will sometimes use superscripts to disambiguate between different indices with the same numerical value.
For instance, if we draw an $[s,s]$ sized wire representing the attention matrix entering an $s$-SIMD box, this leaves it ambiguous with dimension is being expanded by the SIMD box.
Thus, we write $[s^{(q)},s^{(k)}]$ for the tensor on the wire, and also write $s^{(q)}$ for the SIMD box.
Other cases where disambiguation like this may be required are tensor contractions.

Note in principle one could include such superscripts on all tensor indices of the diagram, and do so in a consistent way, due to the nature of tensor contraction -- however, we will only include superscripts below if disambiguation is required.

\subsection{Motivating Vaswani et al. from Bahdanau et al.}
\label{ssec:appendix_bahd_vas}

Here we present an account of how one could arrive at the self-attention mechanism using our graphical rewrites, starting from the Bahdanau et al. architecture.
We use the less precise coloured, ticked wire convention.
This is not intended to be a highly rigorous explanation for, nor are we making precise claims about the expressive power of these models.
Instead this shows how feasibly, using our rewrites can recover relationships between models which were in reality derived from each other, via human inspiration.

Everything in this sequence of steps proceeds by expressivity rewrites involving trainable boxes, except one rewiring step that involves a structural change, in which we plug inputs into what were originally the hidden states passed into the RNN-decoder cells.

The Bahdanau et al. architecture is depicted below: we use slanted thick blue ticks below to represent the reversal operation on the sequence dimension (for the bi-RNN), and the merging the sequences at the end of the bi-RNN via concatenating corresponding vector pairs.
The four trainable boxes below are -- two RNN-cells for the bi-RNN encoder, the score function (which has trainable weights), and the decoder RNN cell.
\[\input{bahdvas/0_j.tikz}\] 
We break apart the decoder RNN cell and discard most of its internal dependency.
This is a move that reduces a lot of power, as it causes decoder to lose access to other information about decoder outputs and become essentially parallel.
\[\quad \textcolor{gray}{\mapsto} \quad \input{bahdvas/1_j.tikz}\] 

\[
\quad = \quad 
\input{bahdvas/1_2_j.tikz}\] 
\[
\quad = \quad 
\input{bahdvas/2_j.tikz}\] 

\[
\quad = \quad 
\input{bahdvas/3_j.tikz}\] 

Here we do the one non-expressivity rewrite -- we effectively turn this into a self-attention mechanism, by plugging the input data stream into what are effectively the queries.

\[ \textcolor{gray}{\overset{\star}{\mapsto}} \quad \input{bahdvas/4_j.tikz}\]

Now we do some rewriting to get rid of the bi-RNN encoder.

\[ \textcolor{gray}{\mapsto} \quad \input{bahdvas/5_j.tikz}\]

\[ \textcolor{gray}{\mapsto} \quad \input{bahdvas/6_j.tikz}\]

In the next step, we omit one half of the bi-RNN output from our diagram. 
This can be implemented for example by having the discarded half of the bi-RNN output a matrix of 0s, with that portion of the data stream being discarded by the trainable boxes further down the line.
\[ \textcolor{gray}{\mapsto} \quad \input{bahdvas/7_j.tikz}\]

Now shrink down the orange SIMD boxes, and recall matrix multiplications have a copy-SIMD structure.
\[ = \quad \input{bahdvas/8_j.tikz}\]

\[ = \quad \input{bahdvas/8_2_j.tikz}\]

Now we specify the score function to be a scaled-dot product, with learnable query and key functions.

\[ \textcolor{gray}{\mapsto} \quad \input{bahdvas/10_j.tikz}\]

\[ = \quad \input{bahdvas/11_j.tikz}\]

Push the first trainable box through the copy node.
We use a variable $x$ to indicate that two of the trainable boxes now have shared weights.
\[ = \quad \input{bahdvas/12_2_j.tikz}\]
If we assume the $x$ box is invertible (a reasonable assumption for learned square value matrices, say), we may formally absorb it into the traniable box corresponding to keys matrix.
\[ \textcolor{gray}{\mapsto} \quad \input{bahdvas/14_j.tikz}\]
Thus arrive at the usual self-attention mechanism (one attention head).

\subsection{Attention with generic similarity function}
\label{ssec:appendix_generic_sim}

As a demonstration of fully precise reasoning using SIMD boxes, we start from Equation ~\ref{eqn:similarityattention} and show in explcit detail how to obtain the computational graph for an attention mechanism with unspecified similarity function (Diagram~\ref{diag:transformerwithsimilarity}).
Equation ~\ref{eqn:similarityattention} takes the following diagrammatic form:
\[\input{appendix_general_sim/similarity_eqn_austere.tikz}\]

We have such a diagram for all $1\leq i\leq s$.
These $s$ copies can be compiled into a single diagram by wrapping the diagram above in the appropriate SIMD box.
\[\input{appendix_general_sim/similarity_austere_step1.tikz}\]

Now we apply SIMD box rewrites to bring this into a nicer form. 
Firstly, we separate out the similarity function.
\[\input{appendix_general_sim/similarity_austere_step2.tikz}\]

Copy maps are SIMD under the hood, so it can separate out too:
\[\input{appendix_general_sim/similarity_austere_step3.tikz}\]

Next, pull out the vector of 1's from the SIMD box, incurring a copy node at the interface.
\[\input{appendix_general_sim/similarity_austere_step4.tikz}\]

Recalling that matrix multiplications, and tensor contractions more generally, have a nested copy-SIMD structure, we can split off more functions from the SIMD box.
\[\input{appendix_general_sim/similarity_austere_step5.tikz}\]

We split off again, absorbing a copy-SIMD box into the final matrix multiplication, which takes us finally to 
Diagram~\ref{diag:transformerwithsimilarity} for an abstract attention mechanism involving an unspecified similarity function.
We include this diagram here again in fully precise form (one adds the preparation of the queries, keys, and values at the start to get back the diagram in the main text).
\[\input{appendix_general_sim/similarity_eqn_final.tikz}\]
This diagram is a common parent of both the vanilla transformer attention mechanism and the linear transformation attention mechanisms.
If we insert $sim(x,y)=\exp\left(\frac{x\cdot y}{\sqrt{d_k}}\right)$ into Diagram~\ref{diag:transformerwithsimilarity} 
we can recover the usual scaled dot-product attention of Vaswani et al.

\subsection{Linearizing the transformer} 
\label{ssec:appendix_vas_to_linear}

To obtain the general form of the family of linear transformers, we choose $sim(x,y)=\phi(x)\cdot\phi(y)$, as described in the main body, 
giving Diagram~\ref{diag:linear_init}.
We show step-by-step how to get from this diagram to the diagram for the actual computational graph of the linear transformer (Diagram~\ref{diag:lineartransformer}).

Firstly, we present Diagram~\ref{diag:linear_init} in the fully detailed notation.
\[\input{appendix_linear_transformer/linear_initial_austere.tikz}\]
Now, pull the $\phi$'s out of the SIMD box.
\[\input{appendix_linear_transformer/linear_austere_step1.tikz}\]
Observe the two nested copy-SIMD boxes around the vector dot-product just forms a matrix-matrix multiplication.
Further, for convenience we expand again the $s^{(q)}$-SIMD box at the end of the graph to include the final matrix multiplication.
This yields:
\[\input{appendix_linear_transformer/linear_austere_step2.tikz}\]

We apply the associativity of matrix multiplication inside this SIMD box.
\[\input{appendix_linear_transformer/linear_austere_step3.tikz}\]
Next, we push the first matrix multiplication through the copy (and revert the $s_q$-SIMD box at the end to only include the final operation):
\[\input{appendix_linear_transformer/linear_austere_step4.tikz}\]

Now, we apply two more instances of associativity of matrix multiplication, which takes us to Diagram~\ref{diag:lineartransformer}: 
\[\input{appendix_linear_transformer/linear_final_austere.tikz}\]

\subsection{Proofs of Lemmas 5.1 and 5.2}
\label{app:lemmaproofs}

\begin{lemma*}
\[\input{greenequivs/copmon.tikz} \quad \equiv \quad \input{greenequivs/proc.tikz}\]
\begin{proof}
\[\input{greenequivs/proc.tikz} \quad \mapsto \quad \input{greenequivs/p2cm.tikz} \quad = \quad \input{greenequivs/copmon.tikz} \quad \mapsto \quad \input{greenequivs/cm2p.tikz} \quad = \quad \input{greenequivs/proc.tikz}\]
\end{proof}
\end{lemma*}

\begin{lemma*}
\[\input{greenequivs/bmon.tikz} \quad \equiv \quad \input{greenequivs/bom.tikz}\]
\begin{proof}
\[\input{greenequivs/bom.tikz} \quad \mapsto \quad \input{greenequivs/m2b.tikz} \quad = \quad \input{greenequivs/bmon.tikz}\]
\[\mapsto \quad \input{greenequivs/b2m.tikz} \quad = \quad \input{greenequivs/bom.tikz}\]
\end{proof}
\end{lemma*}
\section{Experimental Details}
\label{app:experiment}

\begin{figure}
    \centering
    \includegraphics[width=\textwidth]{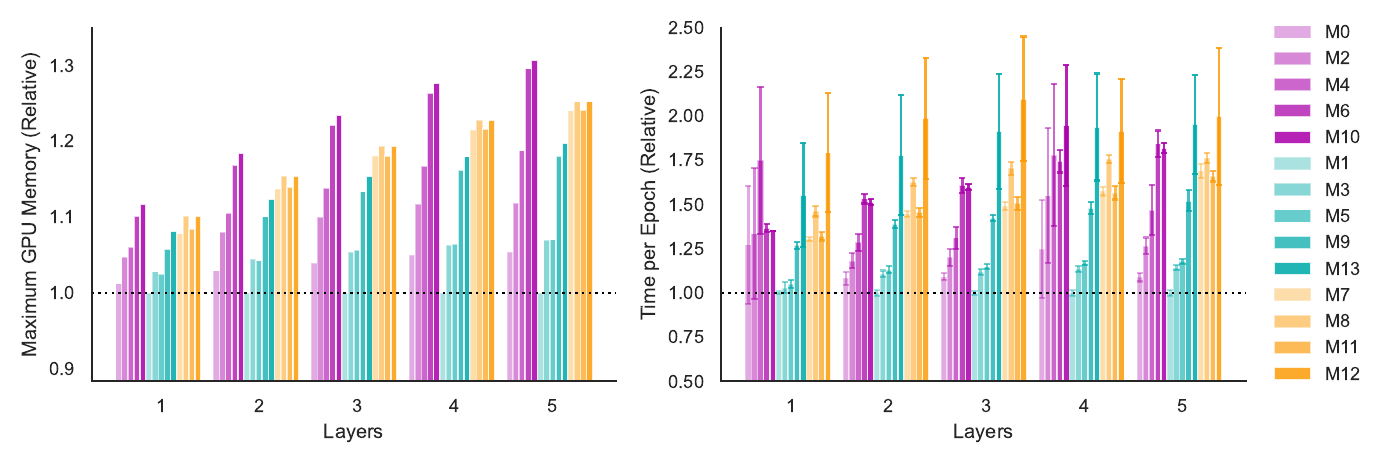}
    \caption{Relative running times and GPU memory required for each of the 14 proposed attention mechanisms, for models of between one and five layers, using the hyperparameters given in Appendix \ref{app:experiment}. The data is scaled so that the performance of M1 (scaled-dot-product attention as in \cite{vaswani2017attention}) is fixed to one (independently for each number of layers), in order to be somewhat platform-independent. All experiments were performed on NVIDIA A30 GPUs. The reported error bars are one standard deviation.}
    \label{fig:appendix-plot}
\end{figure}

To compare the 14 proposed attention mechanisms, we trained them \emph{ab initio} on a language modelling task using the Penn Treebank corpus \cite{marcus1993building}. Our implementation was built with PyTorch, and uses the built-in implementation of FlashAttention-2 \cite{dao2023flashattention2} to implement the \texttt{AttPrep} and \texttt{AttApply} generators. We use the CUDA implementation of causally-masked linear attention from \cite{katharopoulos2020transformers} to implement \texttt{LinAttPrep} and \texttt{LinAttApply}. All trainable universal approximators specified in the generators are implemented as an MLP with one hidden layer with the same dimension as the input - note that this includes the transformations creating the key and query inputs for the \texttt{AttPrep} generator and value inputs for the \texttt{AttApply} generator, in contrast with the usual implementations of these transformations as linear mappings. 

From each attention mechanism we followed \cite{vaswani2017attention} to construct decoder-only Transformer-style models. The method of splitting and recombining the model dimension into multiple attention heads, the placement of residuals, MLPs, and layer normalization was taken directly from the Transformer, with the exception of placing an additional layer normalization after each \texttt{(Lin)AttApply} generator, as recommended in \cite{qin2022devil}. We trained the model with causal masking, with a stride equal to half of the context length. For evaluation, we used the provided test set with a stride of one (so each element is given the maximum context). See below for the hyperparameter choices; an implementation is provided in the supplementary material.

\begin{center}
    \begin{tabular}{llll}
        Attention Heads & 4 & Optimizer & Adam \\
        Layers & 1--5 & Learning Rate Scheduler & \cite{vaswani2017attention} \\
        Model Dimension & 512 & Initial LR Scaling & 0.9 -- 1.1 \\
        Feedforward Dimension & 2048 & Warmup Iterations & 1500 \\
        Model Dimension (per head) & 128 & Weight Decay & 0.0001 \\
        MLP Hidden Dimension (within head) & 128 & Context Length & 35 \\
        Epochs & 20 & & \\
    \end{tabular}
\end{center}

Each model ran on a single NVIDIA A30 GPU, we report maximum GPU memory usage and wallclock time per epoch normalized to M1, which corresponds to the original Transformer, in Figure~\ref{fig:appendix-plot}. Note that the optimal context length for this dataset (at least in our testing) was very short (only 35), so we do not see any speed advantage from the models that are built using \texttt{LinAttPrep/Apply}. This is acceptable as the primary purpose of these experiments was to measure only their relative performances in-task. Similarly, we can see that our models do not come close to state of the art in this task. This is not surprising, as our models are small and we don't do any pre-training - we list some published results for this task below for comparison.
\begin{center}
    \begin{tabular}{ll}
         \toprule
         Test PPL & Model \\
         \midrule
         76.16 & Ours (M11, 4 layers) \\
         78.4 & LSTM \cite{zaremba2015recurrent} \\
         54.55 & Transformer-XL \cite{dai2019transformer}\\
         44.9 & Mogrifier LSTM \cite{melis2020mogrifier} \\
         35.76 & GPT-2 (zero-shot) \cite{radford2019language}\\
         20.5 & GPT-3 (zero-shot) \cite{brown2020language} \\
         \bottomrule
    \end{tabular}
\end{center}

\end{document}